\pdfoutput=1
\documentclass[10pt]{article}
\usepackage{graphicx}
\usepackage{subcaption} 
\usepackage{amsmath}
\usepackage{amssymb}
\usepackage{algorithmic}
\usepackage{algorithm}
\usepackage{booktabs} 

\numberwithin{equation}{section}
\usepackage{url}
\usepackage{hyperref}

\date{}

\author{Brian K. Vogel \\
brian@brianvogel.com}

\title{An NMF-Based Building Block for Interpretable Neural Networks With Continual Learning}

\begin{document}
\maketitle

\begin{abstract}
  

Existing learning methods often struggle to balance interpretability and predictive performance. While models like nearest neighbors and non-negative matrix factorization (NMF) offer high interpretability, their predictive performance on supervised learning tasks is often limited. In contrast, neural networks based on the multi-layer perceptron (MLP) support the modular construction of expressive architectures and tend to have better recognition accuracy but are often regarded as black boxes in terms of interpretability. Our approach aims to strike a better balance between these two aspects through the use of a building block based on NMF that incorporates supervised neural network training methods to achieve high predictive performance while retaining the desirable interpretability properties of NMF. We evaluate our Predictive Factorized Coupling (PFC) block on small datasets and show that it achieves competitive predictive performance with MLPs while also offering improved interpretability. We demonstrate the benefits of this approach in various scenarios, such as continual learning, training on non-i.i.d. data, and knowledge removal after training. Additionally, we show examples of using the PFC block to build more expressive architectures, including a fully-connected residual network as well as a factorized recurrent neural network (RNN) that performs competitively with vanilla RNNs while providing improved interpretability. The PFC block uses an iterative inference algorithm that converges to a fixed point, making it possible to trade off accuracy vs computation after training but also currently preventing its use as a general MLP replacement in some scenarios such as training on very large datasets. We provide source code at \url{https://github.com/bkvogel/pfc}

\end{abstract}

\section{Introduction}
\label{sec:intro}

Current neural networks are often the best performing models on a range of challenging problems. Commonly used neural architectures include fully-connected and convolutional networks, various Recurrent Neural Networks (RNNs), and transformers \cite{vaswani2017attention_v3}. These architectures and others make use of a relatively small number of basic building block types so that the differences between the various architectures are mainly due to which blocks are used and how they are inter-connected. Perhaps the single most fundamental building block is the Multi-Layer Perceptron (MLP) \cite{rosenblatt1958perceptron} \cite{ivakhnenko1965cybernetic} \cite{amari1967theory} since it is the smallest block that provides for learning arbitrarily complex non-linear functions and also tends to account for the bulk of the learnable parameters in existing architectures \cite{cybenko1989approximation} \cite{hornik1989multilayer}. The remaining building blocks are mainly intended to make the optimization process more efficient and/or serve as regularizers to help prevent over-fitting; examples of these include residual (i.e., skip) connections, LayerNorm \cite{ba2016layer}, and Dropout \cite{srivastava2014dropout} layers. Additional building blocks with learnable parameters are sometimes used for specialized architectures, such as linear embedding layers for the case of discrete-valued inputs and to provide the sequence-positional information in the transformer, for example. The transformer includes all of the above-mentioned blocks, as well as the attention block, which itself has an MLP interpretation, as described on the last page of \cite{vaswani2017attention_v3}. We therefore consider the currently popular neural architectures as being ``MLP-based''.

The existing MLP-based models do have some drawbacks, however. The first is that these architectures are often criticized as being ``black boxes'' lacking interpretability. Additionally and perhaps partially due to the first issue is that they tend to experience training difficulties as the distribution of examples starts to deviate from the i.i.d. assumption. This is often referred to as the ``catastrophic forgetting'' problem in the literature and it results in poor continual learning performance \cite{mccloskey1989catastrophic}.

As an alternative to the MLP, there are other existing machine learning methods with better interpretability properties. A simple example is the k-nearest-neighbors (k-NN) algorithm for classification and regression \cite{Cover1967NearestNP}. Since there is no explicit learning step other than retaining the training examples as they become available to the model, k-NN trivially supports continual learning as well as knowledge removal. Other prototype-based algorithms such as LVQ \cite{kohonen1988learning} \cite{kohonen1990improved} can be considered learnable extensions of nearest neighbors and also have good interpretability.

Another method that is often noted for its desirable interpretability properties is non-negative matrix factorization (NMF). In contrast to other prototype-based learning methods such as k-NN and LVQ in which the prototypes represent entire examples, in NMF they instead represent prototypical ``parts'' exemplars which are additively combined to reconstruct the input. 

These more interpretable methods unfortunately also have drawbacks that have prevented them from competing with MLP-based models in terms of predictive accuracy on supervised tasks. Methods such as k-NN and LVQ are not fully differentiable, preventing them from being used as general neural network building blocks that can be composed like the MLP to create expressive architectures supporting arbitrary differentiable loss functions tailored to the task at hand. Although NMF is potentially differentiable, the literature is mainly concerned with its usage for unsupervised learning tasks in which the factor matrices are optimized to minimize an input reconstruction loss only, rather than a supervised classification or regression loss.

\subsection{Contributions}

The main point of this work is to make some progress on developing learning methods with a better balance of interpretability and predictive performance compared to the existing MLP-based approaches. In particular, we make the following contributions:

\begin{itemize}
\item In Section \ref{sec:factorized_layers} we introduce a new neural network building block called the Predictive Factorized Coupling (PFC) block as a more interpretable alternative to the MLP. Since its declarative model is still matrix factorization, it potentially retains the interpretable parts-based nature of NMF, but extends it to support its use as a general differentiable predictive module.
\item In Section \ref{sec:frp_1_layer} we demonstrate that the PFC block has competitive accuracy with the (fully-connected) MLP on MNIST, Fashion MNIST, and CIFAR10 classification. In Section  \ref{sec:frp_2_layer} we demonstrate that a (fully-connected) residual network consisting of two PFC blocks is also able to be trained without optimization difficulties and that it performs similarly. Section \ref{sec:vis_ood} show an example of the increased interpretability by visualizing in-domain vs out-of-domain input examples.
\item In Section \ref{sec:factorized_rnns} we develop a factorized RNN by starting from the well-known vanilla RNN and then replacing its MLP building block with our PFC blocks. This results in an RNN modeled as a single matrix factorization, effectively extending the parts-based nature of NMF to the modeling of sequential data. In Section \ref{sec:standard_nmf_rnn} we demonstrate its interpretability advantages compared to the standard vanilla RNN on a simple sequential learning task with a known interpretable solution and observe that it is consistently able to learn the minimal transition model of the solution, even when the network is heavily over parameterized. In all of our sequence learning tasks we find that the factorized RNN performs either competitively or better compared with the vanilla RNN when both are trained with the usual BPTT. This includes learning a repeating sequence (Section \ref{sec:standard_nmf_rnn}), Copy Task (Section \ref{sec:copy_task}), Sequential MNIST (Section \ref{sec:seq_mnist}), and audio source separation (Section \ref{sec:audio_separation}). We also perform some ablations and find two unexpected results of particular interest: The factorized RNN is able to solve simple tasks such as learning a repeating sequence and the Copy Task using only alternating NMF update rules so that backpropagation is not used at all. We also observe that in both the factorized and conventional RNNs, using backpropagation but disabling BPTT often results in a surprisingly minimal degradation in accuracy. The models do tend to become significantly less parameter efficient without BPTT, however.
\item In Section \ref{sec:split_mnist} we show that our non-replay-based continual learning method is competitive with approaches that rely on replay on the split MNIST Class-IL scenario \cite{vandeven2019scenarios}. Our PFC-based model performs better than the MLP on this task and we introduce a sliding window optimizer in Section \ref{sec:slw_optimizer} for PFC-based models that results in further accuracy improvements.
\item In Section \ref{sec:non_iid_training} we show the superiority of a PFC-based model over the MLP on a non-i.i.d. training task.
\item In Section \ref{sec:knowledge_removal} we show how a PFC-based model can support knowledge removal after training by leveraging the sliding window optimizer that we introduce in Section \ref{sec:slw_optimizer}.

\end{itemize} 

We also mention the following limitations:
\begin{itemize}
  \item The PFC block is slower and consumes more memory during training than the MLP block due to the use of an iterative inference algorithm that effectively turns each replaced MLP into a corresponding RNN. It is therefore not intended to be a general MLP replacement, but rather it is intended to be used in modeling tasks where its better interpretability, parts-based modeling prior, and/or suitability to continual learning is needed.
  \item Since the PFC block is modeled as a matrix factorization, it similarly cannot discriminate between two inputs that differ only by a scale factor, since the corresponding outputs would also only differ by the same scale factor. The NMF modeling assumption also constrains the input features to being non-negative, although this can potentially be relaxed to semi-NMF to support negative data values at the expensive of possibly reduced interpretabiity.
  \item This work is preliminary. Our experiments only involve relatively small datasets. In terms of the suitability of the PFC block as an MLP replacement, we only consider two multi-block architectures in this paper: a 2-block residual network and a factorized RNN. Evaluating on larger datasets and/or more complex multi-block architectures is left as future research. 
\end{itemize} 

\section{Matrix-factorization-based layers}

We can motivate our approach by first reviewing some related methods that will contribute toward its design. These include non-negative matrix factorization (NMF) in Section \ref{sec:nmf_review}, k-nearest-neighbors (k-NN) prediction in Section \ref{sec:knn_review}, and the multi-layer perceptron (MLP) in Section \ref{sec:mlp_review}. We then present the The Predictive Factorized Coupling (PFC) block in Section \ref{sec:factorized_layers}.

\subsection{Review of non-negative matrix factorization (NMF)}
\label{sec:nmf_review}

Non-negative matrix factorization (NMF) is a matrix decomposition method where a data matrix $V \in \mathbb{R}^{M \text{x} N}$ is factorized into two matrices $W \in \mathbb{R}^{M \text{x} R}$ and $H \in \mathbb{R}^{R \text{x} N}$, with the property that all three matrices have no negative elements. The objective is to find two non-negative matrices whose product closely approximates the initial matrix, such that:

\begin{align}
  V \approx W H
  \label{eqn:nmf_v_eq_w_h}
\end{align}

Here, $R$ is a tunable hyperparameter that leads to a more compressed approximation as it is reduced. NMF was originally proposed by \cite{paatero_1994} as \emph{positive matrix factorization} and later popularized by \cite{Lee_seung}.  NMF's strength lies in its ability to provide interpretable decompositions. In particular, the non-negativity constraint is found to lead to sparse and parts-based representations because only additive, not subtractive, combinations are allowed \cite{Lee_seung}. In many real-world applications, such as image representation or document analysis, this aligns well with the nature of the data where the measured quantities (like pixel intensity or word counts) are inherently non-negative. 

It is also possible to relax the non-negativity constraint so that only the factor matrix $H$ is constrained to be non-negative, which is called semi-NMF \cite{ding2008convex}. This allows for more flexibility in representing data but can also reduce the interpretabiity.

The factorization is not unique and is typically computed using iterative algorithms, such as gradient descent or multiplicative updates, for example \cite{lee2000algorithms}. This involves first specifying a differentiable reconstruction loss function such as mean-squared error (MSE), for example, and then jointly optimizing $W$ and $H$ to minimize the loss. These algorithms alternately fix one of the factor matrices and update the other, until approximate convergence to a fixed point. In this paper we will use the same terminology from \cite{vogel2008positive} to discriminate between these two updates, so that the \emph{NMF left update} step refers to applying the update rule once to update the $W$ factor matrix, which we also refer to as the \emph{NMF learning update} step. Likewise, the \emph{NMF right update} step refers to applying the update rule once to update the $H$ factor matrix, which we also refer to as the \emph{NMF inference update} step.

The usual modeling convention is that the columns of $V$ correspond to $N$ input feature vectors, which are $M$-dimensional. For example, the neural network interpretation of NMF shown in Figure 3 of \cite{Lee_seung} corresponds to:

\begin{align}
  v_n \approx W h_n
  \label{eqn:nvf_v_eq_w_h}
\end{align}

where $v_n$ and $h_n$ are now individual column vectors (at columns index $n$) in $V$ and $H$, respectively. Here, $v_n$ represents the observed input features (visible variables), while $W$ represents the neural network weights and $h_n$ represent the hidden variables which are inferred from $v_n$ via the repeated application of the NMF right update rule until convergence. The columns of $W$ are often referred to as the learned \emph{basis vectors} in the literature and we will also sometimes refer to them as (parts-based) \emph{prototypes} in this paper.

We can see from the basic NMF formulation in Eq. \ref{eqn:nmf_v_eq_w_h} that it attempts to learn an approximation to a supplied data matrix. While this can be suitable for many tasks, it also tends to limit its use to unsupervised learning and tasks where NMF is able to provide sufficient modeling expressiveness. For these reasons, NMF is not typically used for supervised learning or for tasks requiring more expressiveness, such as modeling sequential data, for example.

\subsection{Review of nearest neighbor prediction}
\label{sec:knn_review}

The k-nearest-neighbors (k-NN) algorithm \cite{Cover1967NearestNP} is one of the simplest and most interpretable machine learning methods used for classification and regression. It is considered an instance or prototype-based method since the model stores the entire training dataset and then makes predictions using a similarity measure.

Suppose that we have a set of training example pairs $\{(x_1, y_1), (x_2, y_2), \ldots, (x_N, y_N)\}$ where $x_n$ is the input feature vector and $y_n$ is the corresponding target output value or vector. In the classification setting, $y_n$ would contain the ground truth class label and could be represented either as an integer label or a 1-hot vector. In the regression setting, $y_n$ could more generally be an arbitrary real-valued vector. Since the following formulation could apply to either, we will refer to it as nearest neighbor prediction.

In typical machine learning algorithms, the model adjusts its weights (i.e., parameters) to minimize a loss function that measures the difference between the model's predictions and the actual target values. However, there is no such loss function in nearest neighbors. Rather, the learning procedure is extremely simple as it only consists of storing the training examples in a suitable data structure as they become available. We will still refer to this data structure as the ``weights'' since it contains the learned knowledge extracted from the training set, which in this case is just the training examples themselves. For easier connection to the matrix-factorization-based models that follow in later sections, we will formulate nearest neighbor prediction as a special case of matrix factorization. For each training example $(x_n, y_n)$, we create a corresponding column vector $w_n$ in the weights matrix $W$ as the vertical concatenation of the target $y_n$ on top of input feature $x_n$. Specifically, let $y_n \in \mathbb{R}^C$ (i.e., $C$ class labels) and let $x_n \in \mathbb{R}^M$. We then construct $w_n \in \mathbb{R}^{C+M}$ as:

\begin{align}
  w_n = \begin{bmatrix} y_n\\ x_n \end{bmatrix}
\end{align}

As a result, $W$ will be a $(C + M)$ x $N$ matrix containing all of the training examples as its columns. We will find it useful to split $W$ into an upper ``prediction'' sub-matrix $W_y$ consisting of the first $C$ rows (containing the $y_n$ targets), and a lower ``recognition'' sub-matrix $W_x$ consisting of the last $M$ rows (containing the $x_n$ input features):

\begin{align}
  W = \begin{bmatrix} W_{y}\\ W_{x} \end{bmatrix}
  \label{eqn:split_knn_w}
\end{align}

The nearest neighbor algorithm also requires us to define a suitable distance or similarity metric. For this, we can specify a similarity function $sim(x_q, x_k)$ which computes a similarity score between two supplied vectors based on their Euclidean distance or cosine similarity, for example. 

Given a new input example $x$, we can use the model to perform inference and predict the output (either classification or regression) as follows. In the first step, we use $sim(x, x_k)$ to compute the similarity score of $x$ against each of the columns $x_k$ in $W_x$. The general form of the algorithm finds the k-nearest neighbors (k-NN) but we consider only the 1-nearest neighbor (1-NN) method here for simplicity. Let $m$ denote the index of the best-matching column in $W_x$ (which contains the training example $x_m$). We refer to this step as ``recognition'' since the input $x$ was recognized as its nearest neighbor $x_m$, which represent the model's reconstruction of $x$. The second step then involves selecting the same column of $W_y$ so that the model will output $y_m$ as its prediction. We therefore refer to this as the ``prediction'' step.

We can express the inference solution to the 1-NN in the form of a matrix-vector product that will make the reconstructive-predictive aspect of the model explicit. We introduce a 1-hot vector $h \in \mathbb{R}^N$ having the same dimensionality as the column count in $W$, where only index $m$ is set to 1. This allows us to express the model's output prediction in the form of the following matrix-vector product:

\begin{align}
  y_{pred} = W_y h
  \label{eqn:nn_pred_y}
\end{align}

We can likewise  express the model's input prediction as:

\begin{align}
  x_{pred} = W_x h
  \label{eqn:nn_pred_x}
\end{align}

Using Eq. \ref{eqn:split_knn_w}, these can be combined so that both the input and output predictions are given by a single product:

\begin{align}
  \begin{bmatrix} 
    y_{pred} \\ 
    x_{pred}
    \end{bmatrix} = \begin{bmatrix} W_y\\ W_x \end{bmatrix} h
    \label{eqn:nn_pred_both}
\end{align}

Note that given the (1-hot) inference solution in $h$, it picks out a single column in the weights to use as the predictions for both the output and input. For the more general k-NN algorithm, we could consider computing $h$ by setting all elements to 0 except those for indices corresponding to the k-nearest neighbors, for which we would use the corresponding value of the similarity function and then normalize these $k$ values to sum to 1 in $h$. Several interpretable aspects of the 1-NN prediction algorithm are worth mentioning. For example, it provides confidence estimation by making use of the similarity score. When the model makes a wrong prediction (or when an out-of-distribution example is supplied), we can inspect the reconstructed input $x_{pred}$ since it shows what the input was recognized as. Continual learning and knowledge removal are easily supported as well, since the training examples comprise the columns of $W$ and can therefore easily be added or removed as necessary.

A drawback of the 1-NN predictor is that the output $y_{pred}$ is not a differentiable function of $W$ and the input $x$. This prevents us from connecting it to a loss function such as classification loss in order to learn the weights using backpropagation, and also prevents its use as a building block for more complex architectures. The reliance on a manually specified similarity function can limit the prediction accuracy. In addition, inference can be expensive for large datasets since the number of columns in $W$ grows as the number of training examples. Despite these drawbacks, 1-NN and the more general k-NN can still sometimes produce a good balance of predictive performance and interpretability depending on the dataset. It is also worth mentioning that learnable versions of nearest neighbor prediction exist and are referred to as Learnable Vector Quantization (LVQ) \cite{kohonen1988learning} \cite{kohonen1990improved}. However, we are not aware of a fully differentiable version that would enable its use as a building block for more expressive architectures.

\subsection{Review of MLP-based Neural Networks}
\label{sec:mlp_review}

In contrast to nearest neighbor methods, which are non-parametric, most machine learning algorithms use a set of learnable parameters. In modern neural networks, also known as deep neural networks (DNNs), most of the learnable parameters tend to be contained in the MLP blocks as discussed in the Section \ref{sec:intro}. The versatility of DNNs lies in their ability to use backpropagation \cite{rumelhart1986learning} \cite{lecun1985learning} \cite{werbos1974beyond} for learning parameters that minimize a chosen loss function—ranging from classification to regression losses. This flexibility allows us to take a foundational component like the MLP and scale it to create architectures of varying complexity such as convolutional networks, deep residual networks, recurrent neural networks (RNNs), transformer architectures, etc.. It is this ability to optimize the parameters so as to minimize a desired loss function that appears to contribute to DNNs' superior predictive performance over other more interpretable methods.

The basic MLP corresponds to the sequential connection of two affine transformations (linear layers) with a non-linear activation function in between. The hidden layer activation vector $h$ is expressed as the following (differentiable) function of the parameters and input vector $x$:

\begin{align}
  h = \sigma(W^T_{xh} x + b_h)
  \label{eqn:mlp1}
\end{align}

where the weights matrix $W_{xh}$ and bias vector $b_h$ are the parameters of the first linear layer and $\sigma()$ can be an arbitrary differentiable activation function. Common choices for $\sigma()$ include ReLU, GELU \cite{hendrycks2016gaussian}, and tanh, for example. The MLP's predicted output, $y_{\text{pred}}$, is then obtained by applying a second affine transformation to $h$:

\begin{align}
  y_{\text{pred}} = W_{hy} h + b_y
  \label{eqn:mlp2}
\end{align}

where the weights matrix $W_{hy}$ and bias vector $b_y$ are the parameters of the second linear layer. Note that $y_{\text{pred}}$ is differentiable with respect to the parameters of both linear layers, hidden activations, and input. This allows is to be used as a building block to create expressive neural architectures. Also note that unlike a single linear layer or Single Layer Perceptron (SLP), an MLP with at least one hidden layer can approximate complex non-linear functions \cite{cybenko1989approximation} \cite{hornik1989multilayer}. This ability comes from the non-linear activation function $\sigma()$ used in the hidden layer. As discussed in Section \ref{sec:intro}, a drawback of MLP-based models is that they are considered black-box models lacking interpretabiity and have challenges dealing with continual learning and training on non-i.i.d. examples.

Although the above presentation of MLPs might make them seem completely unrelated to nearest neighbor methods, we can make a connection between them. We do this by showing that a particular choice of weights initialization and activation function choice for the MLP makes it equivalent to the k-NN predictor. This is the reason why we overloaded $h$ to refer to both the MLP hidden activations in this section, while also using it to refer to the 1-hot vector (or k-nonzero in the case of k-NN) $h$ in Eqs \ref{eqn:nn_pred_y} \ref{eqn:nn_pred_x} \ref{eqn:nn_pred_both}. To see the connection, we first ignore the MLP bias terms. Instead of using the usual backpropagation procedure to learn the weights, we instead initialize MLP weights $W_{xh}$ and $W_{hy}$ to the corresponding nearest neighbor weights $W_x$ and $W_y$ that were used in Eqs \ref{eqn:nn_pred_y} \ref{eqn:nn_pred_x} \ref{eqn:nn_pred_both} and also normalize the columns of $W_{xh}$ to have unit L2 norm. Recall that $W_x$ contained all of the training input vectors $x_n$ as its columns while $W_y$ contained the corresponding target output vectors $y_n$ as its columns, for a total of $N$ columns in each weight matrix. For the activation function $\sigma()$, we use a k-max activation followed by softmax, which has the effect of only passing the k-largest values and then normalizing them to have unit sum. For example, with $k=1$, the output $h$ of $\sigma()$ will be a 1-hot vector similar to the $h$ vector that we used for nearest neighbors. The final step is that we need to ensure that the input $x$ to the MLP is normalized to have unit L2 norm. Note that with this setup, the matrix product in Eq. \ref{eqn:mlp1} computes the cosine similarity between each of the training inputs in $W_{xh}$ and $x$, corresponding to this choice of similarity measure in nearest neighbors. The activation function then has non-zero outputs only for the indices corresponding to the k-nearest neighbors in $W_{xh}$, where these values must sum to 1 in $h$. As a result, we see from Eq. \ref{eqn:mlp2} that the MLP then outputs a linear combination of the corresponding $k$ columns in $W_{hy}$, (which contain the target training vectors $y_n$). Without using backpropagation, this ``nearest-neighbor" MLP remains interpretable. However, if we attempt to train it, the interpretation of the weights as containing prototypes is potentially lost. For example, with the usual backpropagation training, there is no input reconstruction loss used by default, and so the input prediction $x_{pred} = W_{xh} h$ (analogous to Eq. \ref{eqn:nn_pred_x}) may not necessarily result in a good reconstruction.

\subsection{The Predictive Factorized Coupling (PFC) block}
\label{sec:factorized_layers}

In this section we introduce the Predictive Factorized Coupling (PFC) block and discuss how it relates to the models in the previous background sections. Having covered the related models, we can now easily introduce the PFC block simply by re-interpreting the matrix product shown in Eq. \ref{eqn:nn_pred_both} with the NMF modeling interpretation of Eq. \ref{eqn:nvf_v_eq_w_h} instead of k-NN, so that it represents the predicted input and output vectors after using an NMF algorithm to solve for $h$:

\begin{align}
  v = \begin{bmatrix} 
    y_{target} \\ 
    x
    \end{bmatrix} \approx \begin{bmatrix} 
    y_{pred} \\ 
    x_{pred}
    \end{bmatrix} = \begin{bmatrix} W_y\\ W_x \end{bmatrix} h
    \label{eqn:nn_pred_both_pfc}
\end{align}

Interpreted as NMF, it shows the prediction for a single column vector $v$ of data matrix $V$ in the factorization $V \approx W H$, where $V$ contains the input vectors $x_n$ to be recognized as the columns of its lower sub-matrix $X$ and the corresponding target output vectors ${y_{target}}_n$ to be predicted as the columns of its upper sub-matrix $Y_{targets}$:

\begin{align}
  V = \begin{bmatrix} Y_{targets}\\ X \end{bmatrix}
  \label{eqn:split_v_data}
\end{align}

When used as a neural building block, we consider $X$ to contain the input vectors to the PFC block. The block then infers (solves for) $H$ and predicts $Y_{pred}$ for its output. Similar to neural network training, the corresponding targets $Y_{targets}$ are not available during the prediction (i.e., inference) process. $V$ is therefore partially observed during inference, since only its $X$ sub-matrix is available to the block as input. $Y_{targets}$ is then only used for the purpose of computing the prediction loss when it is available.

Recall that under the NMF modeling constraints, the three matrices of $V \approx W H$ are only required to be non-negative (the non-negativity constraints for $V$ and $W$ can also potentially be removed if we allow semi-NMF, but we will assume NMF here for the purpose of describing the model). We see that $H$ is now less constrained compared to the nearest neighbor model that required $h$ to be 1-hot for 1-NN or contain only $k$ non-zero elements for the k-NN case. Since $W$ is learnable under NMF, we no longer need to use the same number of column vectors (which are often called \emph{basis vectors} in the NMF literature) as training examples. Similar to Section \ref{sec:nmf_review}, we let the hyperparameter $R$ specify the number of learnable basis vectors in $W$, and these can now be initialized to random non-negative values as an alternative to initializing with training examples. $R$ also specifies the dimension of the hidden vector $h$. By keeping $h$ internal to the block, we are free to later add or remove basis vectors from $W$ without changing the external interface of the block. With this interpretation, $h$ corresponds to the inferred hidden activations that represent an encoding of the input in terms of the (parts-based) basis vectors in $W$. From Eq. \ref{eqn:nn_pred_both_pfc} we see that $W$ is also composed of two sub-matrices, $W_y$ and $W_x$, which represent the learned basis vectors as coupled ``input-output'' parts or prototypes. These could also be interpreted as learned key-value factors, where the input vector $x$ serves as the query, and the block is seen to perform a kind of factorized attention over parameters in predicting its output.

The factorization expression in Eq. \ref{eqn:nn_pred_both_pfc} corresponding to the PFC block was first proposed in our earlier work \cite{vogel2008positive} where we referred to it as a ``coupling module'' or ``coupling factorization'' and only considered its use in sequential data models. This contrasts with its more general usage as a building block for both sequential and non-sequential models in the current work. Refer to Section \ref{sec:related_work} for a more detailed discussion of related work.

\subsubsection{Training and inference}

We will show that the PFC block's inference procedure is differentiable. This allows it to be used as a general neural building block in arbitrary computation graphs and trained with the usual backpropagation, similar to how existing MLP-based models are trained. From inspection of Eqs. \ref{eqn:nn_pred_both_pfc} \ref{eqn:split_v_data}, it may not initially be clear that the prediction process is differentiable. We see that the block's predicted output, $y_{pred}$, is expressed as:

\begin{align}
  y_{pred} = W_y h
  \label{eqn:block_pred_output}
\end{align}

and so $y_{pred}$ is clearly differentiable with respect to $W_y$ and $h$, but what about with respect to $W_x$ and $x$? Eq. \ref{eqn:nn_pred_both_pfc} is simply a declarative expression stating that the input $x$ is approximately a linear function of the inferred $h$:

\begin{align}
  x \approx W_x h
  \label{eqn:pred_x_from_h}
\end{align}

We need to show that the corresponding reverse direction imperative process of inferring $h$ from $x$ (and from $W_x$) is also differentiable. Letting   $f()$ represent this inference process, we need to show that $h = f(x, W_x)$ is differentiable with respect to $x$ and $W_x$. Recall from Section \ref{sec:nmf_review} that $h$ is computed by an iterative NMF algorithm consisting of a sequence of right-update steps until approximate convergence of $h$ to a fixed point. Let $h_{k+1} = g(h_k, x, W_x)$ denote the function the computes a single right-update step (the subscript $k$ denotes iteration number here, not column index). If it takes $K$ iterations to converge, then $f()$ corresponds to the $K$-fold composition of $g()$ so that we have:

\begin{align}
f = g \circ g \circ \cdots \circ g = g^{(K)}
\end{align}

It then only remains to show that $g()$ is differentiable. There are several options for $g()$ but we will use the simple and well-known SGD update steps which we review in Appendix \ref{sec:sgd_nmf}. Eq. \ref{eqn:sgd_HUpdate_with_gradients_expanded} shows the right-update step for the more general case of (batched) matrix input $X$ rather than vector input $x$, and so we repeat it here for the vector case:

\begin{align}
  h_{k+1} =  relu(h_k - \eta_H W^T (W h_k - x))
  \label{eqn:vector_h_nmf_update}
\end{align}  

The inference learning rate $\eta_H$ controls the step size. With the choice of Eq. \ref{eqn:vector_h_nmf_update} as $g()$, we see that it is indeed differentiable. In summary, the inference procedure given an input $x$ is as follows. We first apply the NMF right-update rule in Eq. \ref{eqn:vector_h_nmf_update} $K$ times (assuming convergence is reached by then) to infer $h$. We then apply the final linear prediction step in Eq. \ref{eqn:block_pred_output} to compute the predicted output. Note that the targets $y_{target}$ in Eq. \ref{eqn:nn_pred_both_pfc} are masked while performing inference, since we are predicting them as $y_{pred}$. For this reason, we will also refer to this extension of NMF as \emph{masked predictive NMF}.

Since NMF is used to infer $h$, we can interpret it as follows. If the input $x$ is well modeled as consisting of a mixture of parts, then the NMF solution can potentially discover those parts (assuming an appropriate learning algorithm for $W_x$). Using NMF terminology, the columns of $W_x$ can be interpreted as the learned parts or ``basis vectors'' so that the inferred $h$ then specifies an additive encoding of the input in terms of the basis vectors. Intuitively, we can interpret \ref{eqn:pred_x_from_h} as expressing that the input $x$ is approximately generated (reconstructed) as a linear function of the inferred $h$. That is, the reconstruction is given as $x_{pred} = W_x h$. Thus, the process of running an optimization algorithm to solve for $h$ corresponds to the model trying to recognize its input in terms of the already-learned ``parts''. When the recognition is successful, the model is able to find an encoding $h$ of the input in terms of these parts that results in a low reconstruction error $e_{reconstruction} = x - x_{pred}$. When it is unsuccessful, $e_{reconstruction}$ will be large. So although we now need to do more computation compared to the MLP, we get a useful new property: the PFC block can now give us feedback through the reconstruction and its error to tell us how well it was able recognize its input.

We note that it is possible to automate the selection of the learning rate and to accelerate the inference procedure so that fewer iterations are required by leveraging a modified version of the Fast Iterative Shrinkage-Thresholding Algorithm (FISTA) \cite{beck2009fast} algorithm that removes the shrinkage step and adapts it to NMF as we describe in Appendix \ref{sec:fista_details}. We used this method of unrolling in all of our experiments.

To learn the weights, we can use the PFC block in an arbitrary computation graph and train with the usual backpropagation algorithm. For example, we could compute a classification or regression loss between $y_{pred}$ and $y_{target}$, perform backpropagation to compute the error gradients, and use an existing optimizer such as SGD, RMSprop \cite{rmsProp}, etc. to update the weights. When using the NMF modeling constraint, we also need to clip any negative values in the weights to zero after each optimizer update. We can also consider training on mini-batches instead of individual examples by replacing $x$ with a matrix $X_n$ containing a batch of examples.

The general idea of unfolding an iterative and differentiable optimization algorithm, such as NMF in our case, into a computation graph and using backpropagation to learn its parameters is not a new idea. It is sometimes referred to as \emph{algorithm unrolling} or \emph{unrolled neural networks} in the literature \cite{monga2021algorithm}. For details on related work, refer to Section \ref{sec:related_work}.

\section{Factorized RNNs}
\label{sec:factorized_rnns}

In this section we develop a simple matrix-factorization-based replacement for the vanilla RNN \cite{elman1990finding}. Our ``factorized'' RNN is modeled as a single matrix factorization of the form $V \approx W H$ without any activation functions, and containing all of the model's input activations, weights, hidden state activations, and output activations. This makes its declarative model one of the conceptually simplest RNNs that we are aware of. We will often be interested in the case where these matrices are constrained to be non-negative to improve interpretability, so that the model will then more specifically correspond to an instance of non-negative matrix factorization (NMF). Since the model \emph{is} NMF, it retains all of the desirable properties of NMF while also more directly supporting the modeling of sequential data. The models we develop in this section are based on the similar factorized recurrent approach used in Section 3 of \cite{vogel2008positive}, although here we use a different recurrent architecture and propose modified and more effective backpropagation-based learning methods using the algorithm unrolling method from Section \ref{sec:factorized_layers}. 

\subsection{Review of the vanilla RNN}
\label{sec:vanilla_rnn_review}

To motivate the idea, we first need to review the vanilla RNN. In the following, we initially assume that the weights ($W$ matrices) have already been learned, so that we only need to consider the \emph{inference} or \emph{forward pass} through the network to compute its output predictions from its inputs. Given an input sequence of length $T$ containing the feature vectors $x_0, x_1, \dots, x_{T-1}$, the RNN maps them in order, one at a time, into a corresponding output sequence of vectors $y_0, y_1, \dots, y_{T-1}$. The subscript denoting the position in the sequence is often called the ``time slice'', even when there is no notion of time involved. The reason they must be mapped one at a time is because the RNN maintains an internal hidden state vector $h_k$ which is consumed as an additional (hidden) input during each time slice, modified, and produced as a (hidden) output for use in the next time slice. So, in time slice $k$, the RNN will consume input $x_k$ and previous hidden state input $h_{k-1}$. It then produces a new hidden state $h_k$ and output $y_k$. The vanilla RNN does this in two stages. In the first stage, we first update the hidden state

\begin{align}
  h_k = \sigma(W^T_{h} h_{k-1} + W^T_{x} x_k + b_h)
  \label{eqn:vanilla_state}
\end{align}

where $\sigma()$ denotes an arbitrary activation function and $b_h$ is the bias vector. In the second stage, we compute the output from the updated hidden state:

\begin{align}
  y_k = W_y h_k + b_y
  \label{eqn:rnn_y_as_linear_layer}
\end{align}

Note that Eq. \ref{eqn:vanilla_state} can be rewritten as a single linear layer followed by nonlinear activation function:

\begin{align}
  h_k =& \sigma(\begin{bmatrix} W^T_{h} & W^T_{x} \end{bmatrix} \begin{bmatrix} h_{k-1}\\ x_k \end{bmatrix} + b_h) \notag \\
  =& \sigma(W_z^T z_k + b_h)
  \label{eqn:h_as_linear_layer}
\end{align}

where we let $W_z$ refer to the combined weights:

\begin{align}
  W_z = \begin{bmatrix} W_{h}\\ W_{x} \end{bmatrix}
\end{align}

and let $z_k$ refer to the combined inputs:

\begin{align}
  z_k = \begin{bmatrix} h_{k-1}\\ x_k \end{bmatrix}
  \label{eqn:z_from_state_and_input}
\end{align}

Combining Eq. \ref{eqn:rnn_y_as_linear_layer} and Eq. \ref{eqn:h_as_linear_layer}, we can express a single time slice of the RNN computation as:

\begin{align}
  y_k = W_y \sigma(W^T_z z_k + b_h) + b_y
  \label{eqn:vanilla_rnn_1_slice}
\end{align}

This shows that each time slice $k$ of the RNN can be interpreted as an MLP that takes input $z_k$ and produces outputs $y_k$. From Eq. \ref{eqn:z_from_state_and_input}, we see that the previous hidden state $h_{k-1}$ appear together with $x_k$ in the input $z_k$, and the updated hidden state $h_k$ corresponds to the hidden layer of the MLP after the $\sigma()$ activation as shown in Eq. \ref{eqn:h_as_linear_layer}. 

\subsection{The factorized RNN}
\label{sec:the_factorized_rnn}

Section \ref{sec:vanilla_rnn_review} showed that the each time slice of the vanilla RNN can be interpreted as an MLP. With this interpretation, it is now interesting to consider the model that results from replacing each of these MLP blocks with a corresponding PFC block. This corresponds to keeping the output linear layer in Eq. \ref{eqn:rnn_y_as_linear_layer} (although with the bias term removed). We replace the input linear layer and activation function from Eq. \ref{eqn:h_as_linear_layer} with the following vector factorization for the $k$'th time slice:

\begin{align}
  z_k \approx W_z h_k
  \label{eqn:factorized_rnn_state}
\end{align}

With this change, we have now reversed the direction of the linear mapping compared to the MLP so that the input $z_k$ is approximately a linear function of the hidden state $h_k$ (contrast this to the MLP in Eq. \ref{eqn:h_as_linear_layer} where the pre-activation $h_k$ is a linear function of $z_k$). Our factorized representation is also simplified compared to Eq. \ref{eqn:h_as_linear_layer} since we have removed the need for an activation function and bias term. The tradeoff is that now when we are given an input $z_k$, we will require an iterative NMF update algorithm to solve for $h_k$, which could be more computationally costly compared to the MLP. 

With the computed $h_k$, we then compute the output as in the vanilla RNN using Eq. \ref{eqn:rnn_y_as_linear_layer}. Applying Eq. \ref{eqn:rnn_y_as_linear_layer} and Eq. \ref{eqn:factorized_rnn_state} to all time slices $k \in 0 \dots T-1$, we finally arrive at the factorized vanilla RNN expressed as a single matrix factorization of the form $V \approx W H$:

\begin{align}
  \begin{bmatrix} 
    y_0 & y_1 & y_2 & \dots & y_{T-1} \\ 
    z_0 & z_1 & z_2 & \dots & z_{T-1}
    \end{bmatrix} \approx \begin{bmatrix} W_y\\ W_z \end{bmatrix} \begin{bmatrix} h_0 & h_1 & h_2 & \dots & h_{T-1} \end{bmatrix}
\end{align}

Using Eq. \ref{eqn:z_from_state_and_input}, we can also express the left matrix $V$ in terms of $y_k$, $h_{k-1}$, and $x_k$. This brings us to the key result which lets us express the the factorized RNN with all of the model inputs, outputs, hidden states, and weights together in a single matrix factorization:

\begin{align}
  \begin{bmatrix} 
    y_0 & y_1 & y_2 & \dots & y_{T-1} \\ 
    h_{-1} & h_0 & h_1 & \dots & h_{T-2} \\
    x_0 & x_1 & x_2 & \dots & x_{T-1}
    \end{bmatrix} \approx \begin{bmatrix} W_y\\ W_h \\ W_x \end{bmatrix} \begin{bmatrix} h_0 & h_1 & h_2 & \dots & h_{T-1} \end{bmatrix}
    \label{eqn:factorized_vanilla_rnn}
\end{align}

Note that for a single time slice, our model corresponds to the following vector factorization:

\begin{align}
  \begin{bmatrix} 
    y_k \\ 
    h_{k-1} \\
    x_k
    \end{bmatrix} \approx \begin{bmatrix} W_y\\ W_h \\ W_x \end{bmatrix} h_k
    \label{eqn:factorized_vanilla_rnn_1_slice}
\end{align}

If we use $W$ to denote the three stacked weights sub-matrices, the notation simplifies even further to the following:

\begin{align}
  \begin{bmatrix} 
    y_k \\ 
    h_{k-1} \\
    x_k
    \end{bmatrix} \approx W h_k
    \label{eqn:factorized_vanilla_rnn_1_slice_simp}
\end{align}

Now contrast the simplicity of the factorized RNN for a single time slice in Eq. \ref{eqn:factorized_vanilla_rnn_1_slice_simp} with the corresponding vanilla RNN expression for a single time slice in Eq. \ref{eqn:vanilla_rnn_1_slice}. Note that they differ in that Eq. \ref{eqn:factorized_vanilla_rnn_1_slice_simp} is a declarative representation while Eq. \ref{eqn:vanilla_rnn_1_slice} is imperative. That is, for the factorized RNN we will still need to find suitable algorithms to actually solve the factorization, whereas the vanilla RNN expression explicitly tells us the steps needed to produce the output.

We can further simplify the notation by replacing each of the sequences in Eq. \ref{eqn:factorized_vanilla_rnn} with their respective sub-matrices. Letting $Y = \begin{bmatrix} y_0 & y_1 & y_2 & \dots & y_{T-1} \end{bmatrix}$, $H_{prev} = \begin{bmatrix} h_{-1} & h_0 & h_1 & \dots & h_{T-2} \end{bmatrix}$, and $X = \begin{bmatrix} x_0 & x_1 & x_2 & \dots & x_{T-1} \end{bmatrix}$ results in the following:

\begin{align}
  \begin{bmatrix} 
    Y \\ 
    H_{prev} \\
    X
    \end{bmatrix} \approx \begin{bmatrix} W_y\\ W_h \\ W_x \end{bmatrix} H
    \label{eqn:factorized_vanilla_rnn_submats}
\end{align}

Regarding the hidden states, we see that the initial ``previous'' $h_{-1}$ only appears in the left $V$ matrix while the final state $h_{T-1}$ only appears in the right $H$ matrix. The other hidden states are duplicated since $h_k$ for  $k \in [0 \dots T-2]$ appear in both $H_{prev}$ and $H$. Also note that $W_h$ must be an $R$ x $R$ square sub-matrix of $W$ since if the $h_k$ are $R$-dimensional then each of $W_y$, $W_h$, and $W_x$ must also have $R$ columns.

\subsection{Training with alternating NMF update rules}
\label{sec:standard_nmf_factorized_rnn}

A simple method of training the factorized RNN consists of performing alternating NMF updates to $W$ and $H$, while also copying the inferred states from $H$ to the corresponding duplicated positions in the data matrix after each inference update step to satisfy the consistency constraints. This is similar to the approach that we used for the sequential models in \cite{vogel2008positive}. The details of this are as follows. We use the simple SGD-based algorithm which is reviewed in Appendix \ref{sec:sgd_nmf} for our experiments, but a variety algorithms can potentially be used.. Starting from the factorization in Eq. \ref{eqn:factorized_vanilla_rnn}, we begin by initializing the weights $W = {W_y, W_h, W_x}$ to small random values and initializing the hidden states to either zeros or small random values. This corresponds to setting sub-matrices $H_{prev}$ and $H$ to zeros or small random values in Eq. \ref{eqn:factorized_vanilla_rnn_submats}. Let $X = \begin{bmatrix} x_0 & x_1 & x_2 & \dots & x_{T-1} \end{bmatrix}$ represent the training inputs and $Y = \begin{bmatrix} y_0 & y_1 & y_2 & \dots & y_{T-1} \end{bmatrix}$ represent the corresponding target output values that we want to predict. Since we want to predict the outputs, using only the provided inputs, we first need to infer the hidden states. For that we use the subpart of Eq. \ref{eqn:factorized_vanilla_rnn} corresponding to Eq. \ref{eqn:factorized_rnn_state}:

\begin{align}
  \begin{bmatrix}
    h_{-1} & h_0 & h_1 & \dots & h_{T-2} \\
    x_0 & x_1 & x_2 & \dots & x_{T-1}
    \end{bmatrix} \approx \begin{bmatrix} W_h \\ W_x \end{bmatrix} \begin{bmatrix} h_0 & h_1 & h_2 & \dots & h_{T-1} \end{bmatrix}
    \label{eqn_factorized_rnn_inf_subpart}
\end{align}

The task is then to solve for the $h_k$, which we can do by alternating between matrix factorization updates of the right $H$ matrix followed by enforcing the constraint that the duplicated $h_k$ must have equal values. We can do this by simply copying the $h_k$ from the $H$ (right factor matrix) to $H_{prev}$ (left data matrix) after each NMF update to $H$. Once the updates converge, we can then use the top-most sub-factorization of Eq. \ref{eqn:factorized_vanilla_rnn} to compute the predicted outputs $\hat{y_k}$ as a linear function of the $h_k$:

\begin{align}
  \begin{bmatrix} 
    \hat{y_0} & \hat{y_1} & \hat{y_2} & \dots & \hat{y_{T-1}}
    \end{bmatrix} = W_Y \begin{bmatrix} h_0 & h_1 & h_2 & \dots & h_{T-1} \end{bmatrix}
\end{align}

Now that the inference part of the algorithm has complete, we perform the learning updates. We replace the predicted outputs with the target outputs above and perform a NMF update on $W_y$. Similarly, we perform an NMF update on $W_h$ and $W_x$ in Eq. \ref{eqn_factorized_rnn_inf_subpart}. We then repeat the above procedure until convergence.

Since the hidden state updates propagate one slice forward for each NMF update, it will take the same number of updates as the sequence length for information from the first slice to potentially reach the last slice. As a result, if the sequence of long, the initial NMF updates late in the sequence could be considered wasted computation since they will be operating on hidden states that only contain information from nearby slices. Whether or not this actually becomes an issue in practice would seem to depend on how far information can actually propagate in an RNN, which is outside the scope of this paper.

As an alternative to performing the ``full batch'' updates as above, we can consider performing the inference ``one time slice at a time''. Specifically, we start at the first slice $k=0$ and wait for the inference procedure to converge on that slice before containing with the next. Since the (output) inferred hidden state $h_0$ has now converged, we then copy it into the duplicated (input) location in the next ($k=1$) slice of $H_{prev}$. We can now increment the current slice to $k=1$ and continue in the same way so that the input $h_k$ states in the current slice of $V$ can always be considered to have already converged. This is similar to how inference is carried out in the vanilla RNN, since both RNNs share the same temporal dependency ordering between the hidden states. Although the slice-at-a-time inference option seems less able to take advantage of parallel hardware, it also seems potentially more efficient when the sequence length is extremely long since we perform the minimum number of iterations needed for convergence on each state before moving on to the next. Both options seem interesting but a detailed empirical comparison of their relative efficiency is outside the scope of this paper. Perhaps a batch-wise version could also be interesting as a topic for future research. We only consider the latter slice-at-a-time option in the experiments in Section \ref{sec:standard_nmf_rnn}.

\subsection{Training by unrolling NMF inference and backpropagation}
\label{sec:unrolled_factorized_rnn}

The training algorithm in Section \ref{sec:standard_nmf_factorized_rnn} used standard MF update rules to learn the model weights. These updates optimize the local reconstruction loss of the data matrix. Depending on the task, we empirically observed that such algorithms can sometimes be sufficient and we demonstrate example of this in the experiments Sections \ref{sec:standard_nmf_rnn} \ref{sec:copy_task}. However, we generally found this method to fail to perform well on more complex tasks. For this reason we use the algorithm unrolling approach as discussed in Section \ref{sec:factorized_layers} for all other experiments.

It is straightforward to apply unrolling to the factorized RNN since the inference algorithm remains unchanged: we evaluate the computation graph and backpropagate through it. Since we replaced each MLP of the vanilla RNN with a corresponding PFC block, the computation graph dependencies result in the inference progressing one time slice at a time, similar to the vanilla RNN. However, note that in performing the inference in a given time slice, the PFC block's unrolled NMF update steps form another RNN (internal to each PFC block) with length corresponding to the number of unrolled iterations. We therefore have a computation graph corresponding to an RNN within an RNN.

We compute the loss using the MSE loss between the targets and the predicted values. Since the factorized RNN has three predicted sub-matrices in $V$ (Eq. \ref{eqn:factorized_vanilla_rnn_submats}), this means we will have three loss terms. Once the inference procedure converges so that the inferred hidden states are available in $H$, we can predict $\hat{V}$ as follows:

\begin{align}
  \begin{bmatrix} 
    \hat{Y} \\ 
    \hat{H_{prev}} \\
    \hat{X}
    \end{bmatrix} = \begin{bmatrix} W_y\\ W_h \\ W_x \end{bmatrix} H
    \label{eqn:factorized_vanilla_rnn_submats_predicted}
\end{align}

The total loss is the the sum of the three loss terms with an arbitrary non-negative scale factor to adjust the relative strength of each term:

\begin{align}
  loss = \lambda_y MSE(Y, \hat{Y}) + \lambda_h MSE(H_{prev}, \hat{H_{prev}}) + \lambda_x MSE(X, \hat{X})
\end{align}

We then backpropagate through the loss to compute the error gradients and use an existing SGD-based optimizer such as RMSprop etc. to update the weights. Since the inference procedure operated one time slice at a time, we can see that the gradients then flow backward through the entire sequence, effectively making it a instance of backpropagation through time (BPTT).

\section{An optimizer for continual learning and non-i.i.d. training}
\label{sec:slw_optimizer}

To motivate our approach, recall that k-NN simply stores the training examples as is and then performs classification by finding the nearest training examples to the current input. Such an instance-based model can easily support continual learning since we simply append the new examples to the model weights as they become available. Knowledge removal after training (unlearning) is also easily supported by simply removing any desired subset of ``bad'' training examples from model weights. In contrast, the knowledge representation in the PFC-based model is more distributed since the SGD optimizer update from any particular training example or batch could potentially result in modifications to any of the weights. Additionally, each optimizer update only slightly modifies the weight values, so that many updates are needed before the weights can be considered fully learned. With this understanding, we can modify the optimizer update so that only a narrow unmasked ``learnable window" of basis vectors in each weights matrix $W_i$ in the model are able to be modified. We can allow the window to slowly sweep through $W_i$ (e.g., from left to right), advancing slightly with each optimizer update, so that each basis vector ideally remains inside the window long enough to be effectively learned, but not so long as to be overwritten if or when the distribution of training examples changes. If we keep track of the mapping from training batch index to window position during training, then we will be able to identify the (small) subset of weights that can potentially contain the corresponding learned knowledge. For example, this would be the case if the ordering of training examples within each each does not change, such as using a fixed random shuffling. If the training distribution changes (e.g., in continual learning and/or non-i.i.d. training), the learned basis vectors will be protected from being overwritten because they are only learnable for limited number of optimizer updates over which we assume the training distribution to be relatively unchanging.

We now introduce a new ``sliding learnable window'' (SLW) optimizer that can be used with PFC-based models to support improved continual learning, training with non-i.i.d. examples, and knowledge removal after training. Specifically, suppose each weights matrix $W_i$ has $R$ columns (basis vectors) in total. The learnable window will have a width of $L$ basis vectors, where $L$ is a tunable hyperparameter and $L \ll R$. We denote the current position of the window by the index $r$ of its left-most column in $W$. When training starts, we initialize the learnable window to consist only of the leftmost $L$ basis vectors of each weight matrix by setting $r = 0$. As training progresses, we then increment $r$ by some small fractional amount, which is specified by the hyperparameter $sweep\_speed$. This results in the learnable window slowly sweeping to the right within its weight matrix. We can see that this will have the effect of leaving all basis vectors to the left of $r$ frozen at their current values. Only the $L$ basis vectors at column indices $[r, r+L)$ receive optimizer updates from the current training batch. Likewise, all basis vectors to the right of the window (i.e., with index $k$ such that $k \ge r + L$) consist of unused weights which have not yet made their way into the learnable region. If the sweep speed is chosen too small, we will end up with many unused basis vectors at the end of training. However, if it is chosen too large then we will run out of weights storage before training can complete, unless we dynamically allocate additional weights storage and concatenate it to the right of $W$ to ensure a continuous supply of unused weights.. If $L$ is chosen too small, any given basis vector might not remain under the learnable window long enough to be fully learned. However, if it is chosen too large then the basis columns could remain learnable too long so that they would then be vulnerable to being overwritten as the training distribution shifts. In the case of non-i.i.d. training or when unlearning capability is required (and assuming the ordering of examples does not change from epoch to epoch), resetting $r=0$ at the beginning of each epoch will ensure that the optimizer update for any given training batch is always mapped to same location $r$ in $W_i$.

\section{Experiments}

In this section we present experimental results demonstrating the use of the PFC block as a replacement for the MLP block. Section \ref{sec:mlp_basline_image_classifier} first establishes baseline accuracy results of an MLP-based classifier. In Section \ref{sec:frp_1_layer} we then replace the MLP with a PFC block and evaluate its accuracy on the same datasets. We also conduct ablation experiments to compare the effect of different modeling constraints such as NMF vs semi-NMF, as well as the effect of either including or disabling the input reconstruction loss term. We compare these MLP and PFC-based models on a continual learning task in Section \ref{sec:split_mnist} and on a non-i.i.d. training task in Section \ref{sec:non_iid_training}. We demonstrate how PFC-based models can support knowledge removal after training in Section \ref{sec:knowledge_removal}. We show another example of interpretabiity by visualizing in-domain vs out-of-domain inputs in Section \ref{sec:vis_ood}. 

A usable MLP replacement must also be capable of supporting architectures with multiple blocks. The remaining experiments consider two simple multi-block architectures. In Section \ref{sec:frp_2_layer} we show results for a 2-block (fully-connected) residual network using PFC blocks instead of MLP blocks and demonstrate that it produces competitive accuracy with corresponding MLP-based models. Replacing the MLP blocks of the vanilla RNN with PFC blocks results in a factorized RNN as introduced in Section \ref{sec:factorized_rnns} and we present results for these sequential architectures in Sections \ref{sec:standard_nmf_rnn} \ref{sec:copy_task} \ref{sec:seq_mnist} \ref{sec:audio_separation}. We also perform the following ablations: We compare either using the default backpropagation (i.e., unrolled NMF inference algorithm) or disabling it and using NMF-based weight update rules similar to \cite{vogel2008positive}. When using backpropagation-based training, we also evaluate the effect of disabling BPTT.

We conduct the image classification experiments using fully-connected models instead of architectures such as convolutional networks that arguably have a more suitable inductive prior. Our accuracy results will therefore be significantly below state of the art on the datasets used. Since the purpose of these experiments is to evaluate the suitability of the PFC block as a more interpretable replacement to the MLP block, we are therefore concerned with comparing the relative accuracy of these two blocks rather than attempting to achieve state of the art results. For the same reason, we compare the relative accuracy of the vanilla RNN against the corresponding factorized RNN rather than using more sophisticated RNN or transformer models on the sequence modeling tasks.

All experiments were carried out using single RTX 4090 GPU. Due to the limited compute, we did little hyperparameter tunning and worked with small datasets. Consequently, it seems possible that further improvements to accuracy and/or efficiency could be obtained with additional hyperparameter tuning and it remains unknown how well these results will scale to larger datasets. We did not perform ablations on the number of iterations required for reliable convergence. It is potentially possible to improve efficiency by dynamically unrolling the inference algorithm only for the number of iterations needed for reasonable convergence based on the current inputs. However, we do not attempt this in these experiments and leave it as future research. Results for the MLP-based models were generally run 3 times, but due to the limited compute, the PFC-based model results are shown for a single run and not averaged unless otherwise mentioned. 

We trained the models using early stopping with an 85\%-15\% train-validation split. For simplicity and convenience, we use the MSE loss everywhere, for both the classification/regression loss and the input reconstruction loss (applicable to PFC-based models only). We used equal weighting between the reconstruction and prediction loss terms for simplicity. We use the RMSprop optimizer \cite{rmsProp} since it is a simple optimizer that we found to perform well with minimal hyperparameter tuning. For the PFC-based models, parameters are initialized to uniform random values in $[0, 1e-2]$ by default, corresponding to the NMF modeling constraint. In some experiments we allow negative parameters, corresponding to the semi-NMF modeling constraint. When negative parameters are allowed, they are initialized to uniform random values in $[-1e-2, 1e-2]$. The inferred values for the $H$ factor matrices are always constrained to be non-negative. We initialize the $H$ values to zeros, but note that it is also an option to initialize them to small random values. Unless otherwise mentioned, we set the learning rate to 3e-4 for the PFC-based models and 1e-4 for the MLP models. The weight decay was set to 1e-4 unless otherwise mentioned. For the MLP experiments, weights were initialized using the default PyTorch \texttt{LinearLayer} initializer and negative parameters were always allowed since attempting to use non-negative weights with MLPs resulted in optimization difficulties. The GELU activation \cite{hendrycks2016gaussian} was used as the MLP hidden layer activation function.

\subsection{MLP baseline for image classification}
\label{sec:mlp_basline_image_classifier}

We first evaluate a simple 1-hidden-layer MLP-based classifier as a baseline model on the MNIST \cite{lecun1998gradient}, Fashion MNIST \cite{xiao2017/online}, and CIFAR10 \cite{krizhevsky2009learning} datasets. For each dataset, we train models for hidden dimensions sizes of 300, 2000, and 5000. The input feature size is equal to the number of image pixels when the image is flattened into a vector. This is $28 x 28 = 784$ for MNIST and Fashion MNIST since they contain 28x28 grayscale images and $32 x 32 x 3 = 3072$ for CIFAR10 since color images are used. The output layer dimension is 10 since all three datasets have 10 class labels. Table \ref{table:mlp_image_classification} shows the accuracy results.

\begin{table}[h]
  \centering
  \caption{Results of the baseline MLP image classifier model (averaged over 5 training runs).}
  \label{table:mlp_image_classification}
  \begin{tabular}{lcc}
      \toprule
      Dataset  & Hidden Dimension & Test Accuracy \\
      \midrule
      MNIST & 300  & 98.01\% \\ 
      MNIST & 2000 & 98.26\% \\
      MNIST & 5000 & 98.32\% \\
      Fashion MNIST & 300  & 88.07\% \\ 
      Fashion MNIST & 2000 & 88.72\% \\
      Fashion MNIST & 5000 & 88.85\% \\
      CIAFAR10 & 300  & 51.59\% \\ 
      CIAFAR10 & 2000 & 52.78\% \\
      CIAFAR10 & 5000 & 53.17\% \\
      \bottomrule
  \end{tabular}
\end{table}

\subsection{PFC network for image classification}
\label{sec:frp_1_layer}

We train a 1-block PFC-based network on the same datasets and with the same parameter sizes as the MLP from Section \ref{sec:mlp_basline_image_classifier}. Recall that the PFC basis vector count corresponds to the MLP hidden dimension. We also train with and without enforcing non-negative parameters (i.e., NMF vs semi-NMF) and also evaluate the effect of including vs disabling the input reconstruction loss term.

We train with early stopping after 20 epochs with no validation loss improvement. Table \ref{table:pfc_1_layer_image_classification} shows the accuracy results, averaged over 3 training runs. We see that using semi-NMF generally leads to slightly better accuracy. The impact of input reconstruction loss on accuracy is less clear and seems to vary depending on the specific dataset and parameter configuration. Since both the NMF constraint and enabling reconstruction loss can potentially lead to better interpretability, we will enable the reconstruction loss in all remaining experiments. We will also use the NMF parameter constraint in the remaining experiments unless otherwise mentioned.

Comparing the PFC results in Table \ref{table:pfc_1_layer_image_classification} with the MLP results in Table \ref{table:mlp_image_classification} shows the PFC-based models to perform competitively. We see that the PFC-based model performs significantly better compared to the MLP on CIFAR10, slightly better on Fashion MNIST, and similarly on MNIST, although the MLP does perform slightly better on MNIST for the case of 300-dimensional hidden dimension when the NMF constraint is used on the PFC-based network.

\begin{table}[h]
  \centering
  \footnotesize 
  \caption{Results of the 1-block PFC-based image classifier model (averaged over 3 training runs).}
  \label{table:pfc_1_layer_image_classification}
  \begin{tabular}{lcccc}
      \toprule
      Dataset  & Basis Vector Count & Parameter Constraints & Input Reconstruction Loss & Test Accuracy \\
      \midrule
      MNIST & 300  & NMF & Yes & 96.83\% \\ %
      MNIST & 300  & NMF & No & 97.54\% \\ %
      MNIST & 300  & Semi-NMF & Yes & 97.51\% \\ %
      MNIST & 300  & Semi-NMF & No & 98.68\% \\ %
      MNIST & 2000 & NMF & Yes & 97.78\% \\ %
      MNIST & 2000 & NMF & No & 98.14\% \\ %
      MNIST & 2000 & Semi-NMF & Yes & 98.47\% \\ %
      MNIST & 2000 & Semi-NMF & No & 98.84\% \\ %
      MNIST & 5000 & NMF & Yes & 98.08\% \\ %
      MNIST & 5000 & NMF & No & 98.27\% \\ %
      MNIST & 5000 & Semi-NMF & Yes & 98.65\% \\ %
      MNIST & 5000 & Semi-NMF & No & 98.88\% \\ %
      \hline
      Fashion MNIST & 300  & NMF & Yes & 88.62\% \\ %
      Fashion MNIST & 300  & NMF & No & 88.24\% \\ %
      Fashion MNIST & 300  & Semi-NMF & Yes & 88.60\% \\ %
      Fashion MNIST & 300  & Semi-NMF & No & 89.77\% \\ %
      Fashion MNIST & 2000 & NMF & Yes & 90.21\% \\ %
      Fashion MNIST & 2000 & NMF & No & 90.02\% \\ %
      Fashion MNIST & 2000 & Semi-NMF & Yes & 90.58\% \\ %
      Fashion MNIST & 2000 & Semi-NMF & No & 90.56\% \\ %
      Fashion MNIST & 5000 & NMF & Yes & 90.67\% \\ %
      Fashion MNIST & 5000 & NMF & No & 90.64\% \\ %
      Fashion MNIST & 5000 & Semi-NMF & Yes & 90.82\% \\ %
      Fashion MNIST & 5000 & Semi-NMF & No & 90.78\% \\ %
      \hline
      CIFAR10 & 300  & NMF & Yes & 53.25\% \\ %
      CIFAR10 & 300  & NMF & No & 54.42\% \\ %
      CIFAR10 & 300  & Semi-NMF & Yes & 51.54\% \\ %
      CIFAR10 & 300  & Semi-NMF & No & 54.32\% \\ %
      CIFAR10 & 2000 & NMF & Yes & 58.69\% \\ %
      CIFAR10 & 2000 & NMF & No & 58.30\% \\ %
      CIFAR10 & 2000 & Semi-NMF & Yes & 57.46\% \\ %
      CIFAR10 & 2000 & Semi-NMF & No & 55.78\% \\ %
      CIFAR10 & 5000 & NMF & Yes & 60.12\% \\ %
      CIFAR10 & 5000 & NMF & No & 59.68\% \\ %
      CIFAR10 & 5000 & Semi-NMF & Yes & 59.69\% \\ %
      CIFAR10 & 5000 & Semi-NMF & No & 57.90\% \\ %
      \bottomrule
  \end{tabular}
\end{table}

\subsection{Continual learning on the Split MNIST task}
\label{sec:split_mnist}

In this experiment we evaluate and compare the performance of the PFC and MLP-based models on the Split MNIST task \cite{zenke2017continual} under the Class-IL scenario as described in \cite{vandeven2019scenarios}. In Split MNIST, the MNIST dataset is split into 5 task partitions, so that each task only contains two digits: Split 0 contains digits 0 and 1, Split 1 contains digits 2 and 3, and so on up to Split 4. The Class-IL scenario is the most difficult of the three considered in \cite{vandeven2019scenarios}, as it requires the model solve the tasks that have appeared so far as well as inferring the task ID. Since the model is not told which of the 5 tasks it needs to solve, it only receives the image pixels as input and needs predict the correct digit label. The model will therefore have 28x28 = 784 inputs corresponding to the image pixels flattened into a vector and it will have 10 outputs corresponding to the digit labels. We train the model on one split at a time, ordered by the split number so that Split 0 will be the first task. Within each task the examples are presented i.i.d. to the model. The validation loss is computed over all tasks seen so far and early stopping is used to end the current task and move on to the next once the best validation loss is achieved.

\subsubsection{Continual learning performance of baseline MLP-based model}

In \cite{vandeven2019scenarios}, the authors found that regularization-based continual learning methods such as EWC \cite{kirkpatrick2017overcoming} fail completely on the Class-IL scenario and that memory replay-based methods were needed in order to achieve acceptable accuracy. Specifically, on the Split MNIST task under Class-IL, they found that both a baseline MLP model and regularization-based methods such as EWC resulted in accuracies in the 19-20\% range (Table 4 of \cite{vandeven2019scenarios}).

We also evaluate an MLP as a baseline model on this task. We experimented with a range of hidden dimension sizes from 300 - 2000, but only report results for a hidden dimension of 1357 since it corresponds to the same parameter size as the PFC-based model used in Section \ref{sec:baseline_factorized_continual_learning}. We use the RMSprop optimizer with weight decay set to 1e-4. When we tried using a learning rate of 1e-4 that worked well in the MLP for other experiments, it resulted in 19-20\% accuracy on the test set, roughly matching the results reported in \cite{vandeven2019scenarios}. However, when we tuned the learning rate to maximize the validation set accuracy, we were surprised to find much better performance when the learning rate was reduced to 2e-6. This resulted in a test set accuracy of 40.11\%. We therefore suspect that the difference in accuracies between our baseline MLP and that reported in \cite{vandeven2019scenarios} could be due to hyperparameter tuning. Regardless, even 40.11\% is a poor result considering that the upper bound on accuracy when the examples of all tasks are combined together and presented i.i.d to the model is 97.94\% \cite{vandeven2019scenarios}. In the following sections, we will attempt to improve from this 40.11\% baseline accuracy without resorting to replay-based methods.

\subsubsection{Continual learning performance of baseline PFC-based model}
\label{sec:baseline_factorized_continual_learning}

We evaluate a one-block PFC-based model using non-negative weights and 1357 basis vectors, resulting in approximately the same parameter sizes as the baseline MLP. We continue to use the RMSprop optimizer with weight decay of 1e-4 and a learning rate of 1e-5 found through hyperparameter search on the validation set. This model achieved 67.07\% accuracy on the test set, which is significantly higher than the baseline MLP but still far below the upper bound accuracy.

\begin{figure}
    \centering
    \begin{subfigure}{0.3\textwidth}
        \includegraphics[width=\textwidth]{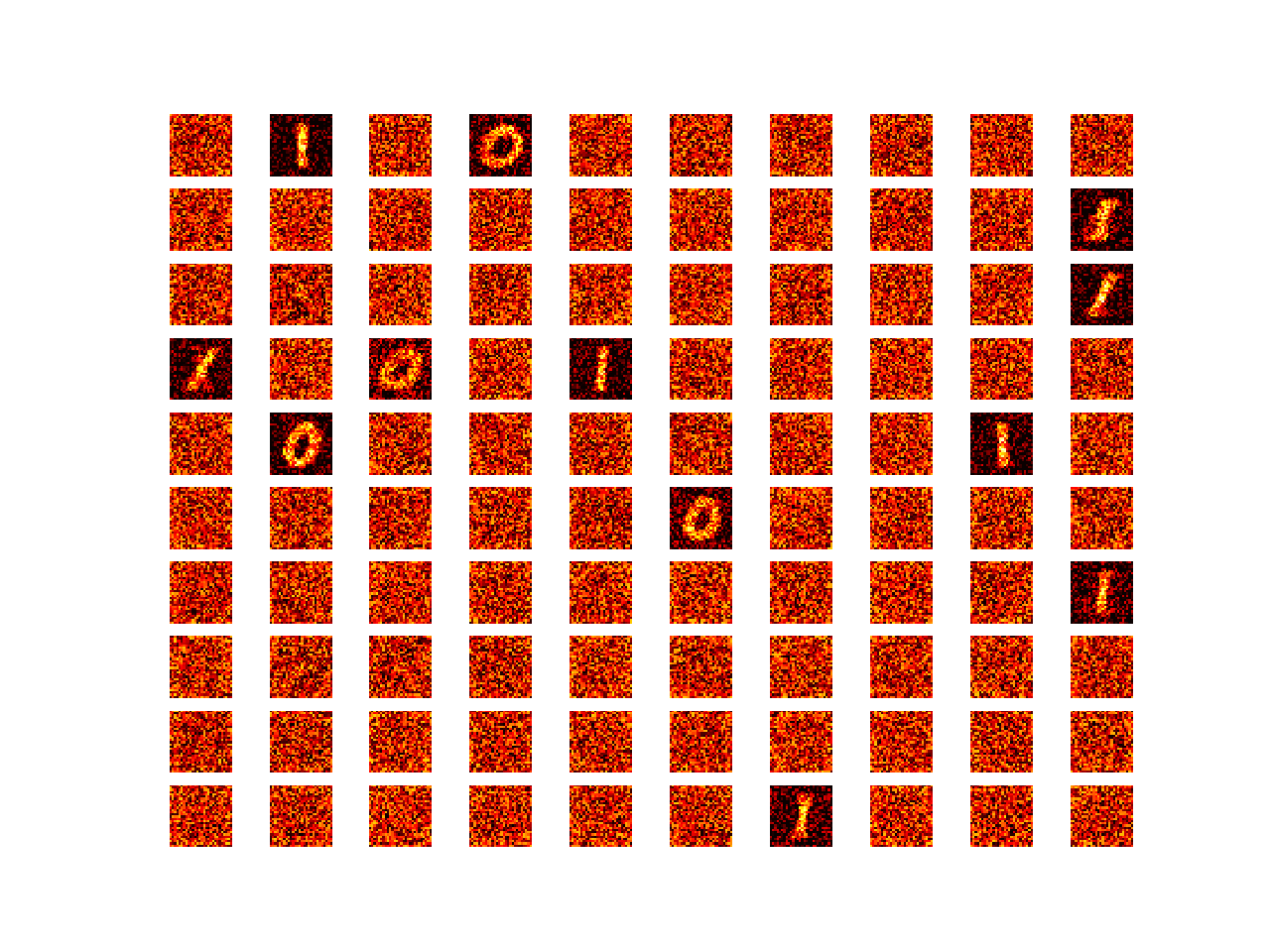}
        \caption{Split 1/5: digits 0 and 1}
        \label{fig:factorized_weights_spit_task_a}
    \end{subfigure}
    \begin{subfigure}{0.3\textwidth}
        \includegraphics[width=\textwidth]{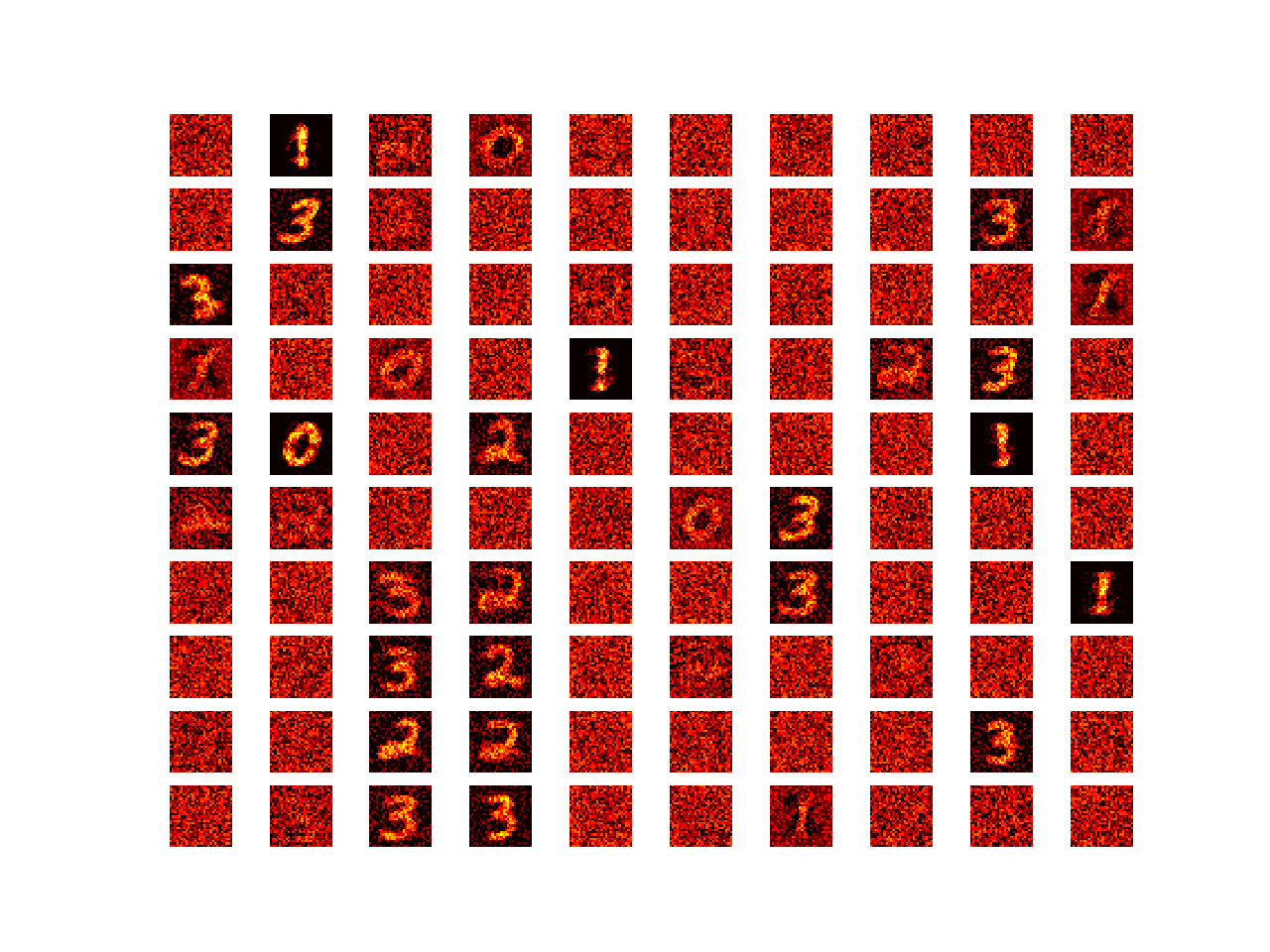}
        \caption{Split 2/5: digits 2 and 3}
    \end{subfigure}
    \begin{subfigure}{0.3\textwidth}
        \includegraphics[width=\textwidth]{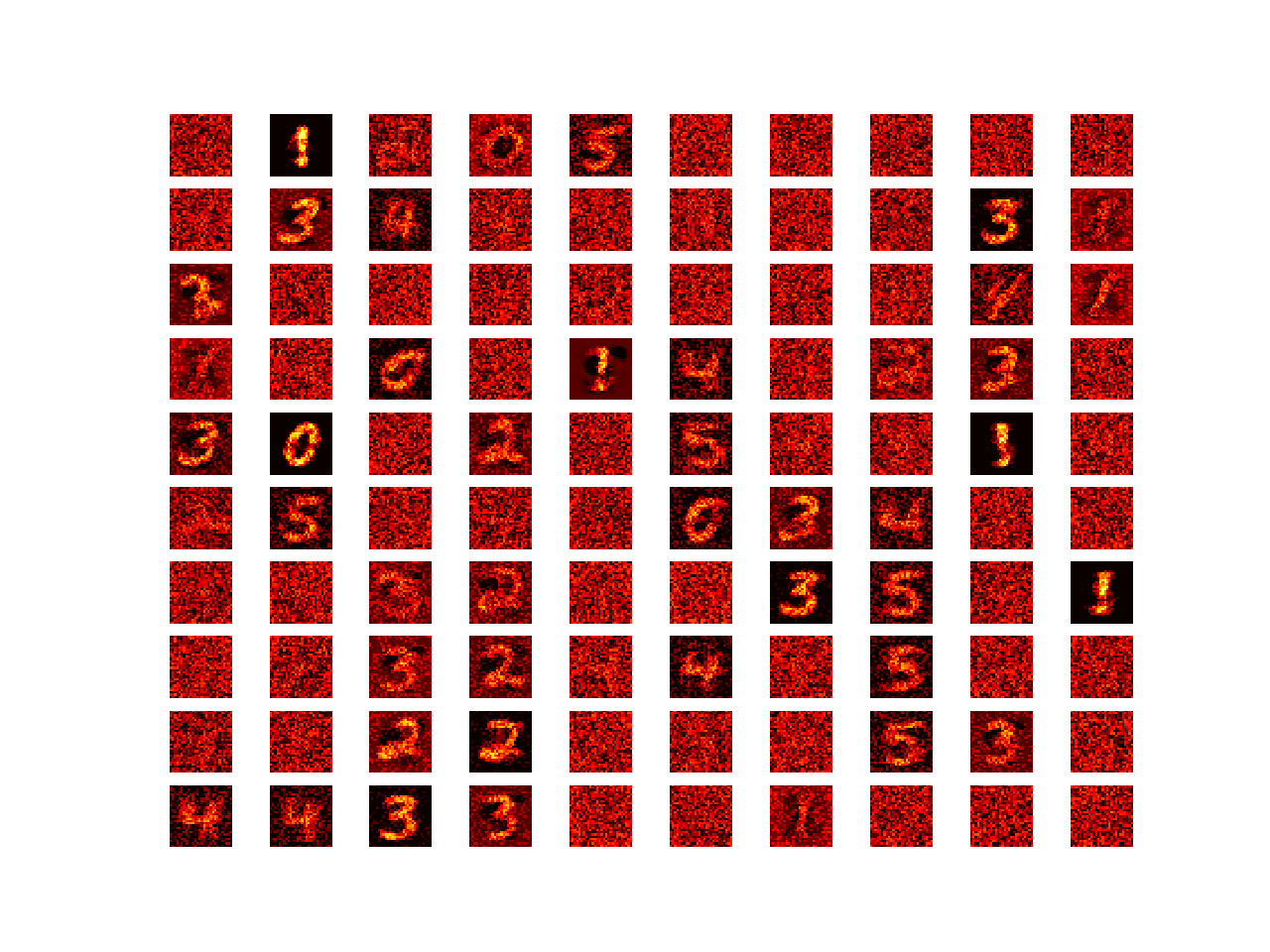}
        \caption{Split 3/5: digits 4 and 5}
    \end{subfigure}
    \\
    \begin{subfigure}{0.3\textwidth}
        \includegraphics[width=\textwidth]{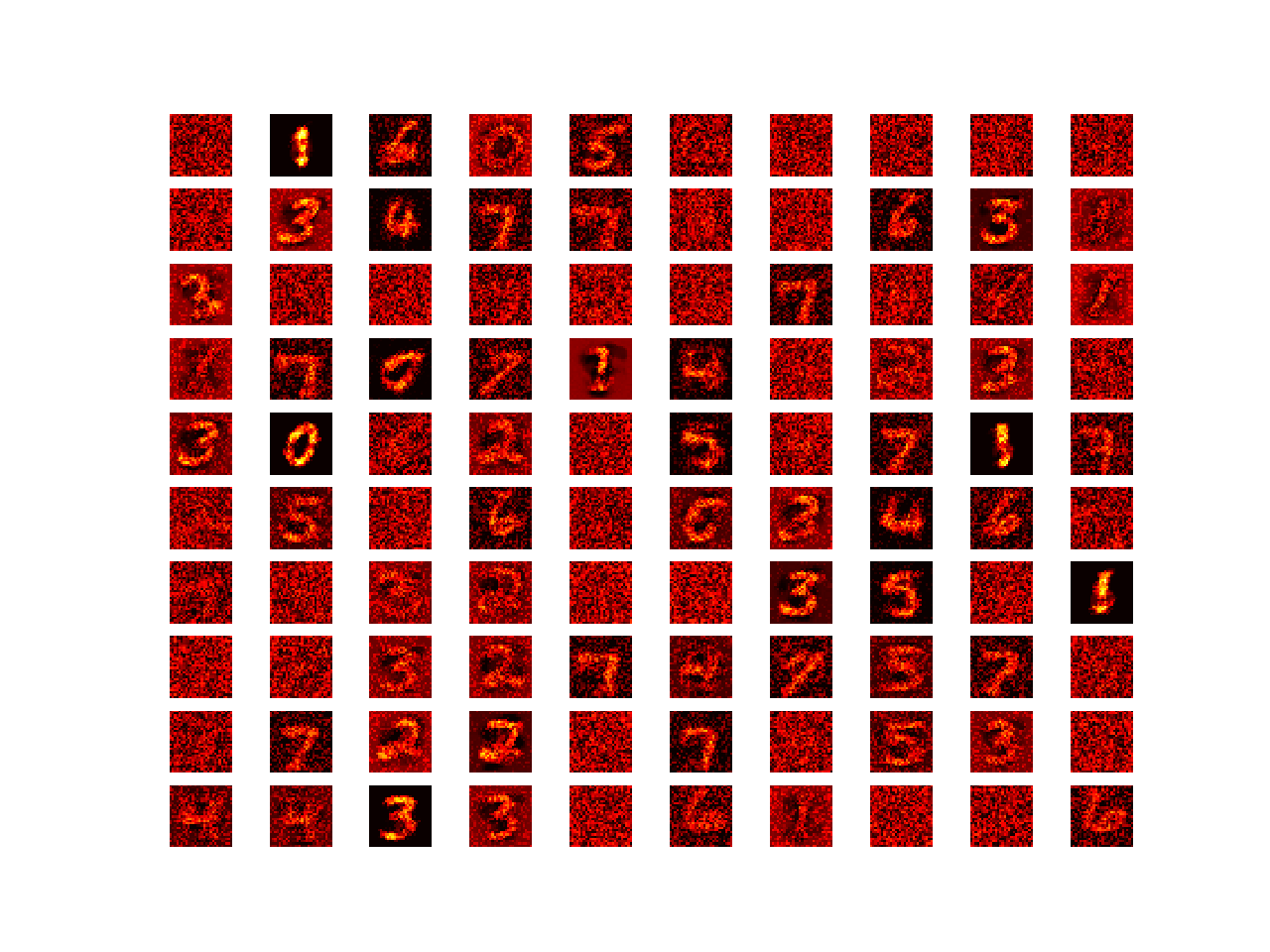}
        \caption{Split 4/5: digits 6 and 7}
    \end{subfigure}
    \begin{subfigure}{0.3\textwidth}
        \includegraphics[width=\textwidth]{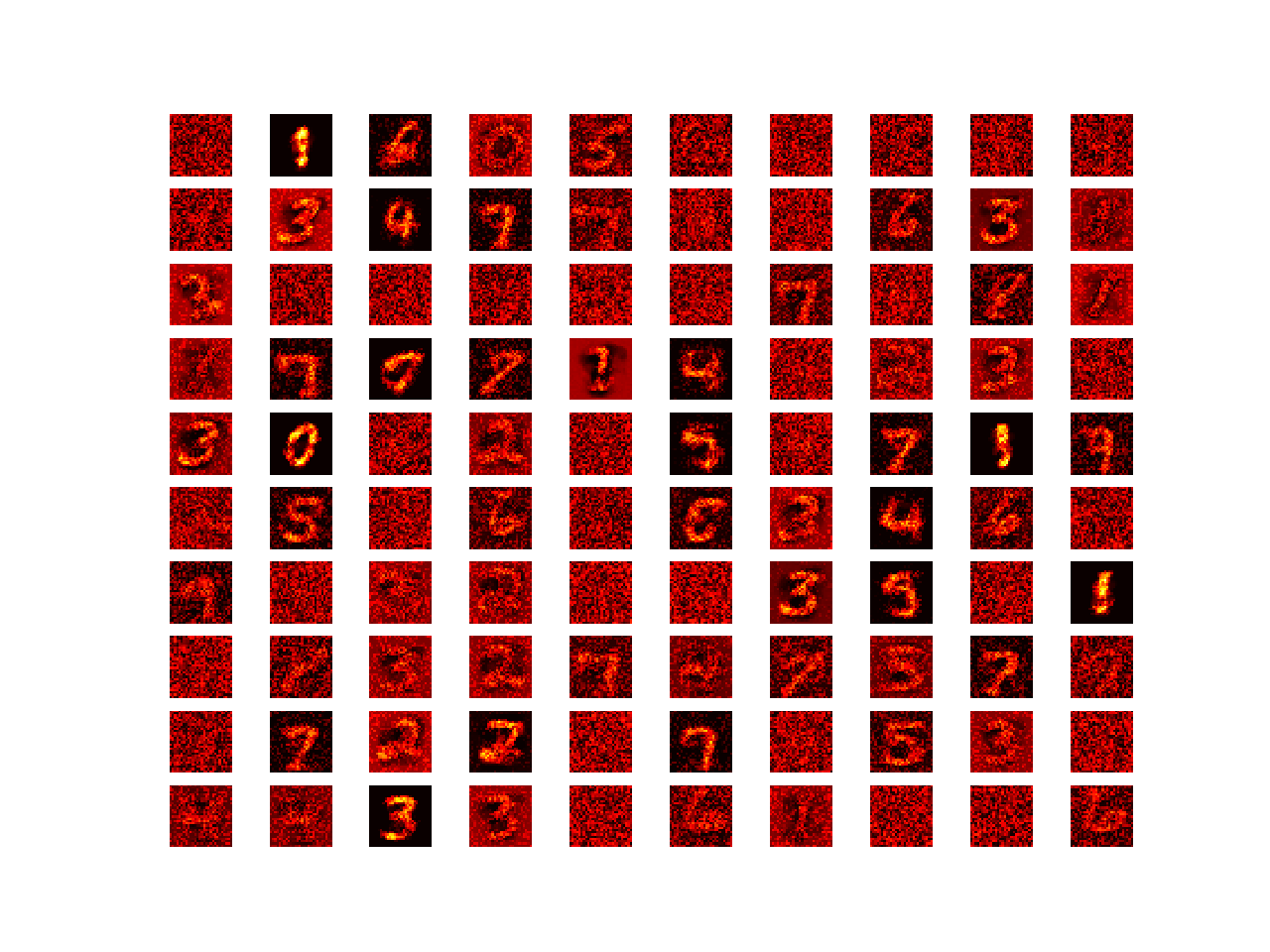}
        \caption{Split 5/5: digits 8 and 9}
        \label{fig:factorized_weights_spit_task_e}
    \end{subfigure}
    \caption{Reconstruction weights of the PFC-based model after learning to classify two MNIST digits in each of the 5 consecutive split tasks. The first 100 weight basis vectors are shown reshaped as images. Notice that two new digit features appear after training on each split. We see that image features (such as the 0 and 1 digits in the first split) of earlier tasks start to degrade as the model learns additional tasks.}
    \label{fig:factorized_weights_spit_task}
\end{figure}

Since this model uses non-negative weights and can potentially learn parts-based representations, it is interesting to visualize the learned weights after training on each of the five tasks. Perhaps we might then be able to better understand what could be causing the model to gradually forget the earlier learned tasks. Figure \ref{fig:factorized_weights_spit_task} shows the first 100 weight basis vectors, reshaped into images after learning to classify the two MNIST digits in each of the 5 consecutive split tasks. Comparing these images, we immediately notice a problem: some of the ``digit'' images learned in earlier tasks gradually fade away as the model learns newer tasks. For example, after completing the first split task, the model can classify the digits 0 and 1, and so we see that the weights in Figure \ref{fig:factorized_weights_spit_task_a} look like zeros, ones, or noise. The model learns to classify digits 2 and 3 during the next split task and so we see these new digits appear in the weights after this task has completed, as expected. However, notice that some of the original 0 and 1 patterns have started to fade or degrade slightly as well. By the time the model has completed training the final split task (i.e., classifying digits 8 and 9), notice that many of the original 0 and 1 patterns are now quite degraded, although a few of them still seem relatively unaffected. We can also see that digits 8 and 9 are difficult to find after learning the final split task in Figure \ref{fig:factorized_weights_spit_task_e}. It seems this is due to the model only training a single epoch on the final task, which maximized the overall validation loss (which is now computed over all splits). It was apparently a better accuracy tradeoff to have relatively poor performance in classifying digits 8 and 9, rather then learning to classify them well and significantly reduce the performance (through forgetting) on all previous tasks. In the next section, we will introduce a simple method to prevent the earlier learned weights from degrading as new tasks are learned.

\subsubsection{Continual learning performance of PFC-based model with a learnable sliding window optimizer}
\label{sec:factorized_continual_learning_learnable_window}

We have implemented this sliding learnable window  idea from Section \ref{sec:slw_optimizer} in a customized RMSprop optimizer which we will refer to as RMSpropSLW. This optimizer takes two additional hyperparameters compared to the standard one: We must specify the learnable width $L$ of the window, as well as $sweep\_speed$, which specifies the (fractional) number of columns that the window advances to the right on each optimizer update.

\begin{figure}
  \centering
  \begin{subfigure}{0.3\textwidth}
      \includegraphics[width=\textwidth]{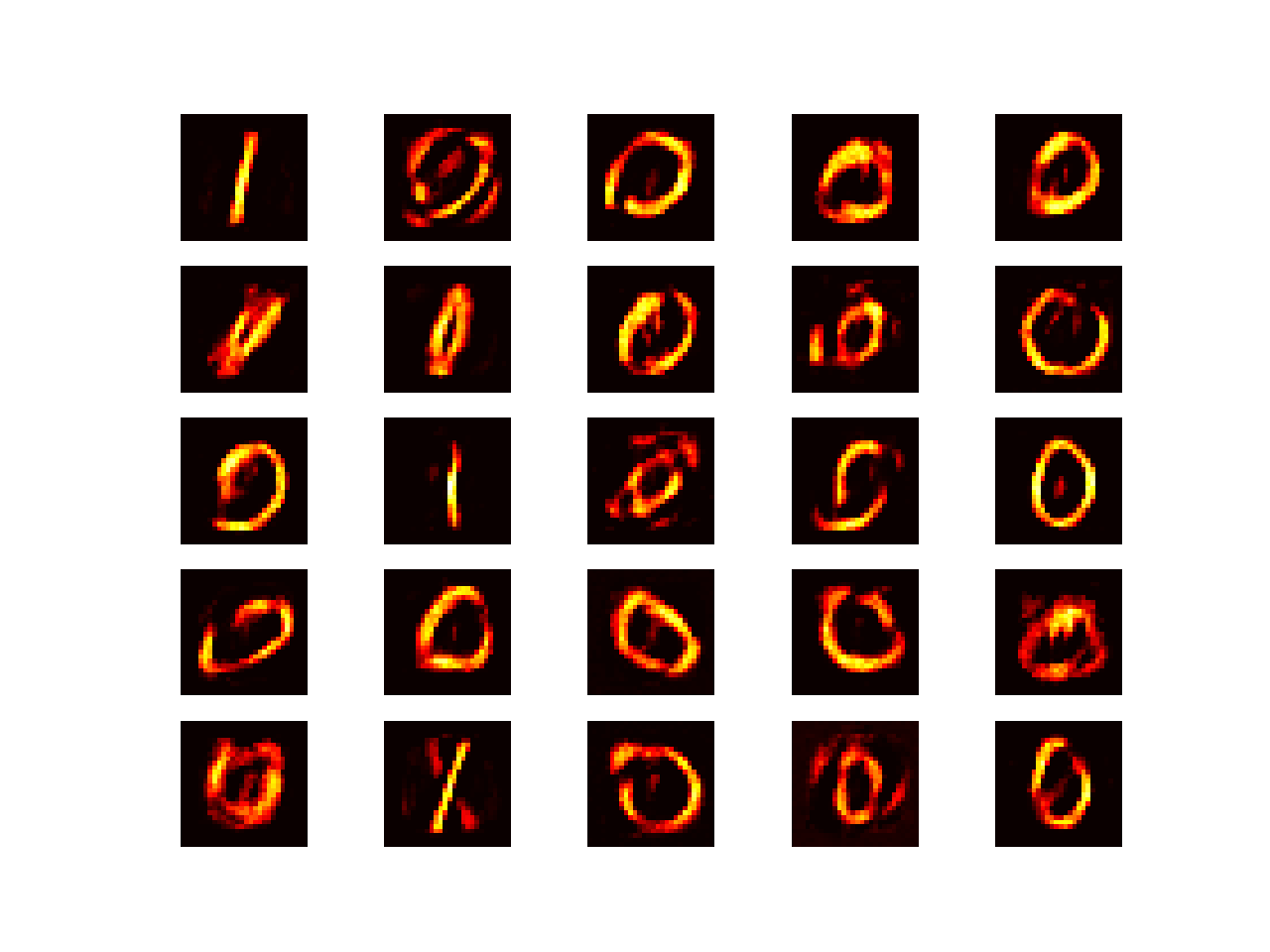}
      \caption{Split 1/5: digits 0 and 1}
      \label{fig:factorized_weights_spit_task_slw_a}
  \end{subfigure}
  \begin{subfigure}{0.3\textwidth}
      \includegraphics[width=\textwidth]{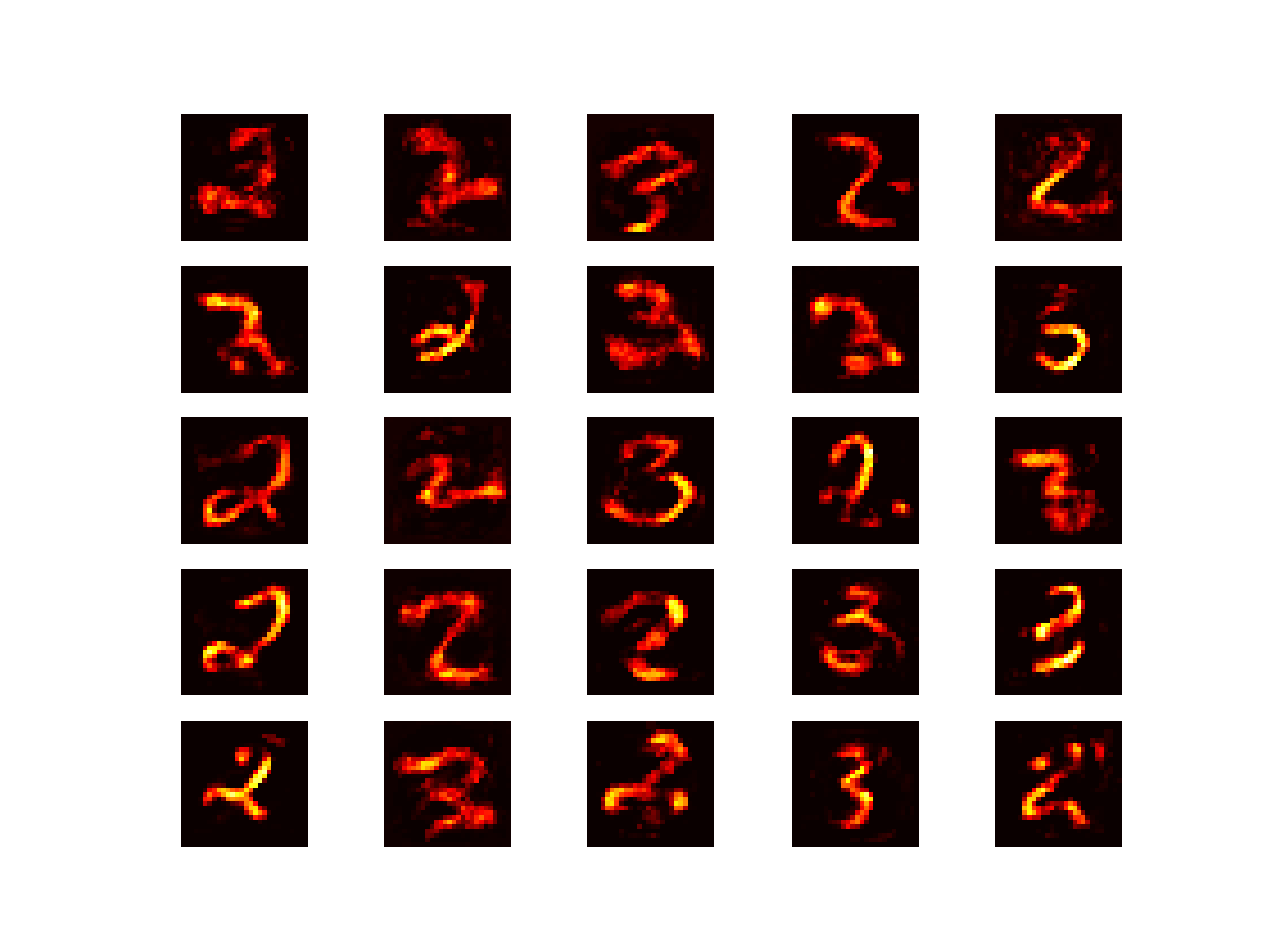}
      \caption{Split 2/5: digits 2 and 3}
  \end{subfigure}
  \begin{subfigure}{0.3\textwidth}
      \includegraphics[width=\textwidth]{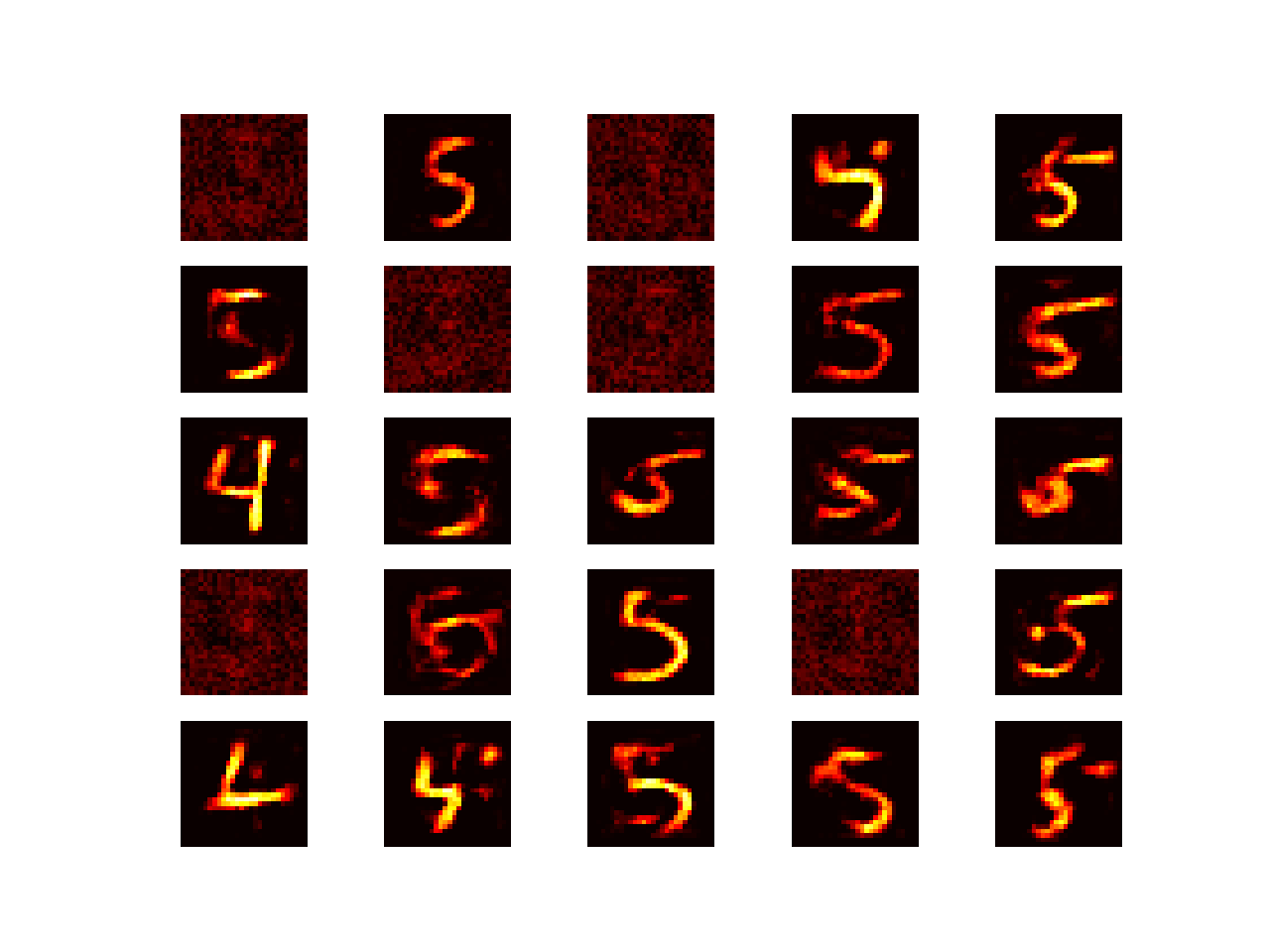}
      \caption{Split 3/5: digits 4 and 5}
      \label{fig:factorized_weights_spit_task_slw_c}
  \end{subfigure}
  \\
  \begin{subfigure}{0.3\textwidth}
      \includegraphics[width=\textwidth]{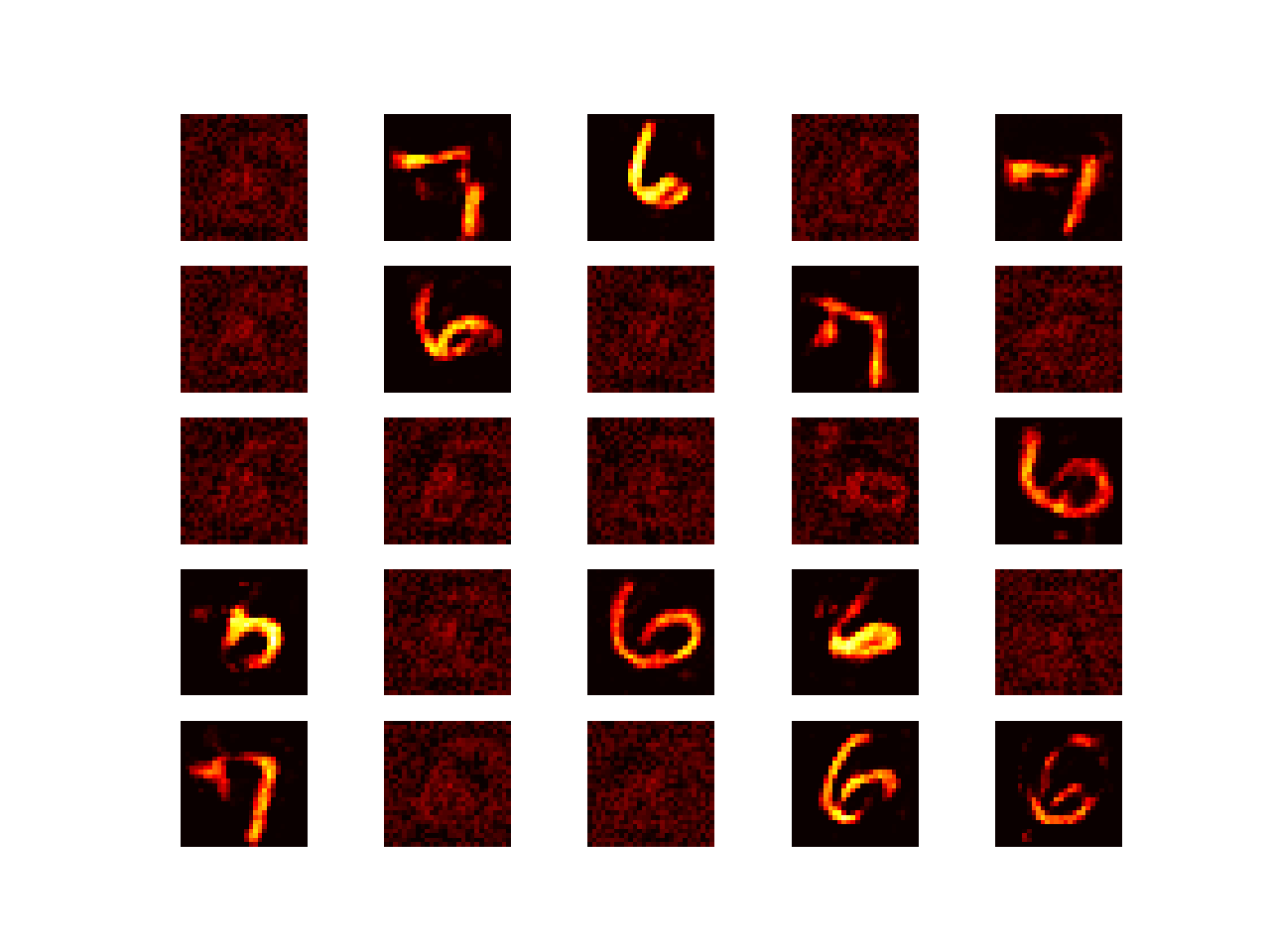}
      \caption{Split 4/5: digits 6 and 7}
  \end{subfigure}
  \begin{subfigure}{0.3\textwidth}
      \includegraphics[width=\textwidth]{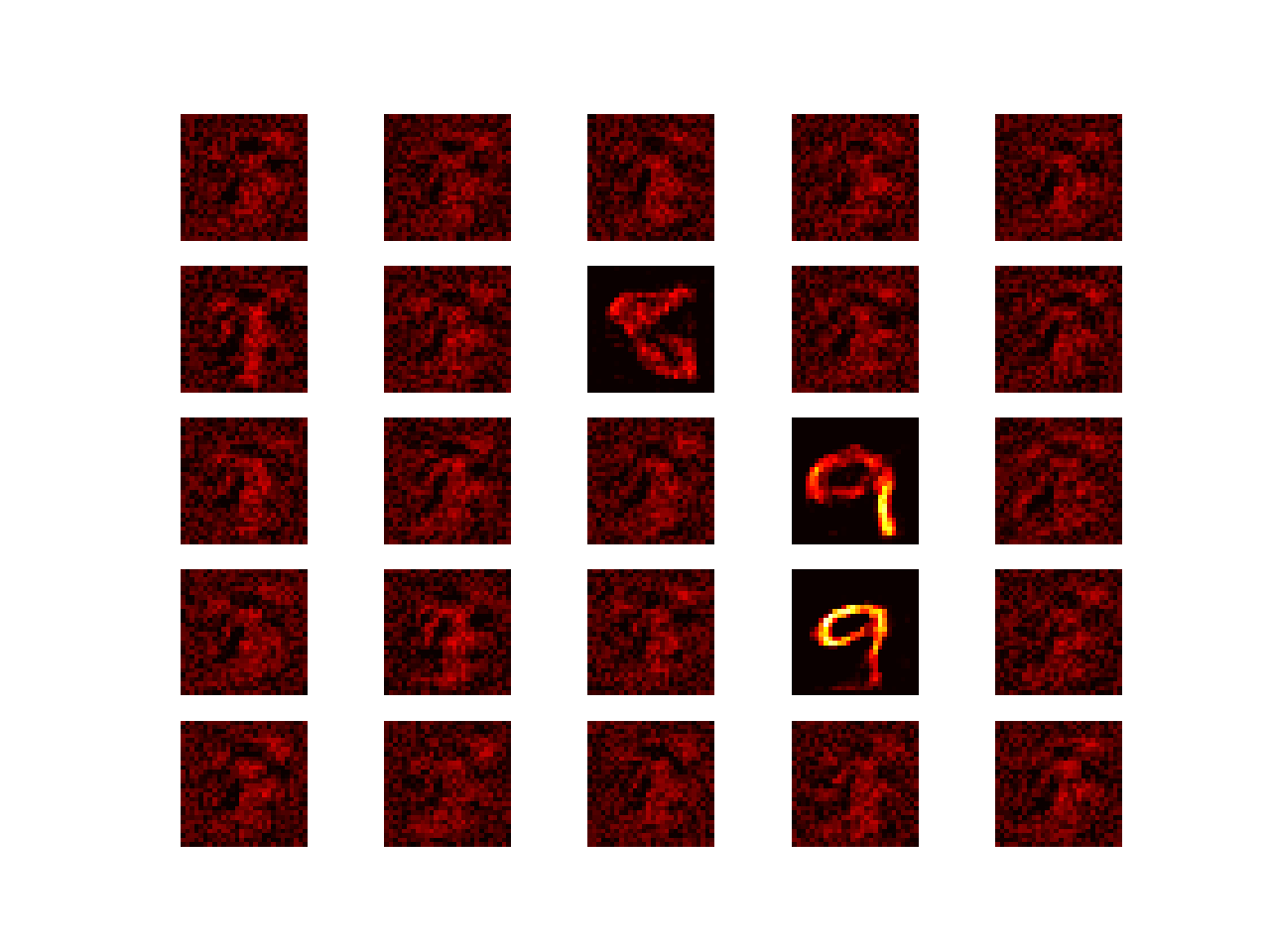}
      \caption{Split 5/5: digits 8 and 9}
      \label{fig:factorized_weights_spit_task_slw_e}
  \end{subfigure}
  \caption{Reconstruction weights of the PFC-based model after learning to classify two MNIST digits in each of the 5 consecutive split tasks, using the sliding learnable window optimizer. The first 25 weight basis vectors are shown, reshaped as 28x28 images. Each sub-figure shows the weights learned during the corresponding split task by extracting the 25 basis vectors to the left of the learnable window just after training of the corresponding task has completed. Since these weights are outside (i.e., to the left of) the sliding learnable window, they remain frozen at their current values and thus protected from degradation during the remainder of training.}
  \label{fig:factorized_weights_spit_task_slw}
\end{figure}

We then repeat the Split MNIST experiment from Section \ref{sec:baseline_factorized_continual_learning}, replacing the RMSprop optimizer with RMSpropSLW. We set the initial (unused) number of basis vectors to 2000 in each of the two weight matrices. We used a sweep speed of 0.25 and a learnable width of 15. We adjusted the slide speed until training was able to complete without using all 2000 basis vectors. Using these values resulted in the right-most column of the learnable window arriving at the 1357'th basis column at the end of training the last task. Note that since early stopping was used to decide when to switch to the next task, the number of in-use basis vectors at the end of training will vary slightly from run to run. Thus, 643 columns remained unused at their initialized random values. This resulted in an accuracy of 93.73\% on the test set, demonstrating the effectiveness of the approach compared to the standard RMSprop optimizer. Figure \ref{fig:factorized_weights_spit_task_slw} shows the weights (reshaped as images) immediately to the left of the learnable window as each of the split tasks is completed. Since these weights are now to left of the sliding learnable window, they remain frozen during the remainder of training. Thus, weights learned during the first split task (i.e., classifying digits 0 and 1) as shown in \ref{fig:factorized_weights_spit_task_slw_a} were protected from being overwritten during later tasks. Notice that the image features visible after training each split only contain the two digits learned during the split. For example, \ref{fig:factorized_weights_spit_task_slw_c} only contains image features for the digits 4 and 5, since the training examples for this split only contain these two digits. Table \ref{table:split_mnist_comparison} summarizes the results on this task for the various approaches discussed. We see that the PFC-based model performs significantly better compared to the MLP even when both use the same RMSprop optimizer. Switching to the RMSpropSLW optimizer further increases the accuracy of the PFC-based model much closer to the upper bound accuracy. Note that neither the RMSpropSLW optimizer nor the model is given any information about which task is active, as required by the Class-IL scenario. Also note that the MLP-based model cannot use the RMSpropSLW optimizer since the modeling assumptions appear to be incompatible.  For reference, replay-based approaches were reported to achieve between 90.79\% and 91.79\% and replay + exemplars were reported to achieve 94.57\% accuracy in Table 4 of \cite{vandeven2019scenarios}. This shows that our non-replay-based PFC + RMSpropSLW approach is somewhat competitive with even replay-based approaches.

\begin{table}[h]
  \centering
  \caption{Comparison of non-replay-based approaches on Split MNIST task under the Class-IL scenario.}
  \label{table:split_mnist_comparison}
  \begin{tabular}{lcc}
      \toprule
      Approach  & Optimizer & Test Accuracy \\
      \midrule
      MLP offline iid training (\cite{vandeven2019scenarios})(upper bound) & RMSprop  & 97.94\% \\ 
      MLP baseline (ours) & RMSprop & 40.11\% \\
      MLP baseline (\cite{vandeven2019scenarios}) & ADAM & 19.90\% \\
      EWC (\cite{vandeven2019scenarios}) & ADAM & 20.01\% \\
      PFC & RMSprop & 67.07\% \\
      PFC & RMSpropSLW & 93.73\% \\
      \bottomrule
  \end{tabular}
\end{table}

\subsection{Non-i.i.d. training: MNIST classification with label-ordered examples}
\label{sec:non_iid_training}

Neural networks are most effectively trained when the stream of training examples is presented i.i.d. and the class labels are balanced. Training difficulties being to arise when distribution shifts are introduced. In this experiment, we train MLP and PFC-based models on the MNIST classification task. However, rather than the usual i.i.d. training in which the examples are presented in shuffled order, we instead present the examples in a fixed label-sorted order. Specifically,  we sort the training examples ascending by class label so that all of the digit 0 examples are presented first, followed by all of the digit 1 examples, and so on, until finally presenting all of the digit 9 examples. We also train the networks with the usual i.i.d. (shuffled) ordering for comparison to provide an upper bound on the achievable accuracy. 

With the PFC-based model, we also have the option of using the sliding learnable window optimizer from Section \ref{sec:slw_optimizer}, which is implemented in RMSpropSLW. Since the examples are presented in the same order each epoch, we simply reset the learnable window to the starting position (i.e., the left-most column of each weight matrix) at the beginning of each epoch. This can potentially improve performance when training on non-i.i.d. data since optimizer updates for a particular example (or its batch) are constrained to a small learnable window of the weights. As a result, only nearby training batches (in terms of the number of training iterations between them) are capable of overwriting previous learned knowledge. More widely spaced batches cannot interfere with each other. 

\subsubsection{Training details}
The batch size was set to 50 for all models. The MLP network has a single hidden layer with a size of 2600. We use the RMSprop optimizer. For for i.i.d. training case, the learning rate is 1e-4. The weight decay is 1e-4. For the label-ordered case, we found a lower learning rate of 5e-7 and no weight decay to give the best validation performance.

For the PFC-based network, we trained networks under the following 4 combinations: using either i.i.d. or label-sorted examples, and using either  RMSprop or RMSpropSLW optimizers. When using RMSpropSLW, we set the maximum weight basis vector count to 3000, of which 2600 were actually used in training. We used a learning rate of 5e-6 and no weight decay. When using RMSprop, we used 2600 basis vectors. This resulted in the PFC-based networks having the same parameter count as the MLP network. The weights were constrained to be non-negative. We used a learning rate of 2e-3 and weight decay of 1e-4.

\subsubsection{Results}
The results are summarized in Table \ref{table:ordered_mnist_comparison}. We see that the MLP and PFC-based models perform similarly when trained on i.i.d. examples and using the RMSprop optimizer. When the PFC-based model is instead trained on the RMSpropSLW optimizer, we see the accuracy is reduced slightly. This is not surprising since the use of a narrow learnable window constrains only a small subset of the weights from receiving optimizer updates. We see that the MLP accuracy suffers worse degradation when label-sorted training is used, only reaching 84.92\%, compared to the PFC-based model's 88.19\% when both use RMSprop. However, the PFC-based model's accuracy only degrades slightly to 96.06\% when using the RMSpropSLW optimizer. Note that when using RMSpropSLW, the PFC-based model has similar accuracy under both the i.i.d. and label-sorted cases. In summary, the PFC-based model was more robust to non-i.i.d. examples than the MLP when both used the same optimizer. When using the sliding window RMSpropSLW (which is only compatible with the PFC-based model), this robustness was further increased.

\begin{table}[h]
  \centering
  \caption{Non-i.i.d. training results on label-ordered MNIST classification task. Corresponding i.i.d. results are also shown as an accuracy upper bound.}
  \label{table:ordered_mnist_comparison}
  \begin{tabular}{lccc}
      \toprule
      Model  & Training Method & Optimizer & Test Accuracy \\
      \midrule
      MLP &  i.i.d. training (upper bound) & RMSprop  & 97.91\% \\ 
      PFC & i.i.d. training (upper bound) & RMSprop & 97.88\% \\
      PFC & i.i.d. training (upper bound) & RMSpropSLW & 96.02\% \\
      MLP & label-ordered & RMSprop & 84.92\% \\
      PFC & label-ordered & RMSprop & 88.19\% \\
      PFC & label-ordered & RMSpropSLW & 96.06\% \\
      \bottomrule
  \end{tabular}
\end{table}

\subsection{Unlearning: removing knowledge from a trained model}
\label{sec:knowledge_removal}

It is sometimes desireable to remove learned knowledge from a model. For example, if it is discovered that certain training batches contained errors, or if certain knowledge needs to be removed for legal reasons. A large model might take a long time to train, and so the ability to quickly remove specific knowledge could be preferable to retaining from scratch.

In this experiment, we train a network on the MNIST classification task in which some of the training examples have incorrect class labels. We show that this reduces the classification performance, as is expected. Provided that the bad examples/batches can be identified after training has completed, we show that it is possible to perform unlearning by removing the subset of model weights that were influenced by these bad examples during training, restoring much of the lost classification accuracy.

We use a model with 1 PFC block and 3000 available basis columns for each of the two weight matrices. For simplicity, we constrain the corrupted training examples so that they appear a contiguous range of batches. We use the same fixed shuffling of examples each epoch so that the i'th batch of each epoch always contains the same examples. There are then a total of 1020 batches per epoch (with 50 examples per batch). Batches 500 through 800 are corrupted by changing the class label to different (and incorrect) class. This is accomplished by incrementing the label modulo the number of classes.

We use the same `RMSpropOptimizerSlidingWindow` optimizer from the continual learning experiments. We reset the sliding window to the left-most position at the beginning of each epoch and use a deterministic dataset loader so that the i'th batch always contains the same examples in each epoch. This will cause the learning updates corresponding to the i'th batch to be stored within a knowable range of columns in the weights matrices, corresponding to the position of the sliding window while the said batch was active.

\begin{figure}
  \centering
  \begin{subfigure}[b]{0.45\textwidth}
    \begin{minipage}[b]{\textwidth}
      \includegraphics[width=\textwidth]{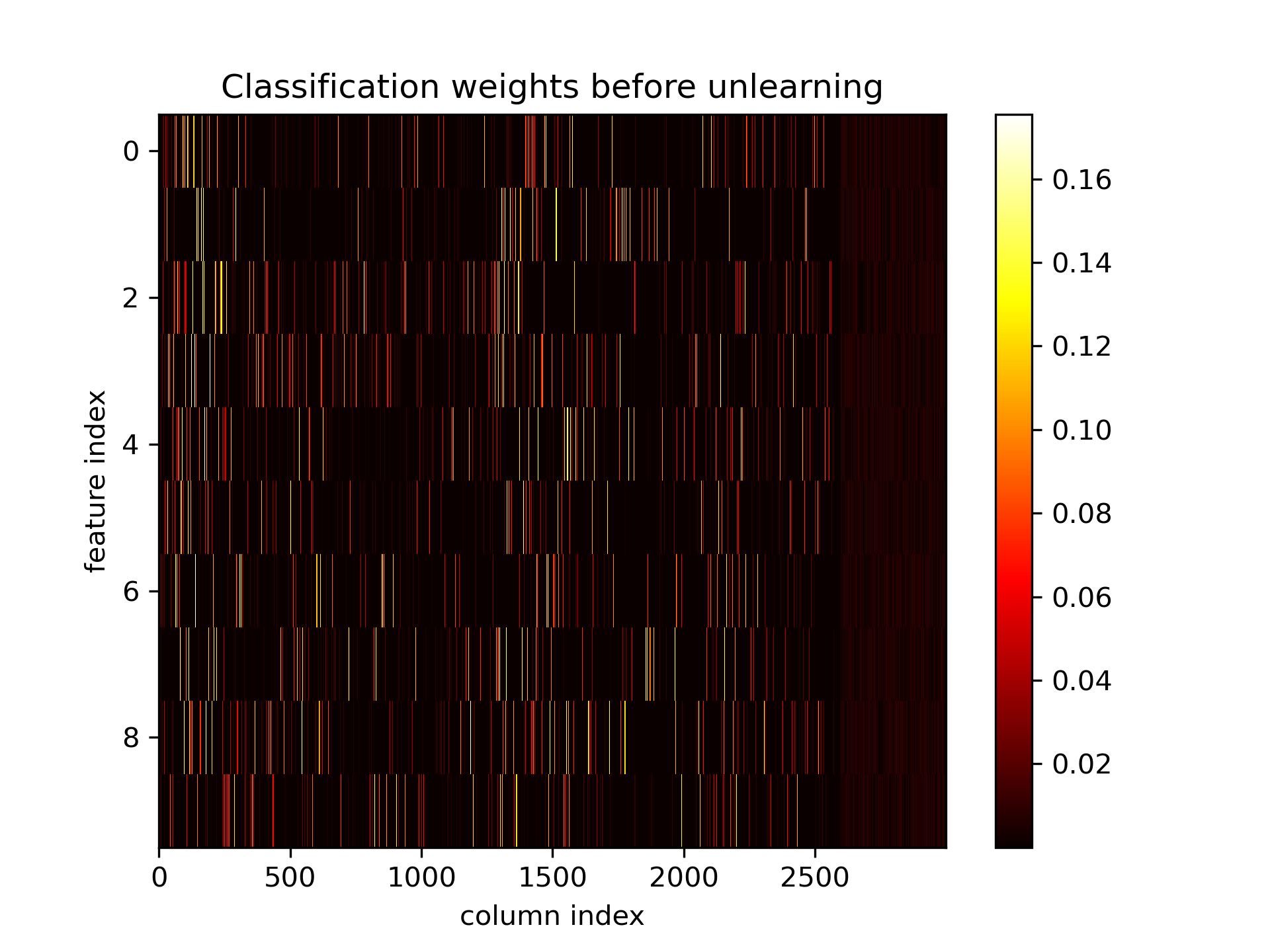}
    \end{minipage}\\ 
    \begin{minipage}[b]{\textwidth}
      \includegraphics[width=\textwidth]{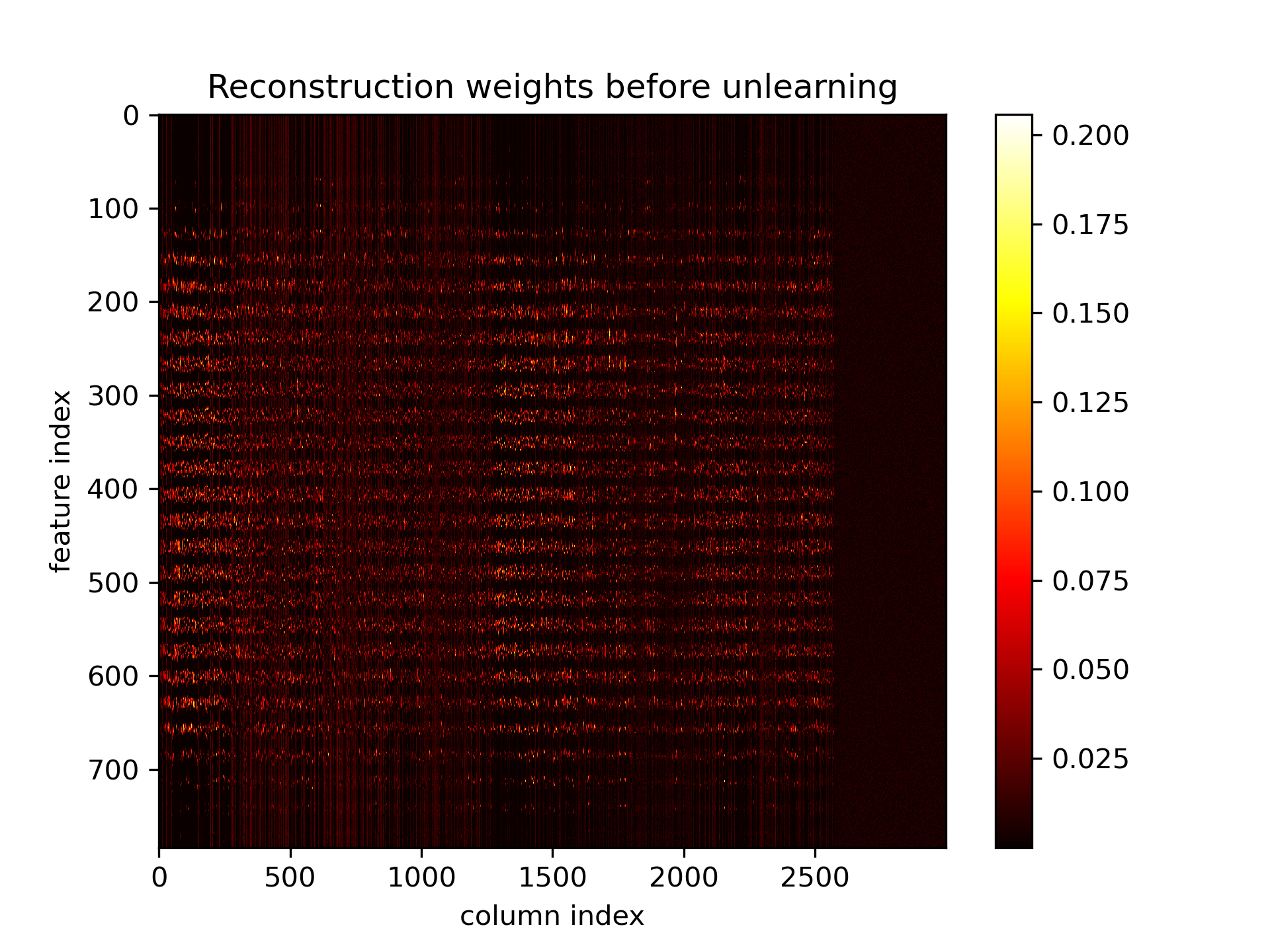}
    \end{minipage}
    \caption{Weights before unlearning}
    \label{fig:unlearning_before}
  \end{subfigure}
  \hfill 
  \begin{subfigure}[b]{0.45\textwidth}
    \begin{minipage}[b]{\textwidth}
      \includegraphics[width=\textwidth]{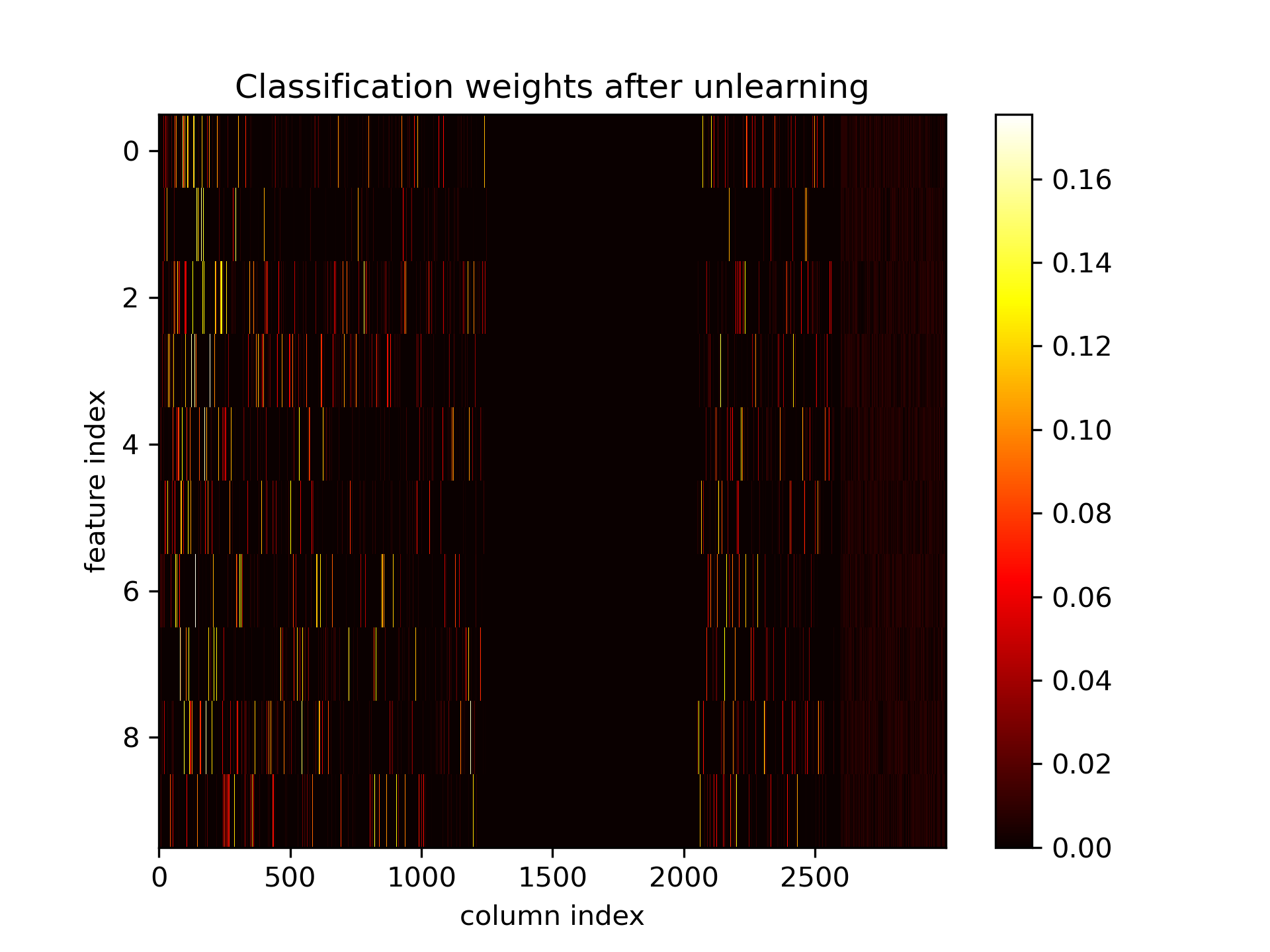}
    \end{minipage}\\ 
    \begin{minipage}[b]{\textwidth}
      \includegraphics[width=\textwidth]{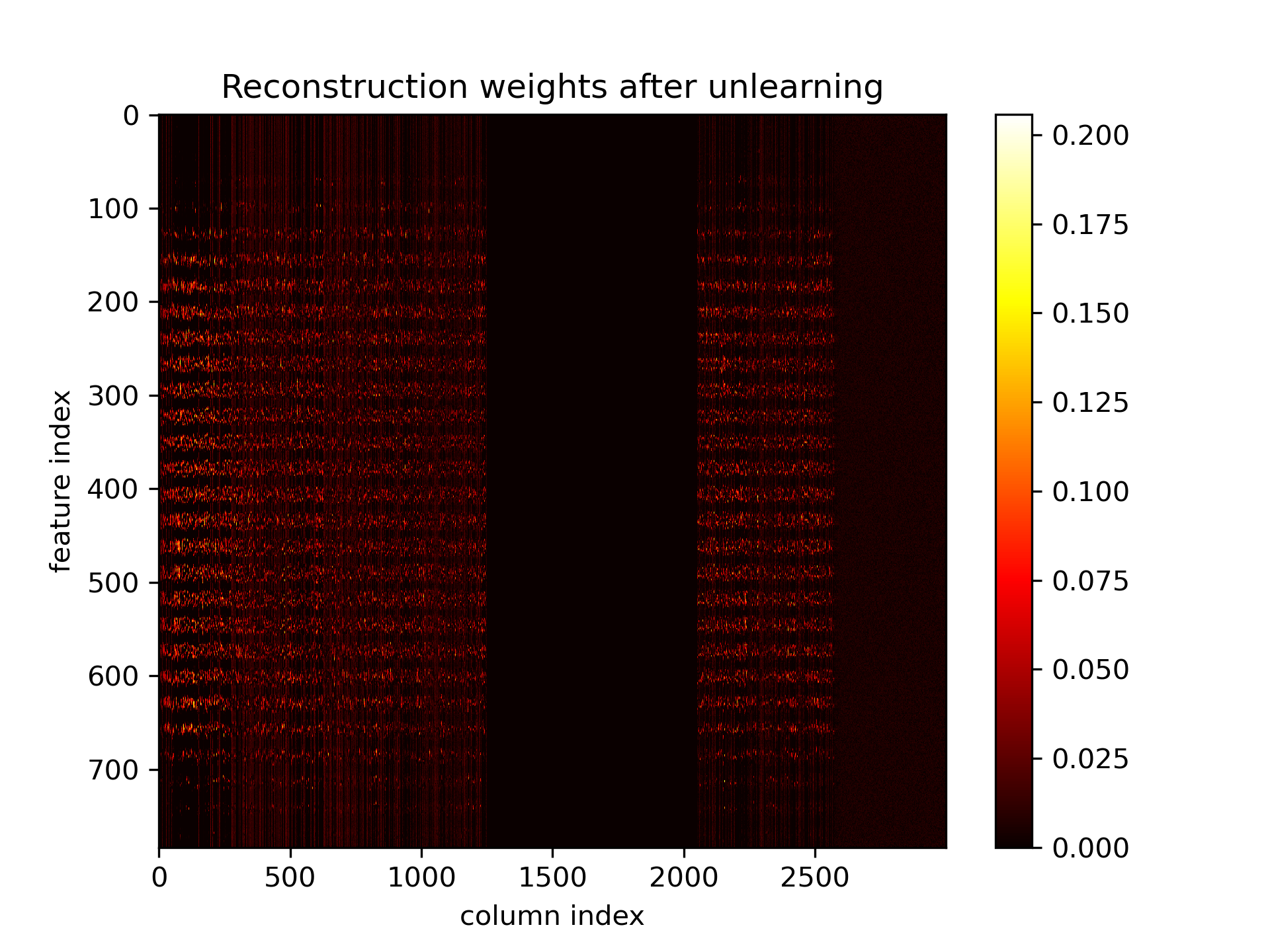}
    \end{minipage}
    \caption{Weights after unlearning}
    \label{fig:unlearning_after}
  \end{subfigure}
  \caption{Model weights before and after unlearning. 2600 out of the 3000 columns of in use, resulting in unused randomly initialized values in the right-most 400 columns. After unlearning, notice that column indices 1250 through 2050 have been removed.}
  \label{fig:unlearing_weights}
\end{figure}

After the model is trained, 2600 of the 3000 basis columns are in use, as Figure \ref{fig:unlearning_before} shows. The right-most 400 columns remain at their randomly initialized values, but this is not a problem and does not affect the results since these columns are not activated during the NMF inference process. Since the training data included a significant fraction of corrupted examples, the classification accuracy is a somewhat low 83.81\% on the MNIST test set.

Next, we perform the unlearning operation, which will attempt to remove the knowledge obtained from the corrupted training examples from the network. This works as follows. First, recall that training batches 500 through 800 were identified as containing the corrupted labels. Since our SLW optimizer constrains the optimizer update for each batch to a small learnable window of the weights, we need to find the corresponding union of all positions of the learnable window during learning updates for these batches. We find that batch index 500 corresponds to the left-most column of the learnable window being at index 1250, and batch index 800 corresponds to the right-most column of the learnable window being at index 2050. That is, during this range of (corrupted example) batch updates, the sliding learnable window covered columns in the weight matrices ranging from 1250 through 2050, so that any knowledge learned from these batches must be in this subset of the weights. It is then straightforward to remove the knowledge, such as by deleting these weights, setting them to zero, or reinitializing to random values. For this experiment, we set these weights to zero, as shown in Figure \ref{fig:unlearning_after}. With the corrupted knowledge removed, we then re-evaluate the network on the test set and see that the accuracy has improved to 92.77\%. For reference, if we train the model from scratch excluding the corrupted examples, we get 95.35\% on the test set, which sets an upper bound on the unlearning accuracy. These results are summarized in Table~\ref{table:mnist-accuracies}. This shows that our method is effective in restoring accuracy without retraining, provided that we are able to identify which training batches were bad. Note that if any training examples are identified as bad, then the entire batch and corresponding region of the weights must be thrown out. This method is therefore most effective when the subset of batches to remove corresponds to a contiguous sequence of batches in the training data.

\begin{table}[h]
  \centering
  \caption{Unlearning performance on MNIST Classification}
  \label{table:mnist-accuracies}
  \begin{tabular}{lc}
      \toprule
      Model Weights at Evaluation  & Test Accuracy \\
      \midrule
      Including only good data (upper bound)  & 95.35\% \\
      Before unlearning & 83.81\% \\
      After unlearning & 92.77\% \\
      \bottomrule
  \end{tabular}
\end{table}

\subsection{Visualizing out of domain interpretability}
\label{sec:vis_ood}

In this section, we train a network containing 1 PFC block on an image classification task and then evaluate it on both in-domain and out-of-domain (OOD) inputs. Since the network produces reconstructed input features during the recognition process, it is interesting to visualize and compare these reconstructions on in-domain vs OOD inputs. We might expect that the learned weights would correspond to the parts of the in-domain images, so that the reconstruction quality should be better on in-domain vs OOD inputs. This is because when the network is given an OOD image, it is forced to attempt to reconstruct it using on the ``in domain'' parts, potentially reducing the reconstruction quality compared to in-domain inputs. 

We now empirically investigate these effects using MNIST as the in-domain images and Fashion MNIST as the OOD images. We train a small network on the MNIST classification task using 100 weight templates to enable the easy visualization of all weights in a single plot figure. We use the combined input reconstruction and classification loss as usual, except that here we use an adjustable trade-off between the two in the form of a hyperparameter $\lambda \in [0, 1]$:

\begin{align}
  L = \lambda L_{classification} + (1 - \lambda) L_{reconstruction}
\end{align}

\begin{figure}[h]
  \centering
  \begin{subfigure}[b]{0.32\textwidth}
      \centering
      \includegraphics[width=\textwidth]{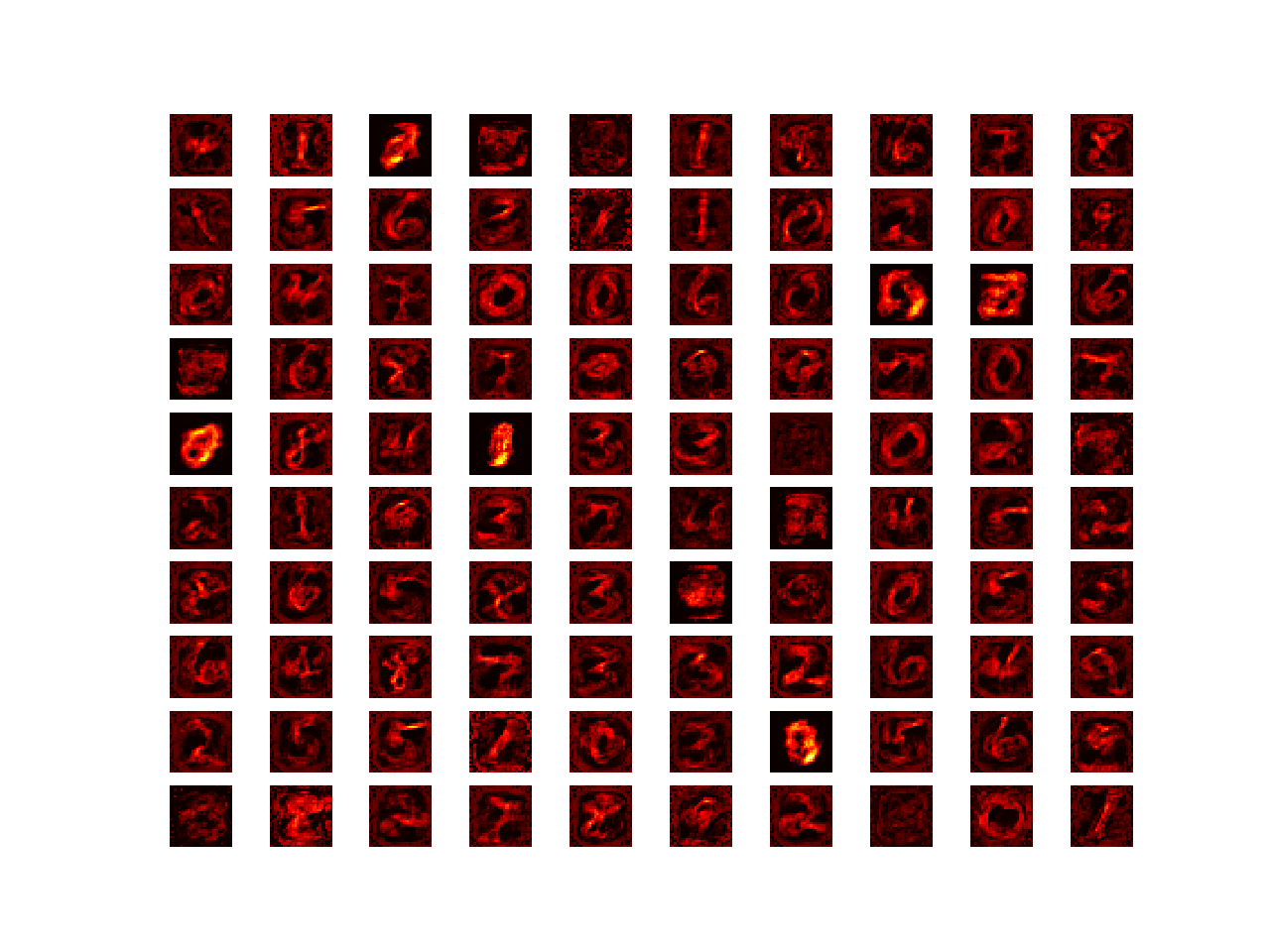}
      \caption{$\lambda = 0.9$, accuracy = 96.21\%}
      \label{fig:weights_vs_lambda_a}
  \end{subfigure}
  \begin{subfigure}[b]{0.32\textwidth}
      \centering
      \includegraphics[width=\textwidth]{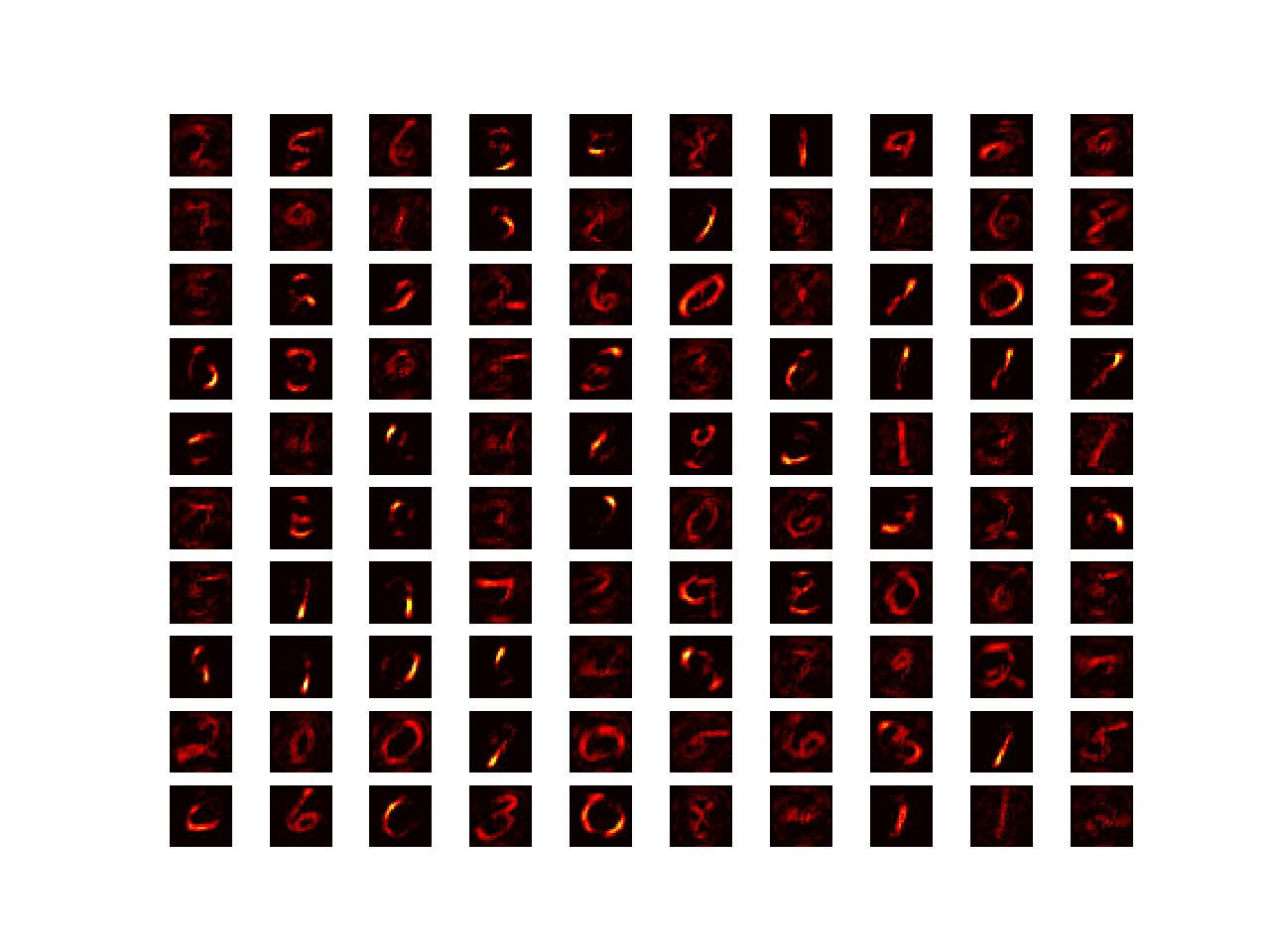}
      \caption{$\lambda = 0.5$, accuracy = 95.21\%}
      \label{fig:weights_vs_lambda_b}
  \end{subfigure}
  \begin{subfigure}[b]{0.32\textwidth}
      \centering
      \includegraphics[width=\textwidth]{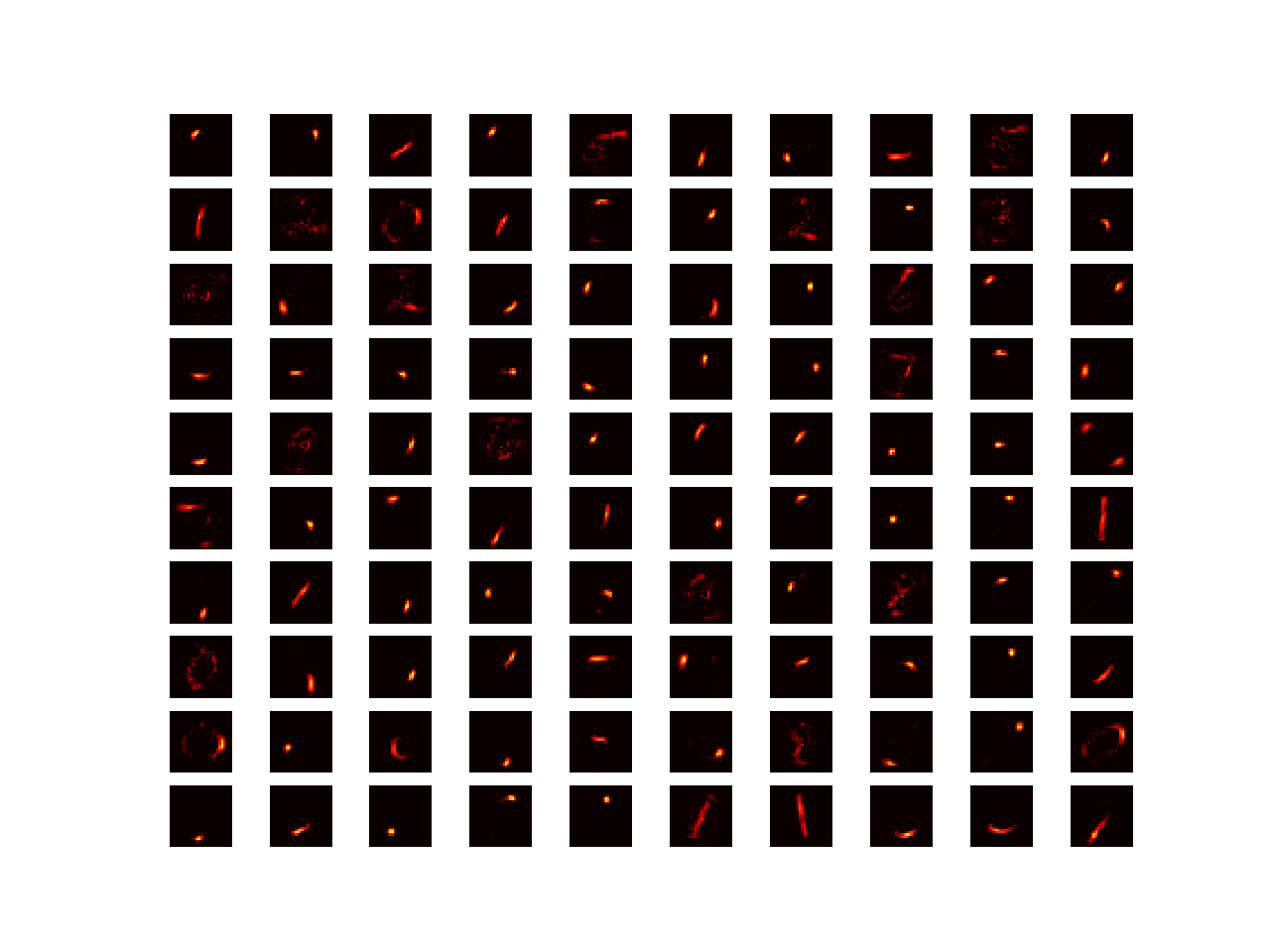}
      \caption{$\lambda = 0.1$, accuracy = 85.94\%}
      \label{fig:weights_vs_lambda_c}
  \end{subfigure}
  \caption{Visualization of model weights for different trade-offs between classification accuracy and input reconstruction quality, for different values of $\lambda \in [0, 1]$. Corresponding classification accuracy is also shown. As $\lambda$ is increased, the strength of the classification loss increases and the reconstruction loss decreases.}
  \label{fig:weights_vs_lambda}
\end{figure}

Figure \ref{fig:weights_vs_lambda} shows the weights, reshaped into images, for models trained using different values of $\lambda$. As $\lambda$ is increased, the strength of the classification loss increases and the reconstruction loss decreases. Thus, sub-figure \ref{fig:weights_vs_lambda_a} corresponds to a loss that emphasizes classification accuracy since only a small reconstruction loss is used. This is reflected in the good classification accuracy. We see that some of the weights resemble MNIST digits, but the images appear somewhat ``noisy''. The middle sub-figure \ref{fig:weights_vs_lambda_b} shows the weights using the default blend of classification and reconstruction loss used in most of the other experiments. Here we see that the weights resemble MNIST digits or parts thereof. Finally, sub-Figure \ref{fig:weights_vs_lambda_c} shows the effect of a strong reconstruction loss. We see that the weights now appears as more localized image parts and that the classification accuracy is significantly lower. 

\begin{figure}[h]
  \centering
  \begin{subfigure}[b]{0.45\textwidth}
      \centering
      \includegraphics[width=\textwidth]{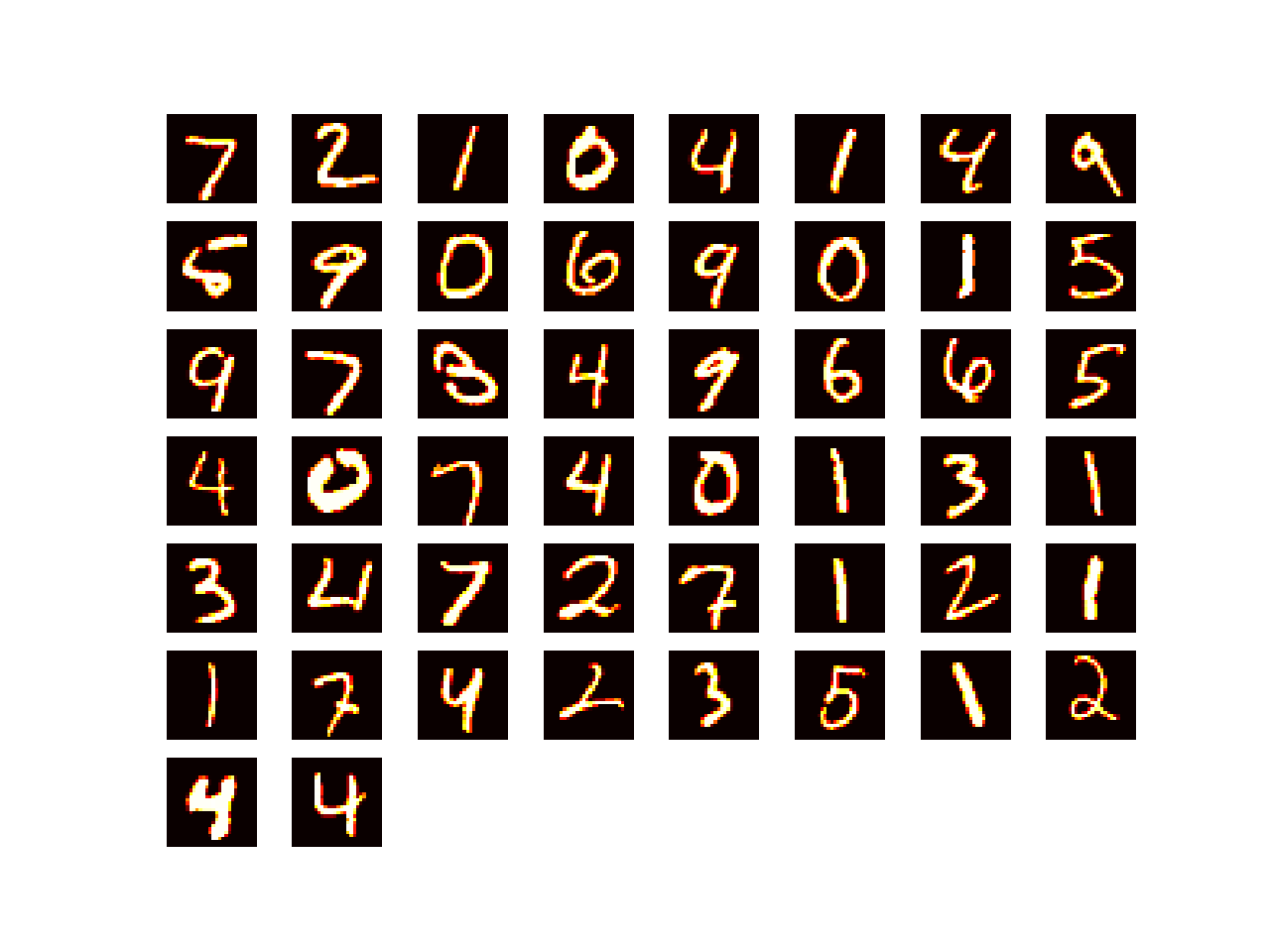}
      \caption{Input images (in-distribution, MNIST test set)}
      \label{fig:in_distribution_response_a}
  \end{subfigure}
  \hfill  
  \begin{subfigure}[b]{0.45\textwidth}
      \centering
      \includegraphics[width=\textwidth]{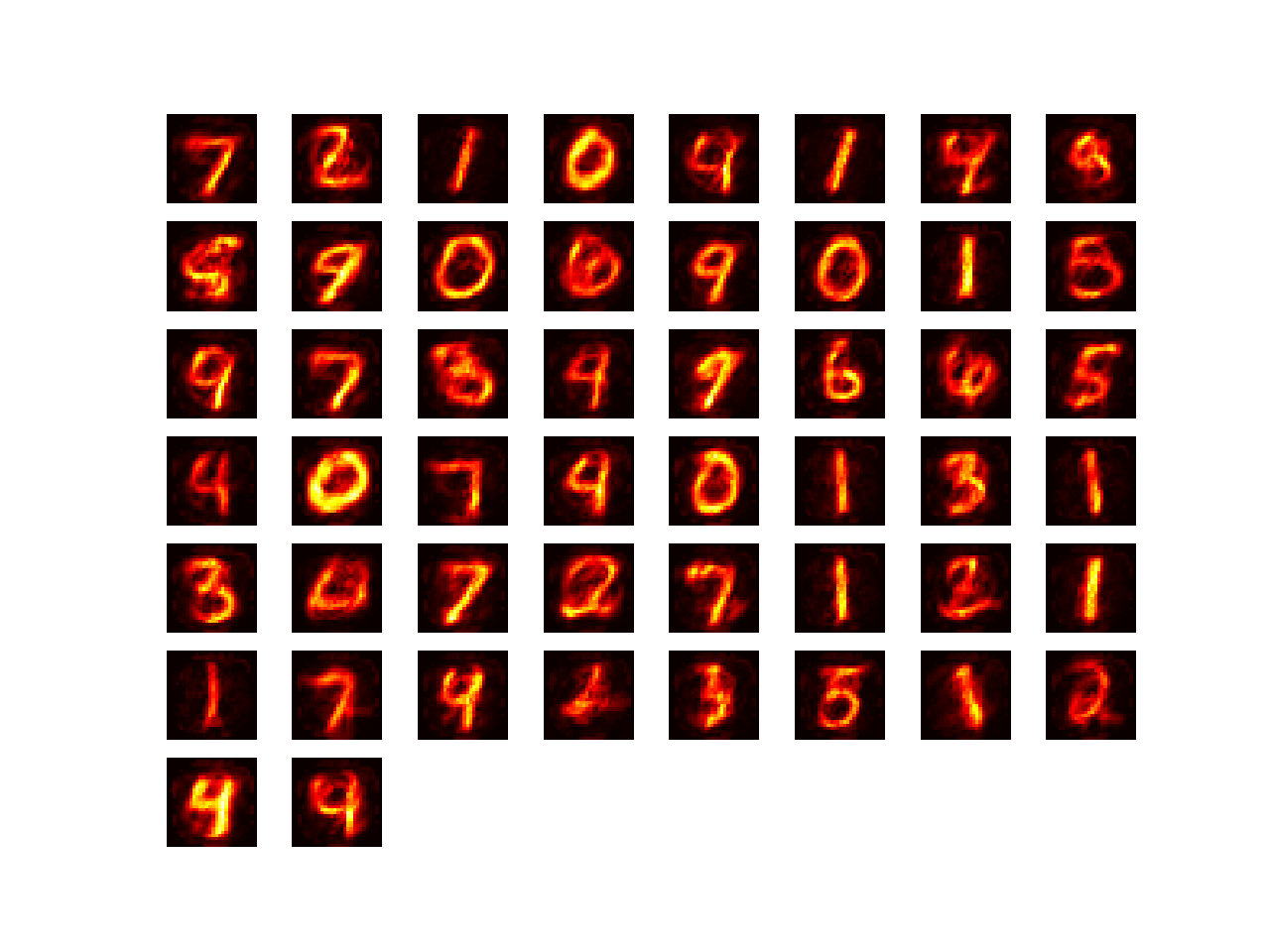}
      \caption{Reconstructed images}
      \label{fig:in_distribution_response_b}
  \end{subfigure}
  \caption{Model response to in-distribution inputs. The model was trained on an MNIST classification task and also evaluated on MNIST images here.}
  \label{fig:in_distribution_response}
\end{figure}

We now compare the input reconstructions produce by the model for in-distribution vs OOD examples. For the following, keep $\lambda$ set to the default value of 0.5. For the in-distribution visualization, we train on MNIST and also evaluate on a batch of MNIST test images, as shown in Figure \ref{fig:in_distribution_response}. Sub-figure \ref{fig:in_distribution_response_a} shows a batch of input MNIST test images and sub-figure \ref{fig:in_distribution_response_b} shows the corresponding input reconstructions generated by the model. We see that many of reconstructed images resemble the corresponding inputs, although some are less recognizable.

\begin{figure}[h]
  \centering
  \begin{subfigure}[b]{0.45\textwidth}
      \centering
      \includegraphics[width=\textwidth]{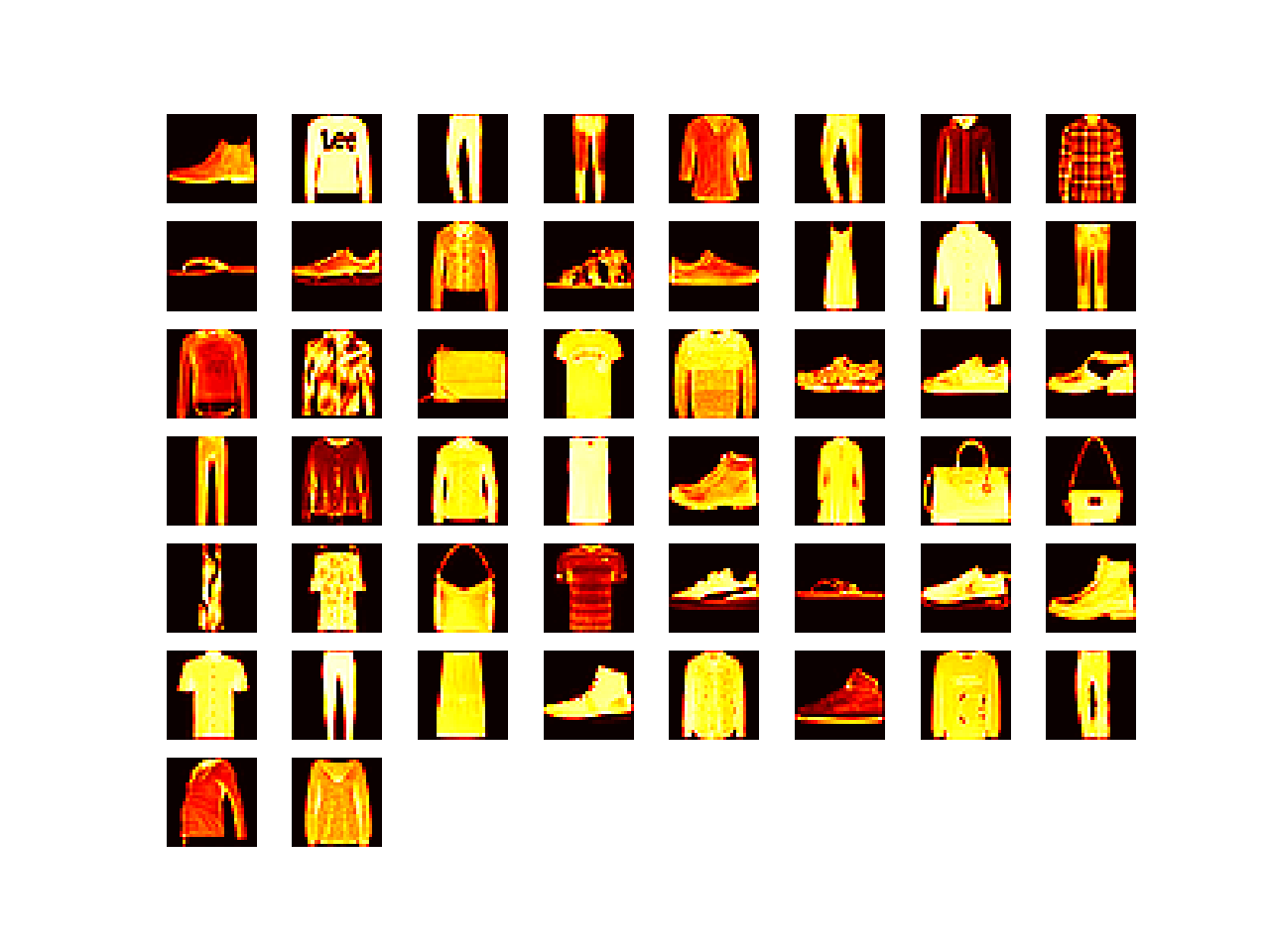}
      \caption{Input images (OOD, Fashion MNIST)}
      \label{fig:oode_a}
  \end{subfigure}
  \hfill  
  \begin{subfigure}[b]{0.45\textwidth}
      \centering
      \includegraphics[width=\textwidth]{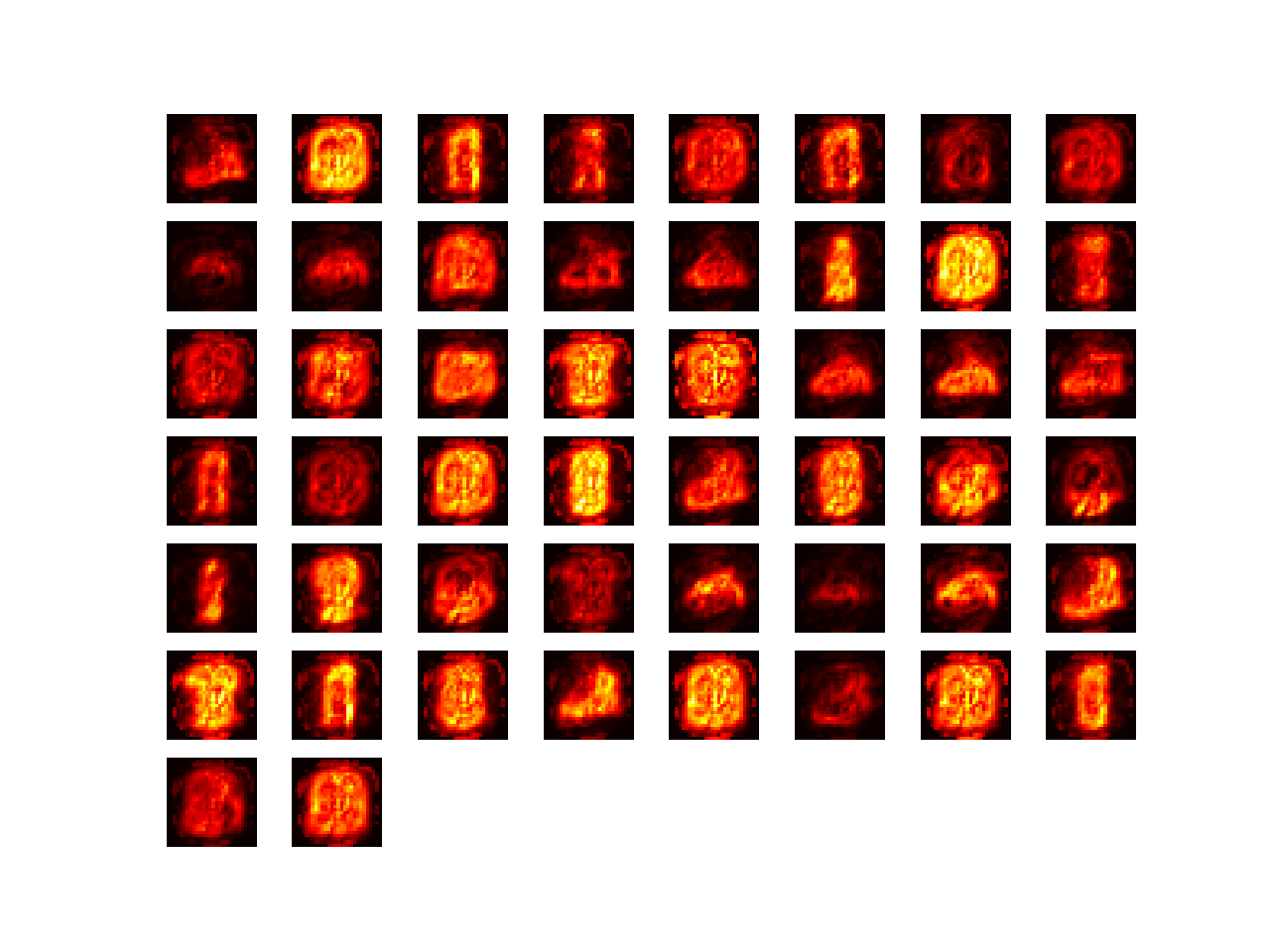}
      \caption{Reconstructed images}
      \label{fig:ood_b}
  \end{subfigure}
  \caption{Model response to OOD inputs. The model was trained on an MNIST classification task and but evaluated on Fashion MNIST images here.}
  \label{fig:ood_response}
\end{figure}

We now move on to the visualization of the model's reconstructed images when given OOD inputs. For this we will use a batch of images from the Fashion MNIST test set, which contains grayscale images of fashion products of the same size as MNIST images. Recall that the model is constrained to reconstruct an input image by solving for the best additive combination of its weights (i.e., using the 100 images in \ref{fig:weights_vs_lambda_b}). However, since the available weights correspond to MNIST images and/or their parts, we might expect the reconstructed Fashion MNIST inputs to have less resemblance to the actual images. Indeed, this is what we observe in Figure \ref{fig:ood_response}, which shows the OOD inputs and their reconstructions.

\subsection{Residual PFC network for image classification}
\label{sec:frp_2_layer}

In this experiment we use more than one PFC block to build a more complex architecture. The main purpose of this is to verify that if there are no optimization difficulties then the network should produce similar accuracy as that 1-block network in Section \ref{sec:frp_1_layer}. We construct a simple fully-connected residual network containing 2 PFC blocks in which the first block uses a skip connection. That is, we let $x$ represent the input to the first PFC block, which then produces output $y_1$. The input to the second PFC block is then given by $x_2 = relu(x - y_1)$. The relu is optional when using the semi-NMF assumption but is needed when using the NMF constraint to prevent the input $x_2$ from becoming negative. The second PFC block then outputs the final prediction $y_{pred}$.

We train on the same datasets as in Section \ref{sec:frp_1_layer} using the same basis vector sizes (equivalent to MLP hidden dimension). Since there are two blocks, the total number of parameters is doubled from the 1-block model. Here we only train using the NMF modeling constraint (non-negative parameters) and we use both prediction and input reconstruction MSE loss terms for each PFC block.
Table \ref{table:pfc_2_layer_image_classification} shows the accuracy results on the test set. We see that the accuracy results appear in a similar range compared to those of the 1-block model in Table \ref{table:pfc_1_layer_image_classification}.

\begin{table}[h]
  \centering
  \caption{Results of the residual 2-block PFC-based image classifier model. Parameters are constrained to be non-negative.}
  \label{table:pfc_2_layer_image_classification}
  \begin{tabular}{lcc}
      \toprule
      Dataset  & Basis Vector Count & Test Accuracy \\
      \midrule
      MNIST & 300  & 97.81\% \\ 
      MNIST & 2000 & 98.30\% \\
      MNIST & 5000 & 98.09\% \\
      \hline
      Fashion MNIST & 300  & 88.92\% \\ 
      Fashion MNIST & 2000 & 89.89\% \\
      Fashion MNIST & 5000 & 90.08\% \\
      \hline
      CIFAR10 & 300  & 54.00\% \\ 
      CIFAR10 & 2000 & 56.26\% \\
      CIFAR10 & 5000 & 58.15\% \\
      \bottomrule
  \end{tabular}
\end{table}

\subsection{Memorizing a deterministic sequence with a factorized RNN}
\label{sec:standard_nmf_rnn}

We developed the factorized RNN in Section \ref{sec:the_factorized_rnn} and modeled it as the matrix factorization of the form $V \approx W H$ shown in Eq. \ref{eqn:factorized_vanilla_rnn} which we repeat here:

\begin{align}
  \begin{bmatrix} 
    y_0 & y_1 & y_2 & \dots & y_{T-1} \\ 
    h_{-1} & h_0 & h_1 & \dots & h_{T-2} \\
    x_0 & x_1 & x_2 & \dots & x_{T-1}
    \end{bmatrix} \approx \begin{bmatrix} W_y\\ W_h \\ W_x \end{bmatrix} \begin{bmatrix} h_0 & h_1 & h_2 & \dots & h_{T-1} \end{bmatrix}
\end{align}

We also provided a more compact notation in which the sequences are replaced by their respective sub-matrices in Eq. \ref{eqn:factorized_vanilla_rnn_submats}, which we repeat here:

\begin{align}
  \begin{bmatrix} 
    Y \\ 
    H_{prev} \\
    X
    \end{bmatrix} \approx \begin{bmatrix} W_y\\ W_h \\ W_x \end{bmatrix} H
\end{align}

We then provided two basic training procedures. Here we consider the first method which uses matrix factorization update rules for both inference and learning as described in Sections \ref{sec:standard_nmf_factorized_rnn}. It is initially unclear whether such simple update rules could even learn a task spanning several time slices, since there is no backward flow of error gradient-like information through the sequence. We therefore think it seems appropriate to start with a relatively simple temporal learning task.

\subsubsection{Training on a repeating sequence}

For this initial task, we present a fixed-length pattern that is repeated over and over in the training data and require the network to predict the next item in the sequence. Since we use a deterministic repeating pattern, the network only needs to identify and memorize this underlying pattern in order to predict with perfect accuracy. We also chose this task because we know the underlying generative model corresponds to a simple finite state machine (FSM) containing the same number of (deterministic) transitions as there are time slices in the repeating pattern. We therefore know the minimum number of model parameters that are needed in principle to solve it and it is straightforward to imagine interpretable solutions to the corresponding RNN factorization ourselves. This task therefore also serves as a simple interpretability test for the model since we can train the model and then visualization the learned weights and see whether they align with the interpretable solutions that we know should be possible.

Specifically, we use the following fixed repeating pattern consisting of 25 4-dimensional 1-hot vectors for easy presentation and visualization, shown here represented as integer-valued tokens for easier readability:

\begin{align}
  [0, 1, 1, 2, 2, 2, 3, 3, 3, 3, 3, 3, 3, 3, 3, 1, 2, 2, 2, 2, 2, 1, 3, 2, 1]
\end{align}

\begin{figure}[h]
    \centering
    \includegraphics[width=0.7\textwidth]{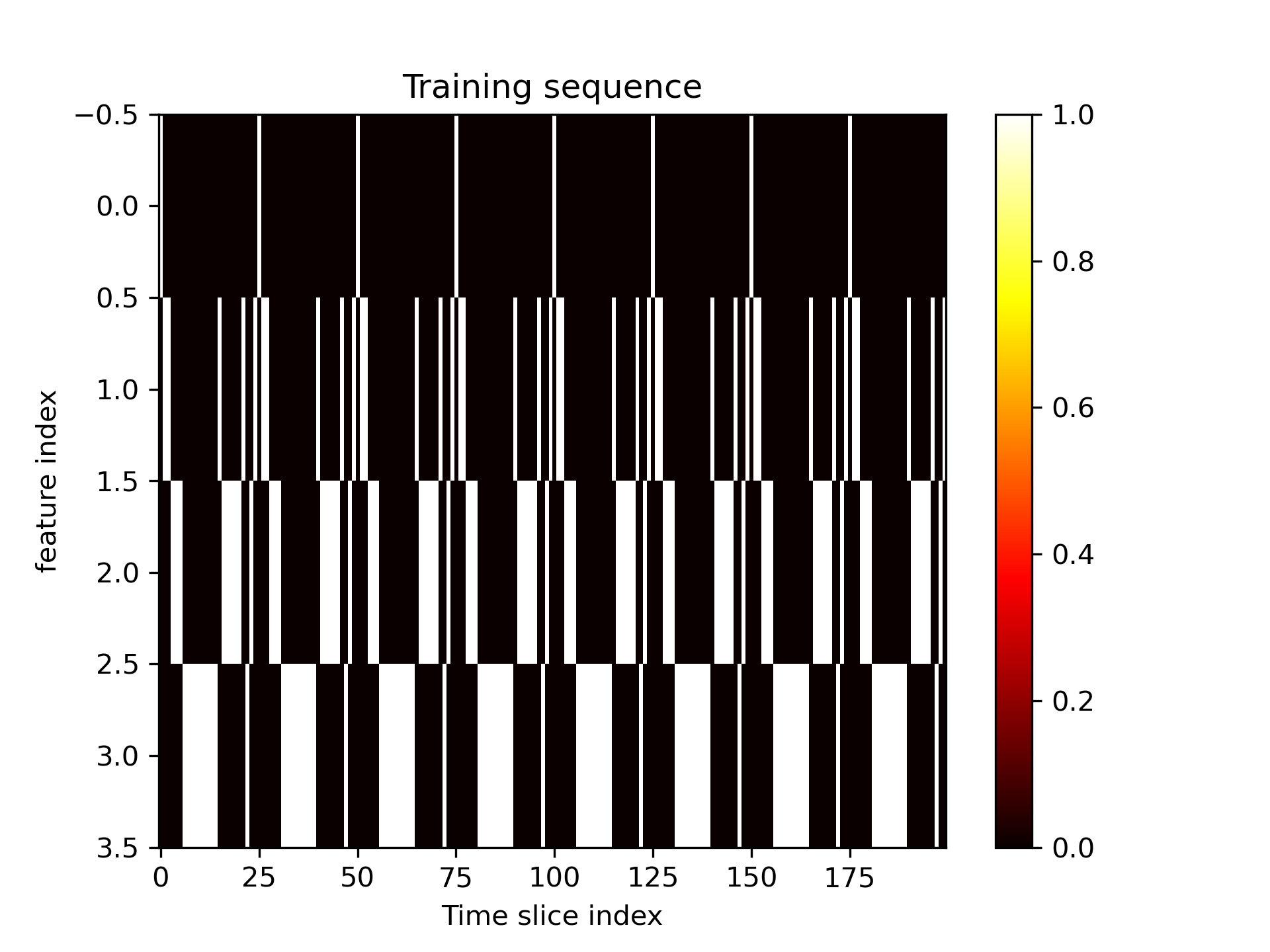}
    \caption{A training sequence $X$ of length 200 consisting of a fixed repeating sub-sequence of 1-hot vectors of length 25.}
    \label{fig:deterministic_training_seq}
\end{figure}

We then repeat this pattern 8 times to create the full training sequence $X = \begin{bmatrix} x_0 & x_1 & x_2 & \dots & x_{T-1} \end{bmatrix}$ shown in Figure \ref{fig:deterministic_training_seq}, which has a length of $T = 200$.

We want the model to memorize the repeating pattern and so it needs to predict the next vector in the sequence. For this we can use an autoregressive model so that $Y = \begin{bmatrix} x_1 & x_2 & x_3 & \dots & x_{T} \end{bmatrix}$ in Eq \ref{eqn:factorized_vanilla_rnn_submats}. With this, Eq \ref{eqn:factorized_vanilla_rnn} looks like the following:

\begin{align}
  \begin{bmatrix} 
    x_1 & x_2 & x_3 & \dots & x_{T} \\ 
    h_{-1} & h_0 & h_1 & \dots & h_{T-2} \\
    x_0 & x_1 & x_2 & \dots & x_{T-1}
    \end{bmatrix} \approx \begin{bmatrix} W_y\\ W_h \\ W_x \end{bmatrix} \begin{bmatrix} h_0 & h_1 & h_2 & \dots & h_{T-1} \end{bmatrix}
\end{align}

We use NMF for the inference and learning updates. Specifically, we use SGD updates with negative value clipping and scaling of the rows and columns of the factor matrices to keep them from exploding as described in Section \ref{sec:sgd_nmf}. For the results in this section, we do not use any weight decay or sparsity regularizers so that only the non-negativity constraint is used. We initialize the weights $W = {W_y, W_h, W_x}$ to non-negative random values uniform in $[0, 1e-2]$ and initialize the hidden states ($H_{prev}$ and $H$) to zero. Recall from Section \ref{sec:the_factorized_rnn} that the hyperparameter $R$ specifies both the dimensionality of the hidden state vectors $h_k$ as well as the number of basis column vectors in $W$ so that $W_h$ is an $R$ x $R$ sub-matrix of $W$. Since the repeating sub-sequence has length 25, that seems to be the minimum value of $R$ that could have any chance of learning the state transition model. We then run the training procedure using alternating the NMF inference and learning update rules as described in Section \ref{sec:standard_nmf_factorized_rnn}. We ran several training runs for values of $R$ ranging between 25 and 500.

\subsubsection{Interpretation of the learned weights}

\begin{figure}[h]
  \centering
  \includegraphics[width=0.7\textwidth]{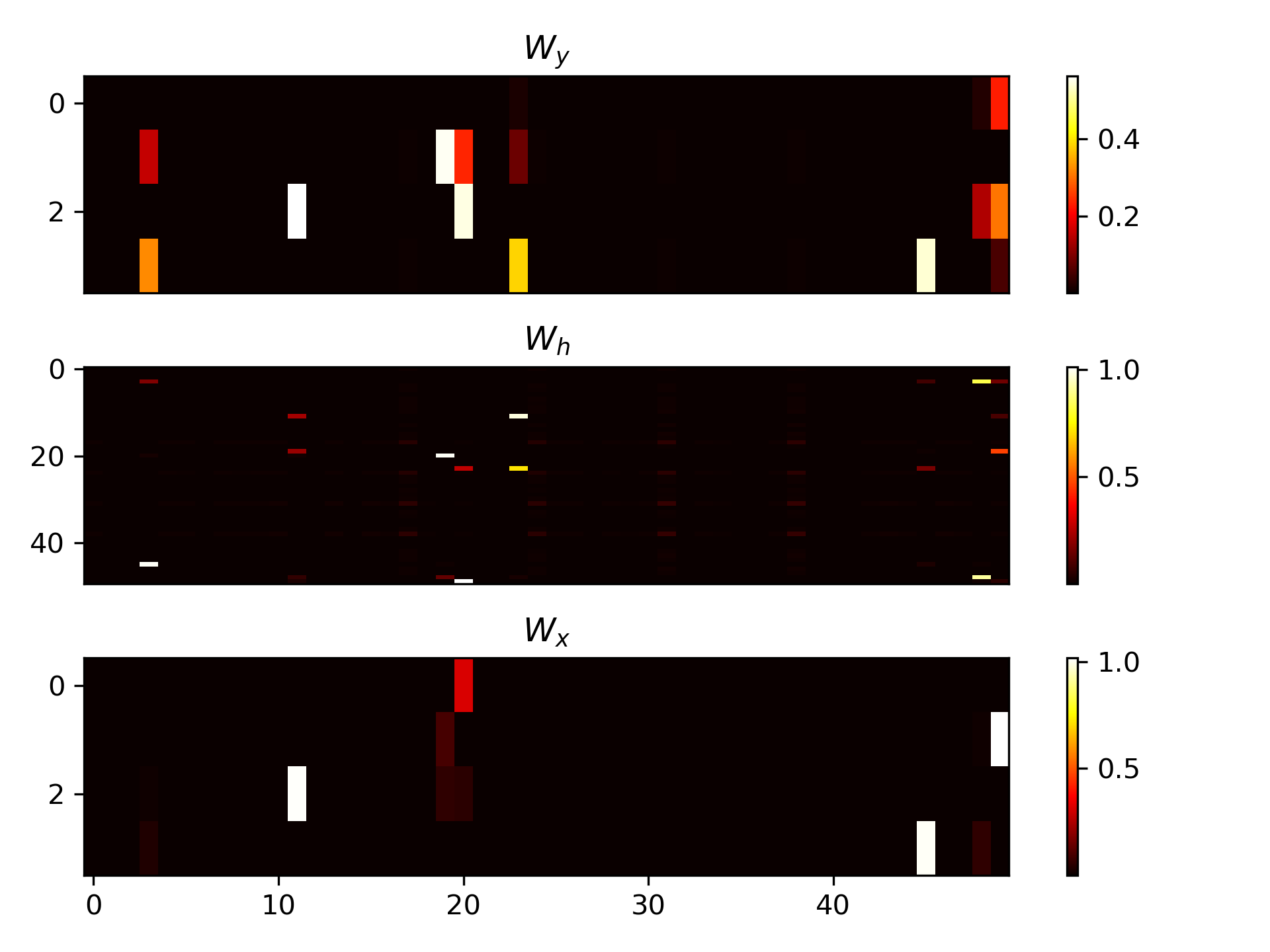}
  \caption{A failure case: learned weights after training on the repeating deterministic sequence in Figure \ref{fig:deterministic_training_seq} for $R = 50$. Multiple training attempts were sometimes needed for values of $R$ between 25 and 100 and this shows one such example of unsuccessful training.}
  \label{fig:deterministic_rnn_hdim_50_fail}
\end{figure}

We observed some interesting and surprising results when training with different values of $R$. The first is that training seems to become faster and more reliable as $R$ is increased. The model was consistently able to learn an exact or nearly exact factorization (training MSE below 1e-7 or so) for $R$ around 100 or larger. However, we saw training gradually became less reliable as $R$ was decreased toward the lower limit of 25, often requiring multiple training attempts to successfully learn the factorization. Figure \ref{fig:deterministic_rnn_hdim_50_fail} shows one such unsuccessful training run with $R=50$, where the training MSE only converged to 0.112. Still, training was still sometimes successful even at the limit value $R = 25$. Since we only observed unreliable training with small values of $R$, we did not attempt any hyperparameter optimization in order to fix it. 

\begin{figure}[h]
  \centering
  \begin{subfigure}[b]{0.32\textwidth}
      \centering
      \includegraphics[width=\textwidth]{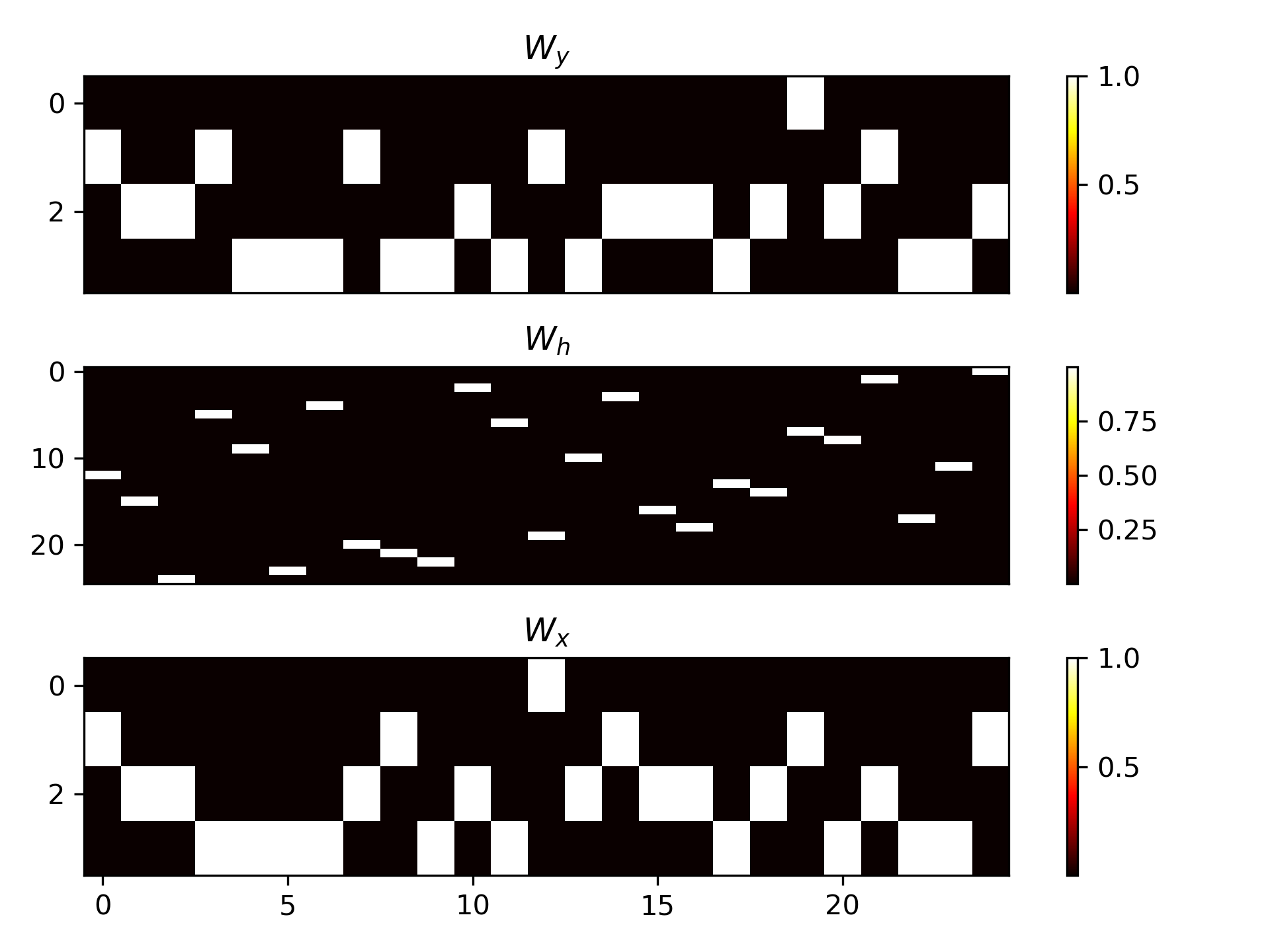}
      \caption{$R = 25$ basis columns}
      \label{fig:learned_weights_for_various_hdims_25}
  \end{subfigure}
  \begin{subfigure}[b]{0.32\textwidth}
      \centering
      \includegraphics[width=\textwidth]{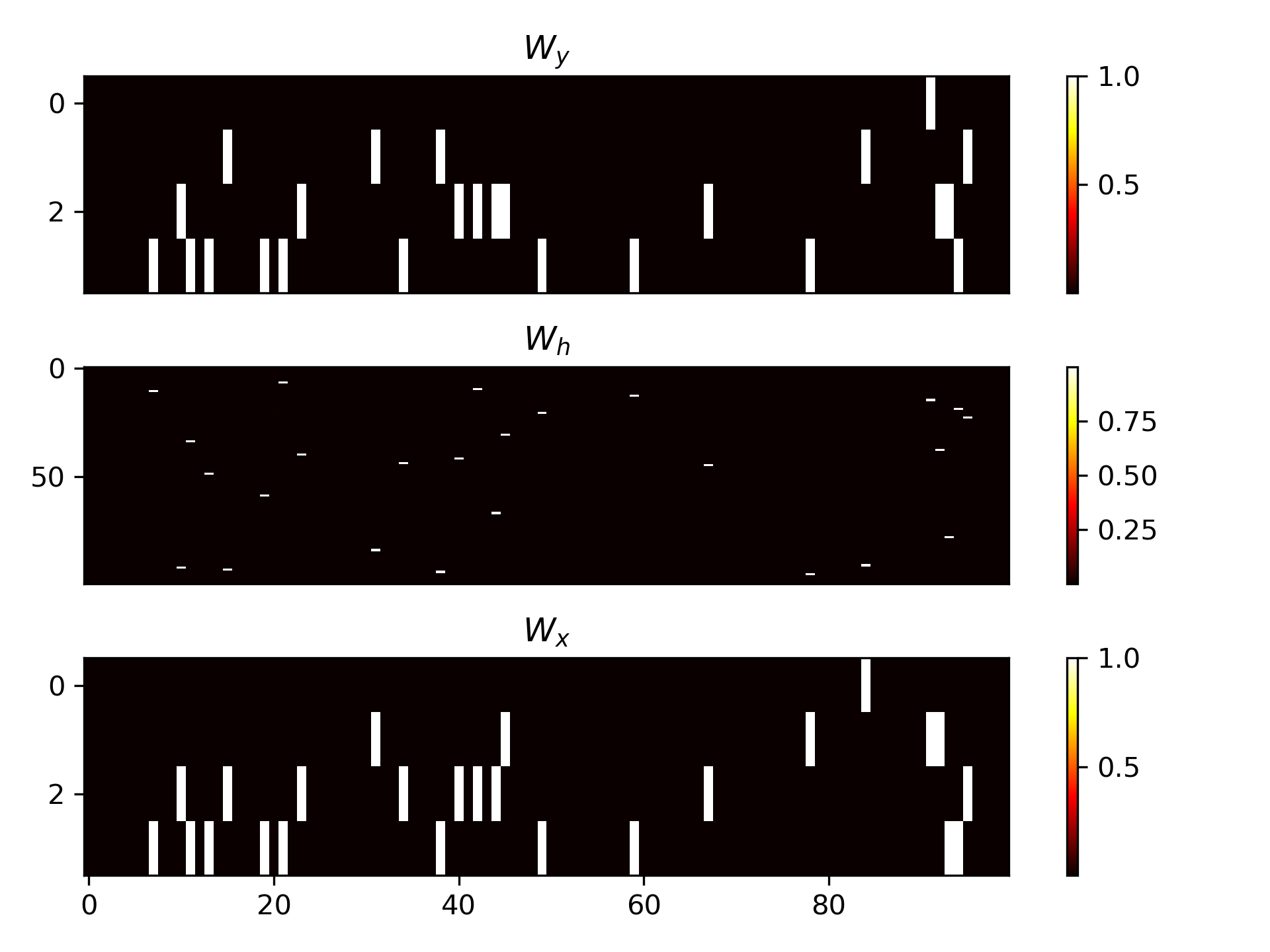}
      \caption{$R = 100$ basis columns}
      \label{fig:learned_weights_for_various_hdims_100}
  \end{subfigure}
  \begin{subfigure}[b]{0.32\textwidth}
      \centering
      \includegraphics[width=\textwidth]{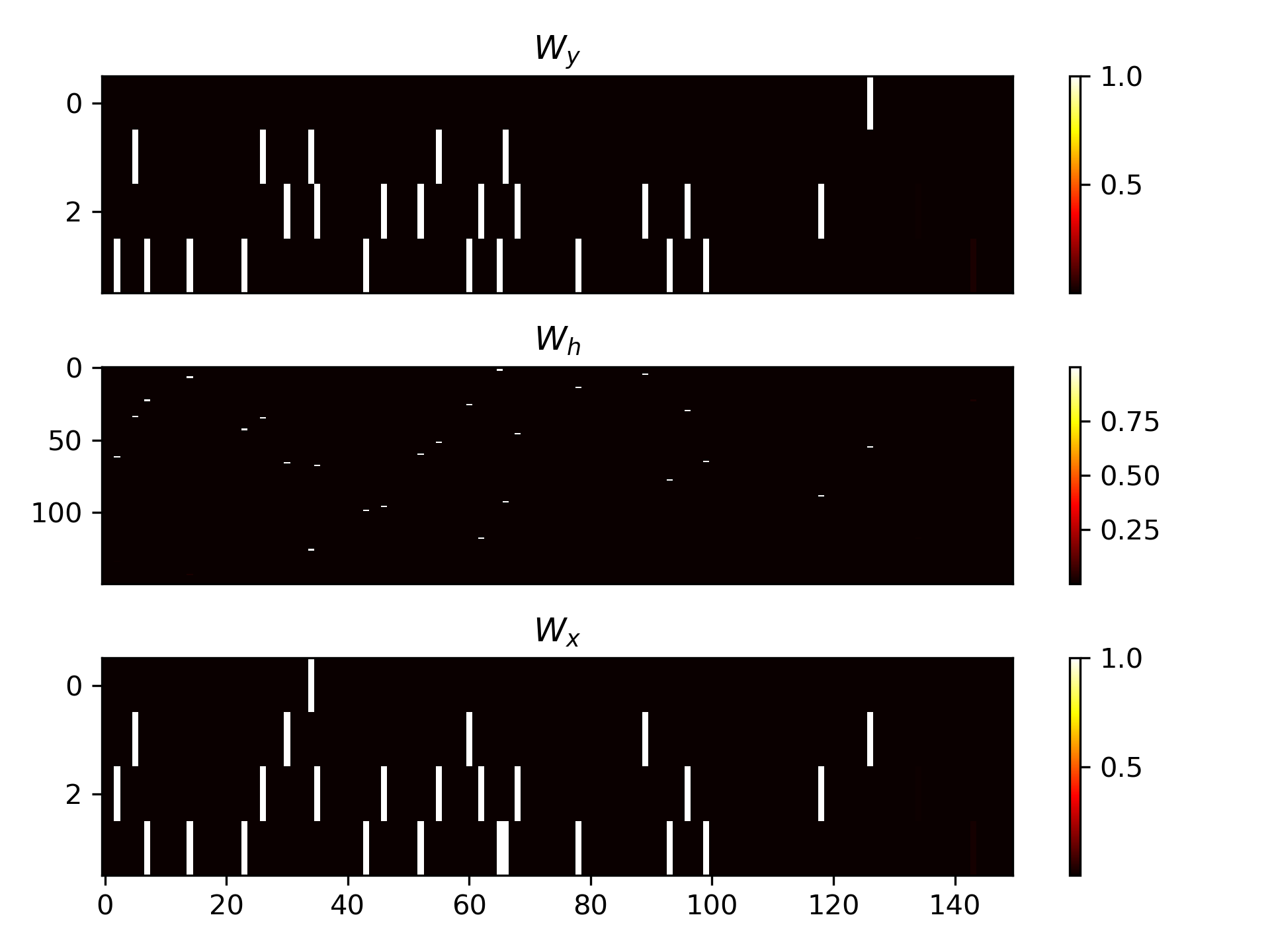}
      \caption{$R = 150$ basis columns}
      \label{fig:learned_weights_for_various_hdims_150}
  \end{subfigure}
  \caption{Learned weights after training on the repeating deterministic sequence in Figure \ref{fig:deterministic_training_seq} for three choices of the hidden state vector dimension $R$, which is equal to the number of basis columns in $W$. We see that in each case, 25 column vectors are learned.}
  \label{fig:learned_weights_for_various_hdims}
\end{figure}

Perhaps our most surprising observation was that the learned models tended to be highly sparse and interpretable even though we did not use any sparsity regularization. Regardless of the value of $R$, a successful training run resulted in the model discovering that only 25 basis vectors were actually needed in $W$, with the other columns tending toward 0. Additionally, the learned columns of the state transition matrix $W_h$ appeared close to 1-hot vectors as Figure \ref{fig:learned_weights_for_various_hdims} shows. Such a learned representation lets us understand the underlying state transition model that was used to generate the repeating sub-sequence by quick visual inspection of the model weights. Recall that for any particular activated basis column in $W$, the corresponding column of bottom sub-matrix $W_x$ additively reconstructs the input $x_k$. Similarly, the same column of $W_h$ additively reconstructs the previous state $h_{k-1}$, and the same column of $W_y$ additively reconstructs $y_k$, which is the prediction for the next input $x_{k+1}$. As a concrete example, consider the right-most basis vector $W[:, 24]$ in Figure \ref{fig:learned_weights_for_various_hdims_25}. Since $W_x[:, 24]$ has a 1 in the second row and $W_y[:, 24]$ has a 1 in the third row, this corresponds to explaining a 1-hot input vector similarly having a 1 in its second dimension and predicting that it will transition to a 1-hot vector having a 1 in its third dimension in the next time step. In the integer 1-hot sequence representation, this would correspond to $[1] \rightarrow [1, 2]$. The same column of $W_h[:, 24]$ is additively reconstructing the previous hidden state (as a 1-hot vector with a 1 in its first dimension). When the right-most basis vector (i.e., column index = 24) of $W$ is activated by a column in $H$ such as $H[24, k] = 1$, it causes the inferred next state to have a 1 in its last dimension. This inferred $h_k$ then becomes the previous $h_{k-1}$ input state in the next time slice, and we see that the third basis column of $W$ from the left has a 1 in the final dimension (i.e., $W_h[24, 2]$ = 1) which would cause this basis column to be activated in the next time slice $k+1$. Continuing in this way, we can easily read out the underlying transition model from inspection of $W$. Note that the ordering of the basis vectors in $W$ is significant because activating column $r$ of $W$ results in a corresponding positive activated value in dimension $r$ of the inferred next state vector. With this understanding, it makes sense that the learned $W_h$ would tend to be sparse even without any explicit sparsity regularization. Each activated basis column of $W_h$ becomes a positive entry in the corresponding dimension of the input previous state $h_{k-1}$ in the next time slice (recalling the inferred states in $H$ are copied into the next time slice of $H_{prev}$), which in turn needs to be explained as an additive combination of the basis vectors in $W_h$. That is, any activated columns in $W_h$ translate to corresponding non-zero (positive-valued) rows in the next time slice's input state vector. If $W_h$ contains a non-zero column that has multiple positive values in different rows and this column is activated by $h_k$, it implies multiple columns of $W_h$ must have been activated in the previous time slice. Consider also the case where there are duplicated columns in $W_h$. The NMF inference algorithm might then choose to activate both of them with some positive strength, again resulting in a non 1-hot $h_k$ that would in turn need to be explained in the next time slice. Intuitively, it then seems to make sense that the model would tend toward the sparsest possible learned representation.

\subsubsection{Evaluation and visualization}

\begin{figure}[h]
  \centering
  \includegraphics[width=0.7\textwidth]{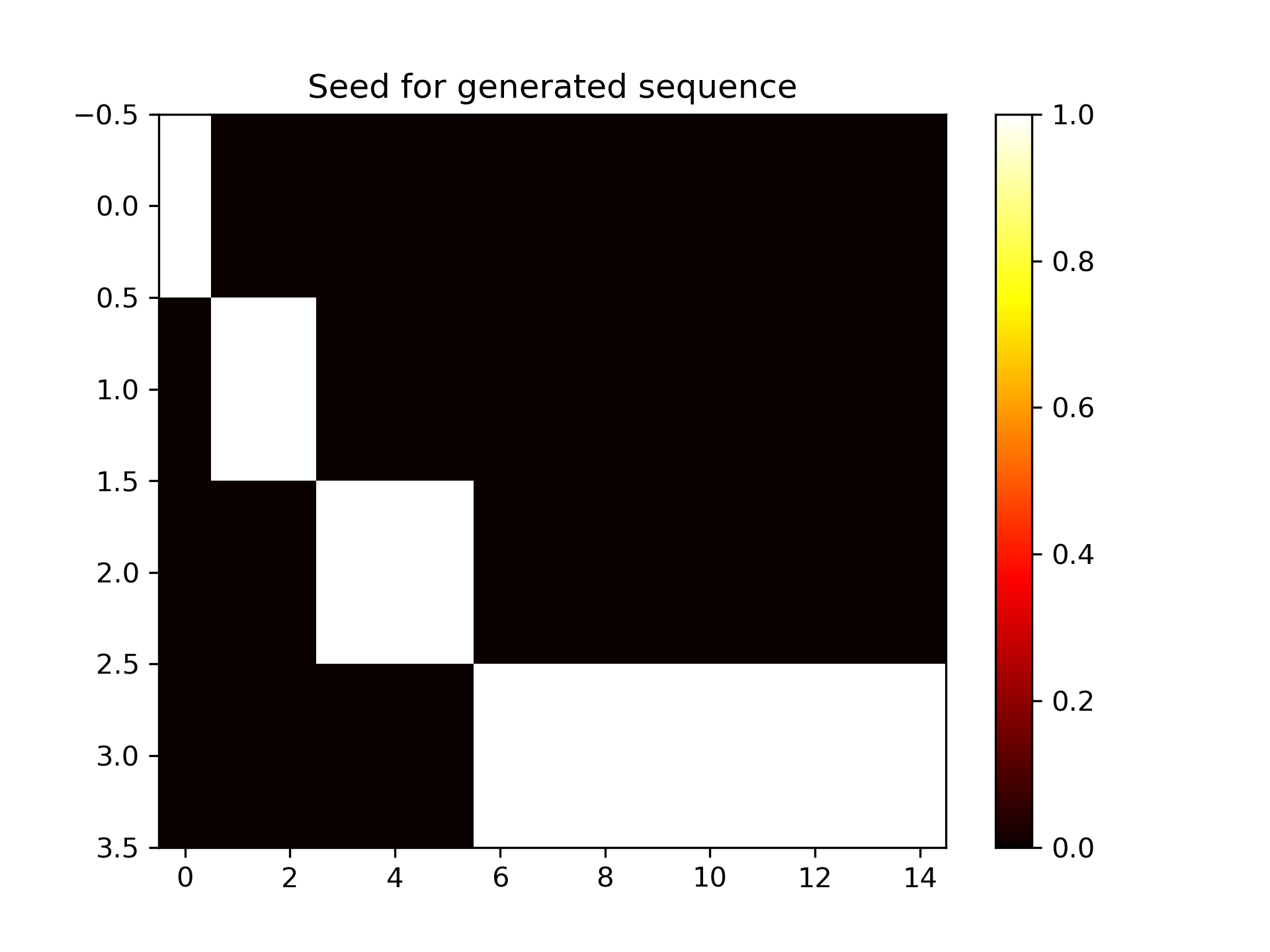}
  \caption{The input seed sequence consisting of the first 15 time slices of the repeating sub-sequence the model was trained on.}
  \label{fig:deterministic_seed_seq}
\end{figure}

\begin{figure}[h]
  \centering
  \includegraphics[width=0.7\textwidth]{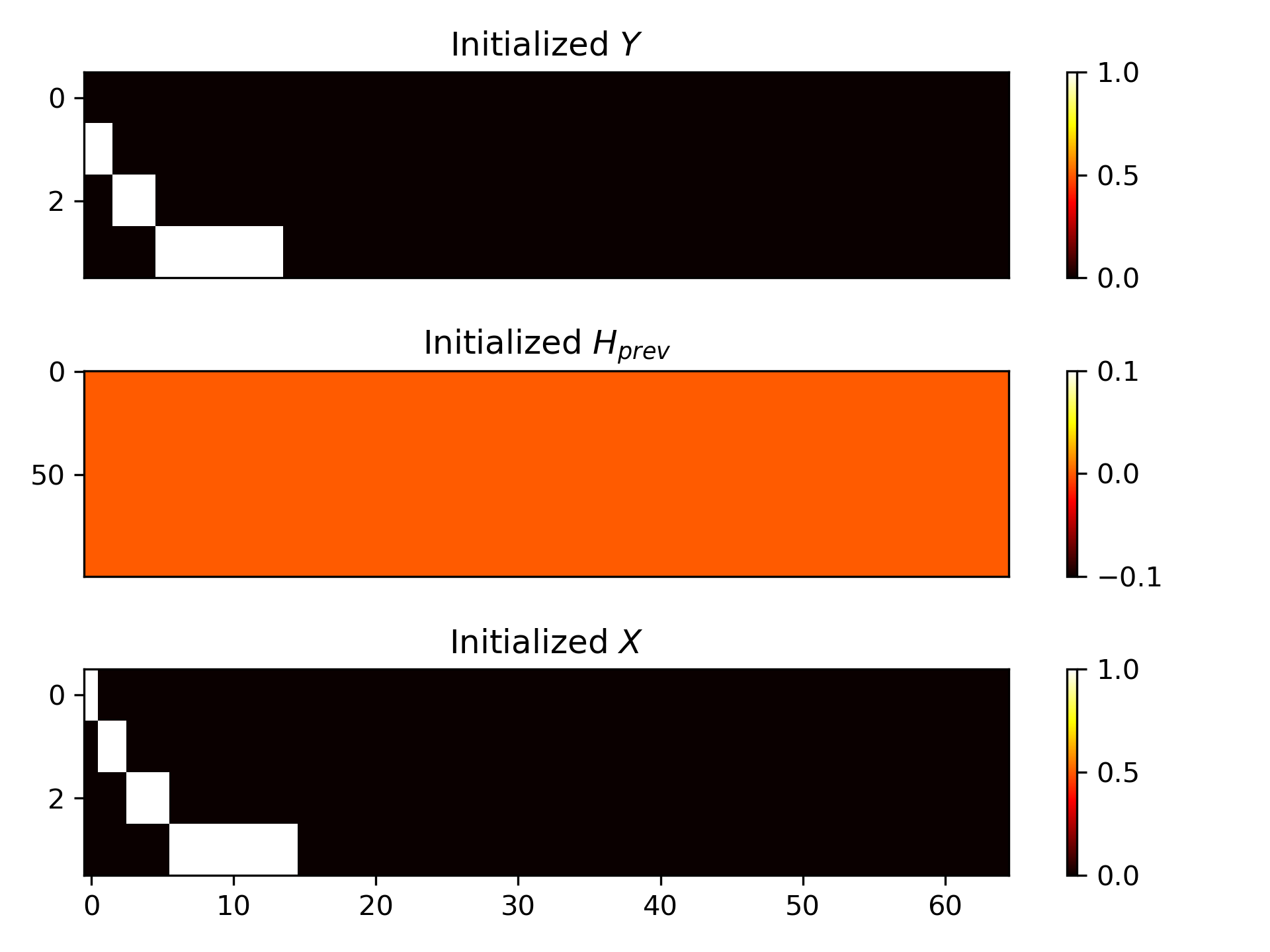}
  \caption{The sub-matrices of $V$ just before the generation task. The first 15 time slices of $X$ are initialized with the seed sequence from \ref{fig:deterministic_seed_seq}. Likewise, the first 14 time slices of $Y$ are initialized with the seed shifted 1 slice to the left. All hidden states are initialized to zeros.}
  \label{fig:initialized_V_for_gen}
\end{figure}

With the model weights trained, we can now verify that it has successfully memorized the sequence. We will provide a short ``seed'' sequence that contains part of the repeating sub-sequence and then ask the model to generate a continuation of it. For this evaluation, we use a model that was trained with $R = 100$ dimensional hidden state vectors. Figure \ref{fig:deterministic_seed_seq} shows the seed sequence, for which we use the first 15 time slices. We will generate 50 additional time slices after the seed sequence. We then initialize the $X$ and $Y$ sub-matrices of $V$ so that they only contain the seed as the initial part at the left as shown in Figure \ref{fig:initialized_V_for_gen}. 

\begin{figure}[h]
  \centering
  \includegraphics[width=0.7\textwidth]{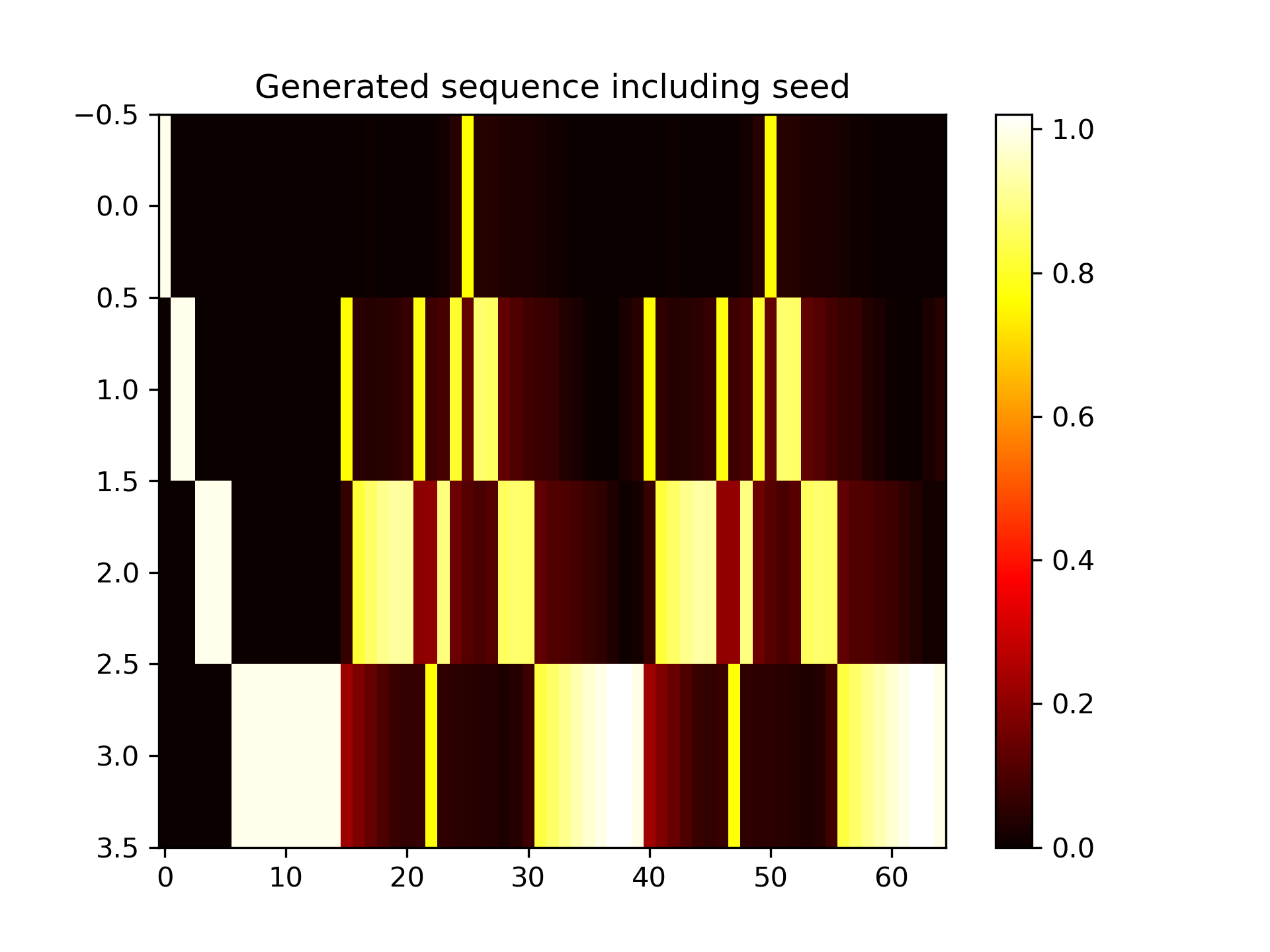}
  \caption{The generated sequence including the seed. The first 15 time slices are the seed and the remaining were generated.}
  \label{fig:generated_seq}
\end{figure}

\begin{figure}[h]
  \centering
  \begin{subfigure}[b]{0.32\textwidth}
      \centering
      \includegraphics[width=\textwidth]{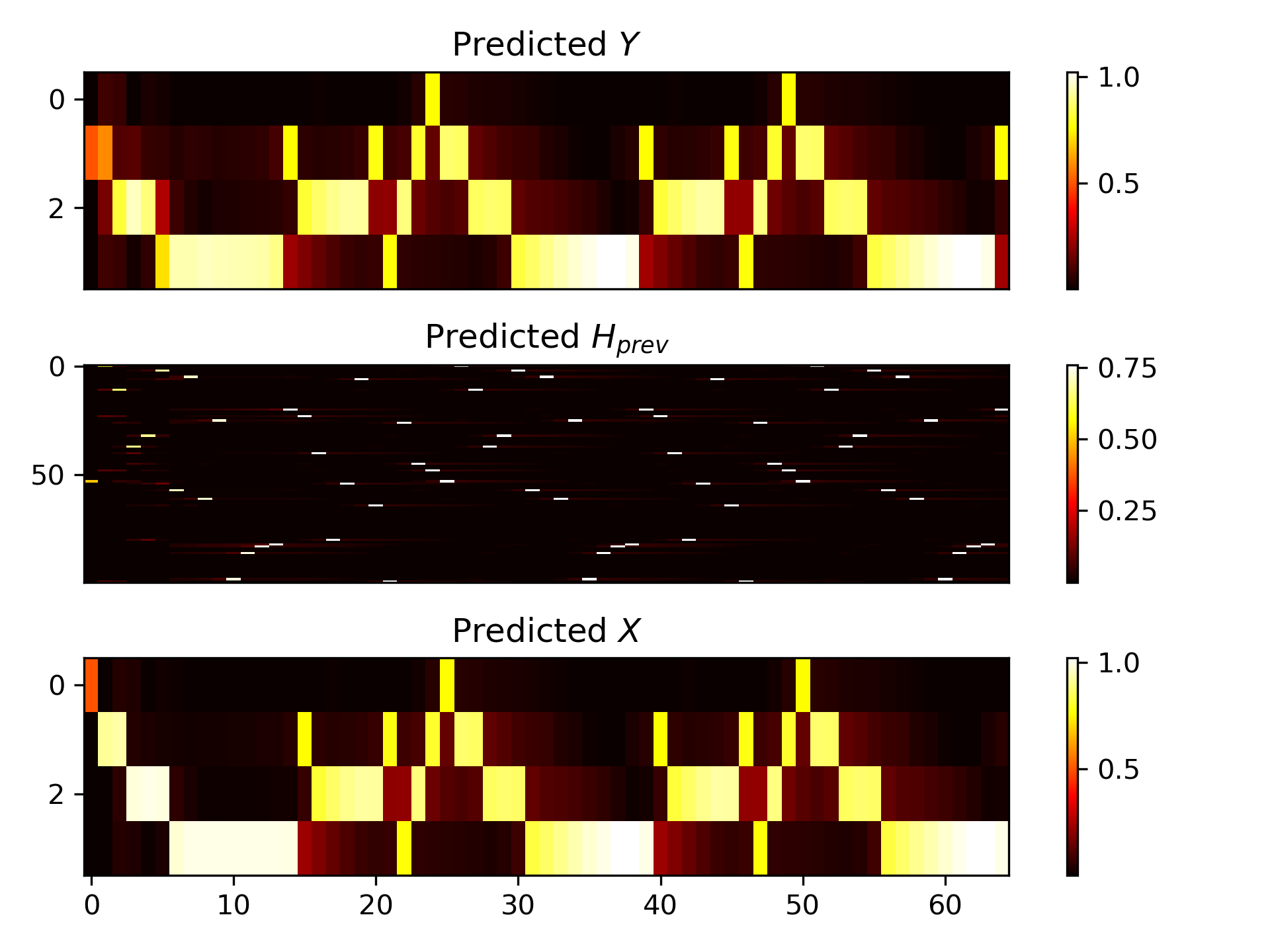}
      \caption{Predicted sub-matrices of $V$}
      \label{fig:factorization_after_gen_from_seed_V_pred}
  \end{subfigure}
  \begin{subfigure}[b]{0.32\textwidth}
      \centering
      \includegraphics[width=\textwidth]{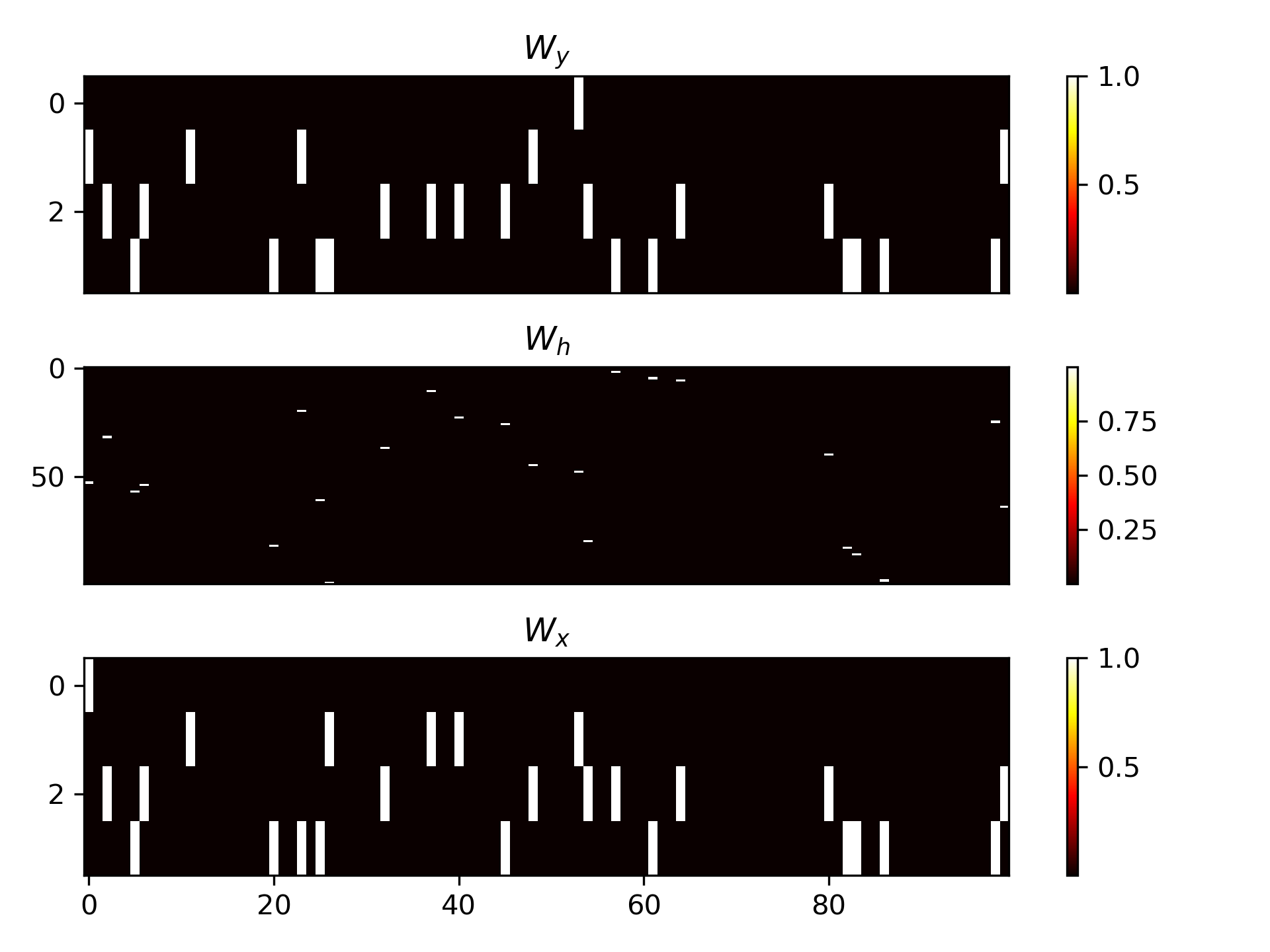}
      \caption{Learned weights $W$}
      \label{fig:factorization_after_gen_from_seed_W}
  \end{subfigure}
  \begin{subfigure}[b]{0.32\textwidth}
      \centering
      \includegraphics[width=\textwidth]{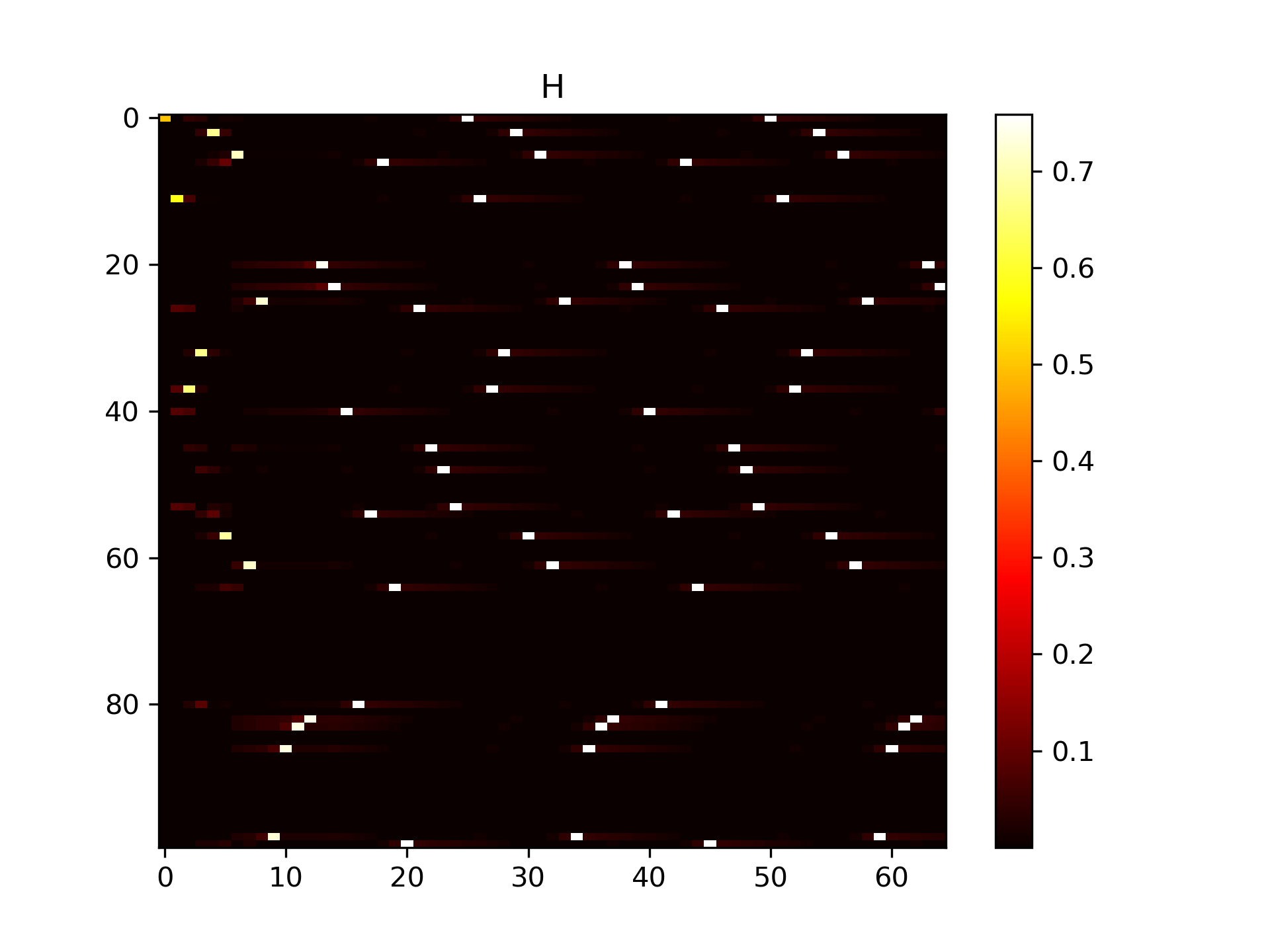}
      \caption{Inferred hidden states $H$}
      \label{fig:factorization_after_gen_from_seed_H}
  \end{subfigure}
  \caption{The factorization $V_{predicted} = W H$ corresponding to Eq. \ref{eqn:factorized_vanilla_rnn_submats} after generating the sequence from the seed.}
  \label{fig:factorization_after_gen_from_seed}
\end{figure}

We then run the inference procedure starting from the first time slice, since the hidden states need to be inferred for all time slices. The generation procedure works as follows. For each time slice $k$, we iterate the NMF updates until the current state vector $h_k$ converges. We copy the inferred $h_k$ into the next time slice of the $H_{prev}$ sub-matrix of $V$. If $k$ is less than the seed length, we leave $y_k$ as is (since it is part of the seed). Otherwise, we also update the current $y_k = W_h h_k$. We also copy this predicted $y_k$ into the $x_{k+1}$ position in the next time slice of $X$, which serves to propagate the predicted sequence vectors forward. However, we should note that we do not sample from the predicted $y_k$ and instead simply copy the predicted vector directly into the following time slice. We then increment $k$ to the next time slice and so on until finally reaching the end of the generated sequence length. Figure \ref{fig:generated_seq} shows the resulting generated sequence including the seed. From this we see that the model can successfully generate the memorized sequence, with a small amount of noise which is seen as slight yellow or red in the ``hot'' colormap that we used. Figure \ref{fig:factorization_after_gen_from_seed} shows all three matrices $V_{predicted} = W H$ corresponding to the factorization in Eq. \ref{eqn:factorized_vanilla_rnn_submats} after generating the sequence from the seed.

\subsection{Copy task}
\label{sec:copy_task}

We now test the factorized RNN and compare to the vanilla RNN on the more difficult Copy Task \cite{hochreiter1997long} using the setup from \cite{arjovsky2016unitary}. This involves generating a random sequence of tokens which are supplied one at a time in each time slice to the network. After this we supply a padding token for some fixed number of time slices. We then supply a ``remember'' token for a number of time slices equal to the length of input sequence that was supplied earlier, during which the network must attempt to output the remembered tokens of the input sequence. This task thus tests the network's ability to recall information that was presented earlier and it gets more difficult as the task spans longer time intervals. We use the same parameters as \cite{arjovsky2016unitary} in which the input sequence is of length 10 and there are 10 distinct token values, which we supply to the network as 1-hot vectors. This combined with the pad and remember tokens lead to 12 distinct token values in total, so that the input 1-hot vectors will be 12-dimensional. We can then still adjust the difficulty by controlling the padding length $T_{pad}$.

\subsubsection{Training details for the factorized RNN}

We used the same model and training hyperparameters as in Section \ref{sec:standard_nmf_rnn}. They only differ in how the loss is applied. In the model of Section \ref{sec:standard_nmf_rnn}, the RNN outputs a prediction at every time slice. However, in the copy task we only care about the predicted tokens during the time slices when the ``remember'' token is being supplied and so we only compute the prediction loss of these time slices here. Note that since we are using the NMF learning (i.e., NMF $W$ update rule) to learn the weights, this corresponds to having implicit MSE reconstruction loss terms on all of the hidden states and inputs as well. We set the hidden state dimension to $R = 1024$. We used 100 NMF iterations per time slice for inference.

With the factorized RNN and non-negative parameters (i.e., NMF), most training runs resulted in perfect validation accuracy for $T_{pad} = 5$. With $T_{pad} = 10$, multiple training runs were needed to reach perfect accuracy. For $T_{pad} = 15$, accuracy was no better than chance level. When we tried allowing negative parameters, the models failed to do better than chance accuracy. These results seem interesting in that the network is able to successfully learn the task with up to 10 padding tokens, even though there is no BPTT-like backward flow of error gradients.

\subsubsection{Training details for the vanilla RNN}

For comparison, we also train a vanilla RNN on the same task. For this we use the network described in Section \ref{sec:vanilla_rnn_review} except that we used LayerNorm on the hidden states as suggested for RNNs in \cite{ba2016layer}. We use the GELU activation. We used the same hidden state dimension $R = 1024$ as the factorized network. Negative-valued parameters were allowed as an all other experiments with MLP-based models. We observed that weight decay of 1e-4 was needed for the best results, although we did not do much hyperparameter tuning.

With the vanilla RNN, we were also able to get perfect accuracy for $T_{pad} = 5$ when BPTT was used. We found that perfect accuracy was possible up to approximately $T_{pad} = 15$. However, when we disabled BPTT, we were not able to get perfect accuracy for any padding length. We were not able to do much hyperparameter tuning, though, and so it remains possible that performance could improve with better tuning.

\subsection{Sequential MNIST classification}
\label{sec:seq_mnist}

The Sequential MNIST classification task is a common task used for evaluating the performance of RNNs. It adapts the MNIST dataset \cite{lecun1998gradient}, which consists of 28x28 pixel grayscale images of handwritten digits (0 to 9), for a sequential data processing context. In the standard MNIST task, the entire image is presented to the model at once, as in Section \ref{sec:frp_1_layer}. However, in the Sequential MNIST task, the image is presented as a sequence of pixels, typically in a row or column-wise manner. We evaluate the column-wise version in this experiment. The image is unrolled column by column, resulting in a 28-time-slice sequence, where each slice is a 28-dimensional vector representing one column of pixels. Each time slice of the RNN outputs a 10-dimensional class prediction vector, although we only compute the loss on the final time slice of the sequence, ignoring the model predictions from earlier time slices.

\subsubsection{Training details}

For the factorized RNN, we used 10 unrolled inference iterations since it allowed us to run more experiments. We noticed only slightly reduced accuracy in one experiment compared to using 50-100 iterations and so we made the trade-off based on our limited computational resources. We used a learning rate of 5e-5 and weight decay of 1e-5 for all of the training runs and for both the factorized and vanilla RNNs. Similar to the other experiments, we used the input reconstruction loss on the factorized model. We trained under both the NMF and semi-NMF parameter constraints. We trained both the factorized and vanilla RNNs with and without BPTT, and for hidden state dimensions of 512 and 2048.

\subsubsection{Results}

Table \ref{table:sequential_mnist_comparison} shows the accuracy results on this task. For the results with BPTT disabled, the factorized RNN has significantly higher accuracy compared to the MLP (96.23\% vs 83.33\% at hidden dimension = 512, and 97.16\% vs 94.06\% at hidden dimension = 2048), although this required the semi-NMF parameter constraint. Using the NMF constraint without BPTT produced significantly worse accuracy. We were somewhat surprised to see both the factorized and vanilla RNNs performing so well without BPTT. Both models performed better with BPTT enabled, but only slightly, and we see that the factorized RNN slightly outperformed the vanilla RNN. Interestingly, when using BPTT, the factorized RNN performed similarly under both the NMF and semi-NMF constraints. Notice also that when BPTT is disabled, both the factorized and vanilla RNNs suffer a more severe accuracy degradation in going from 2048 to 512 hidden dimension size, compared to when BPTT is enabled; It seems that disabling BPTT could be making the models less parameter efficient.

\begin{table}[h]
  \centering
  \caption{Comparison of factorized vs vanilla RNNs on the Sequential MNIST task for various hyperparameter settings.}
  \label{table:sequential_mnist_comparison}
  \begin{tabular}{lcccc}
      \toprule
      Model  & BPTT & Hidden State Dimension & Parameter Constraints & Test Accuracy \\
      \midrule
      Factorized RNN & No & 512 & NMF  & 77.43\% \\ 
      Factorized RNN & No & 512 & semi-NMF & 96.23\% \\
      Factorized RNN & No & 2048 & NMF & 75.81\% \\
      Factorized RNN & No & 2048 & semi-NMF & 97.16\% \\
      Vanilla RNN & No & 512 & N/A & 83.33\% \\
    Vanilla RNN & No & 2048 & N/A & 94.06\% \\
    Factorized RNN & Yes & 512 & NMF  & 98.56\% \\ 
      Factorized RNN & Yes & 512 & semi-NMF & 98.16\% \\
      Factorized RNN & Yes & 2048 & NMF & 98.66\% \\
      Factorized RNN & Yes & 2048 & semi-NMF & 99.00\% \\
      Vanilla RNN & Yes & 512 & N/A & 97.94\% \\
      Vanilla RNN & Yes & 2048 & N/A & 98.66\% \\
      \bottomrule
  \end{tabular}
\end{table}

\subsection{Audio source separation on MUSDB18}
\label{sec:audio_separation}

Audio tends to be well modeled as a mixture of source components. Much of the music we listen to is explicitly mixed together from isolated recordings (often called stems). In the source separation problem, we are interested in separating multiple sources that have been mixed together. For example, suppose we have a recording by a small band in which the vocals, guitar, and drums have all been captured together in a single audio channel. A source separation system would then be tasked with taking this recording as input and producing an output containing only a single desired source such as the vocals.

A useful inductive bias in this case would be to have the model $f()$ satisfy the linear superposition property: $alpha1 * f(x_1) + alpha2 * f(x_2) = f(alpha1 * x_1 + alpha2 * x_2)$. Any scaled mixture of the two sources should produce the corresponding scaled output. This is a desirable property since it implies that if the model is capable of recognizing the sources in isolation, it automatically generalizes to handling the mixture case. Such a property could potentially allow the model to generalize better from limited training data.

NMF can potentially satisfy this property, and there are several existing works in which it has been applied to related audio problems \cite{Grindlay_Ellis_2009}. Since our factorized RNN is essentially a sequential extension of matrix factorization, it seems interesting to apply it to the source separation task and compare its generalization performance against a standard vanilla RNN (which does not have the superposition property due to its use of non-linear activation functions). We train both factorized and vanilla RNNs on the source separation task using the MUSDB18 dataset. MUSDB18 contains 150 music tracks of various genres corresponding to approximately 10 hours of music, split into a training dataset containing 100 tracks and a test dataset containing 50 tracks. It includes the isolated tracks (stems) for drums, bass, other, and vocals. This makes it straightforward to use for training and evaluating source separation models. 

\subsubsection{Training details}

For training we mix together the isolated stems to create the inputs and then supply just one of the stems as the target. In this experiment we chose to use vocals as the target. The audio is aligned during validation and testing but not aligned during training to provide more variation in the examples. During training only, each source is scaled by random per-example values between 0.5 and 2.0.

We used the same MSE output prediction loss (i.e., regression loss) for both models to ensure that the resulting MSE test loss would be comparable between the factorized and vanilla RNN. That is, the factorized RNN only used the MSE output prediction loss term, without the input reconstruction loss term in this experiment. For the audio features, we use the following hyperparameters: The sample rate was 44100 Hz, the short-time Fourier transform (STFT) used a 2048 window size and 1024 hop size. We limited tha audio feature vectors (time slices of the STFT) to contain only the lowest 350 frequency bins. Negative parameter values were allowed in both models. BPTT was used in both models. The hidden state dimension was 1024 in both models. The batch size was 100. The learning rate was 5e-4 in the factorized RNN as 1e-4 for the vanilla RNN. Weight decay was 5e-5 for both models.

\subsubsection{Results}

The vanilla RNN had a MSE test loss of 3.657e-3. The factorized RNN had a test loss of 2.497e-3. These were averaged over 3 runs. Both models seem to be underfitting since the training loss converges to a similar range as the validation loss. Although the factorized RNN produced a better (lower) test loss compared to the vanilla RNN, we should note that both models perform significantly below state of the art on this dataset due to the use of a the simple RNN architecture.

\section{Related work}
\label{sec:related_work}

\subsection{Positive Factor Networks}
\label{sec:related_pfn}

Perhaps the single most related existing work is our previous work on Positive Factor Networks \cite{vogel2008positive}, which similarly proposed NMF-based building blocks that can be composed to build expressive and interpretable architectures. Although the factorized equivalent of the vanilla RNN that we introduce in this paper was not considered, various other factorized-RNN-like architectures (which were referred to as \emph{dynamic positive factor networks}) were considered, including some more sophisticated architectures such as a sequential data model for target tracking, and a model employing dilated deconvolutional layers in a Wavenet-like \cite{oord2016wavenet} architecture. A block with the same factorized model as our PFC block appears in Eq 3.1 of \cite{vogel2008positive}, which was referred to a ``coupling module'' or ``coupling factorization'' in that paper. Similar iterative NMF algorithms were used to perform inference in these models. However, a key distinction between that work and our present work is that it did not use backpropagation for learning the parameters, and instead relied on applying the NMF left-update rules to learn them. We also experimented with NMF left-update rules in the current work for learning the parameters as an ablation in Sections \ref{sec:standard_nmf_rnn}, \ref{sec:copy_task}, \ref{sec:seq_mnist}, but found them to underperform backpropagation on the more challenging learning tasks. As a result of not using backpropagation for parameter learning, the models presented in \cite{vogel2008positive} failed to produce results competitive with other approaches on supervised learning tasks. The PFC block and corresponding neural networks constructed from them that we consider in the present work can therefore be considered as unrolled positive factor networks, or positive factor networks employing backpropagation-based training.

\subsection{Non-negative matrix factorization}

NMF was originally proposed by \cite{paatero_1994} as \emph{positive matrix factorization} and later popularized by \cite{Lee_seung} \cite{lee2000algorithms}. NMF is related to other methods such as sparse coding \cite{olshausen1997sparse} that use a similar dictionary learning model but with differences in the modeling constraints, regularizers, and/or algorithms used. NMF is often observed to provide parts-based decompositions of data, which can be useful when interpretable decompositions are desired. It also potentially satisfies the (non-negative) linear superposition property and as a result has been applied to audio processing tasks such as source separation and music transcription in which the audio features in the input data matrix are assumed to be well modeled as an additive combination of audio sources (i.e., individual instruments and/or notes) \cite{Grindlay_Ellis_2009}.

Our PFC block is related in that its declarative model corresponds to a \emph{masked predictive NMF} in which the data matrix is partitioned along the row dimension into ``input'' and ``output'' vectors of the block, with the right factor matrix $H$ corresponding to inferred hidden activations. The output vectors as masked during recognition (during inference of $H$) and the resulting inferred $H$ is then used to predict the outputs. This allows our block to retain the additive superpositional NMF model, while also supporting the construction of more expressive differentiable architectures such as factorized RNNs. The resulting neural networks (or positive factor networks) can then be interpreted as an extension of NMF that increases its modeling expressiveness and suitability for supervised learning tasks.

\subsection{Unrolled neural networks}

The key enabler of the increased predictive performance of our present work compared to the Positive Factor Networks of \cite{vogel2008positive} is the modification of the learning algorithm to use backpropagation instead of relying on NMF left-update steps. As discussed in Section \ref{sec:factorized_layers}, backpropagation-based training is enabled by unrolling the iterative NMF inference steps into an RNN-like structure in the computation graph, making the PFC block differentiable and therefore compatible with backpropagation training.

This process of unrolling an iterative optimization algorithm into a neural network and training with backpropagation is referred to as \emph{algorithm unrolling} or \emph{unrolled neural networks} in the literature \cite{monga2021algorithm}. An existing example of unrolled NMF appears in \cite{nasser2021deep}. The idea that improved results could be obtained on supervised tasks by adding an additional task-specific loss function to an iterative and differentiable optimization algorithm and using it to optimize the parameters was proposed in \cite{mairal2011task} and \cite{rolfe2013discriminative}, with earlier related ideas being proposed in \cite{bengio1995recurrent} and \cite{seung1997learning}. More recent works have explored the use of unrolled convolutional sparse coding for improved robustness to corrupt and/or adversarial input images \cite{sulam2020adversarial}, \cite{li2022revisiting}. Of these, \cite{sun2018supervised}, \cite{li2022revisiting} also use FISTA to accelerate the convergence of the unrolled inference. As far as we are aware, ours is the first work to explore the use of a modular unrolled block (PFC block) supporting the construction of arbitrary neural architectures, unrolled factorized RNNs, and the first to explore the interpretability advantages of unrolled NMF-based networks for continual learning.

\subsection{Nearest neighbor classification and regression}

As discussed in Section \ref{sec:knn_review}, our PFC block shares some similarities with classification and regression based on the k-nearest neighbors (k-NN) algorithm \cite{Cover1967NearestNP}. We show that the PFC block can be derived by first formulating k-NN prediction as a matrix-vector product, and then generalizing and interpreting it as a matrix factorization.

\subsection{Learning Vector Quantization (LVQ)}
Learning Vector Quantization (LVQ) \cite{kohonen1988learning} \cite{kohonen1990improved} is a prototype-based method that extends the k-nearest neighbors algorithm by introducing the concept of learnable prototypes. An advantage of this approach over matrix-factorization-based methods is its faster recognition since an iterative algorithm is not used. We are not aware of a fully differentiable version of LVQ that could match MLP predictive performance and make it possible to use as a building block for more complex architectures such as RNNs, however.

\subsection{Future research}

This work is preliminary and we leave several unanswered questions and possibilities for future research directions. We list some of them here:

\begin{itemize}
    \item As discussed in \cite{Achler2012AFO}, methods such as ours that perform recognition by iterating to a fixed point can potentially make use of adaptive computation. This could potentially be used to improve recognition efficiency, as well as providing an additional form of confidence estimation, since ``difficult'' inputs could require more iterations to converge.
    \item When unrolling the NMF inference updates, we observed that memory usage can potentially be reduced by computing the initial several iterations ``without gradients'' and only unrolling the last several iterations ``with gradients''. In some cases, this resulted in improved efficiency, but in others it resulted in reduced accuracy, and so a better understanding is still needed.
    \item Although we found FISTA to be one effective method in accelerating the NMF inference step of the PFC block, it could be interesting to also consider other approaches to further accelerate the inference.
    \item Our sliding learnable window optimizer introduced in Section \ref{sec:slw_optimizer} enabled improved continual learning in PFC-based models. However, more sophisticated approaches based on basis-vector-specific learning rates could potentially lead to improved continual learning performance and/or parameter efficiency. For example, it could be interesting to consider methods that identify the important or in-use basis vectors and reduce their learning rates accordingly so as to make them more resistant to being overwritten or modified later in training.
    \item We did not consider any form of sparsity regularization (e.g., L1 penalty) in our experiments, leaving regularization methods other than the non-negativity constraint and weight decay unexplored in the current work.
    \item As Table \ref{table:sequential_mnist_comparison} shows, disabling BPTT in both factorized and vanilla RNNs often resulted in only a small drop in accuracy, although more parameters were needed to recover accuracy. It is also unclear why the semi-NMF parameter constraint resulted in better accuracy compared to NMF when BPTT was disabled. It could be interesting to conduct a more detailed investigation as future research.
    \item The declarative model of the PFC block in Eq. \ref{eqn:nn_pred_both_pfc} is symmetric with respect to the inputs and outputs. This allows us to potentially reverse the prediction direction of the block so that we can consider swapping their roles and make the block compute the ``inputs'' given the ``outputs''. In addition to reversible blocks/layers, it could be interesting to explore the case where only some subset of the input and output are observed and the block then jointly predicts all unobserved elements.
    \item As discussed in Section \ref{sec:factorized_layers}, the PFC block can be interpreted as performing factorized attention over parameters. It could be interesting to also consider its use as an attention block that performs factorized attention over key and value activations (output from other upstream blocks) instead of over parameters, making it more like a factorized version of the attention block used in the transformer. Specifically, in Eq. \ref{eqn:nn_pred_both_pfc} the $W_x$ matrix would be replaced by a \emph{keys} matrix $K_x$ and $W_y$ would be replaced with a corresponding \emph{values} matrix $V_y$ which would then represent output activations from other layers/blocks instead of learnable parameters.
    \item Note that the PFC block has the inductive bias of NMF, which is different than the MLP. In our experiments, we observed it to perform similar to and sometimes better compared to the MLP. However, it is possible that the NMF inductive bias could make PFC-based models particularly well suited or poorly suited depending on the modeling assumptions of the datasets used. Also note that relaxing the non-negativity constraint to semi-NMF could support the use of datasets containing negative values, at the expense of possibly reduced interpretabiity, however. We leave such an exploration as future research.
\end{itemize} 

\section{Conclusion}

In this paper, we presented the Predictive Factorized Coupling (PFC) block, a neural network building block that combines the interpretability of non-negative matrix factorization (NMF) with the predictive performance of multi-layer perceptron (MLP) networks. We demonstrated the versatility of the PFC block by using it to build various architectures, including a single-block network, a fully-connected residual network containing two PFC blocks, and a factorized RNN.

Our experiments showed that the PFC block achieves competitive accuracy with MLPs on small datasets while providing better interpretability. We also demonstrated the benefits of the PFC block in continual learning, training on non-i.i.d. data, and knowledge removal after training. Additionally, we showed that the factorized RNN outperforms vanilla RNNs in certain tasks while providing improved interpretability.

While the PFC block has limitations, such as slower training and inference and increased memory consumption during training, it offers a promising direction for developing more interpretable neural networks without sacrificing predictive performance. Future work includes evaluating the PFC block on larger datasets and exploring its suitability for use in more complex multi-block architectures.

\appendix

\section{Alternating SGD updates for NMF}
\label{sec:sgd_nmf}

One of the simplest methods that can be used to solve for the factors $W$ and $H$ in Equation (\ref{eqn:nmf_v_eq_w_h}) is gradient descent (GD) or stochastic gradient descent (SGD). Whether GD or SGD applies depends on whether we are operating on the full training data or on batches, and so we will simply refer to both cases as SGD in the following. In SGD, we first choose a suitable loss function to quantify the approximation error and then compute its gradients with respect to each factor matrix. We alternately apply small additive update steps in the opposite direction of the gradient to each factor matrix, while clipping any negative values to zero, until convergence. A commonly used choice of loss is the squared Euclidean error and so we use it here. The squared error loss $\mathcal{L}$ between $V$ and the approximation $V_{pred} = W H$ is given by:

\begin{align}
  \mathcal{L}(V || V_{pred}) = \frac{1} {2} \sum_{i,j} (v_{i j} - {v_{pred}}_{i j})^2 = \frac{1} {2} \sum_{i,j} e_{i j}^2
  \label{eqn:nmf_mse}
\end{align}

where $e_{i j} = v_{i j} - {v_{pred}}_{i j}$ is the approximation error of the $i j$'th element of $V$. Let $E$ denote the approximation error matrix with elements $e_{i j}$.  Since $\mathcal{L}$ is differentiable with respect to $W$ and $H$, the loss gradients are:

\begin{eqnarray}
  \frac{\partial \mathcal{L}} {\partial H} = W^T (W H - V) = W^T E \\
  \frac{\partial \mathcal{L}} {\partial W} = (W H - V) H^T = E H^T
\end{eqnarray}

The resulting SGD updates are then given by:

\begin{eqnarray}
  \label{eqn:sgd_HUpdate_with_gradients}
  H &\leftarrow&  relu(H - \eta_H \frac{\partial \mathcal{L}} {\partial H}) \\
  \label{eqn:sgd_WUpdate_with_gradients}
  W &\leftarrow&  relu(W - \eta_W \frac{\partial \mathcal{L}} {\partial W})
\end{eqnarray}

where $\eta_H$ and $\eta_W$ are learning rate hyperparameters which are set to a small nonnegative value. The $relu()$ function is used to prevent negative values in the updated matrices. Substituting the gradients in the above updates gives:

\begin{eqnarray}
  \label{eqn:sgd_HUpdate_with_gradients_expanded}
  H &\leftarrow&  relu(H - \eta_H W^T (W H - X)) \\
  \label{eqn:sgd_WUpdate_with_gradients_expanded}
  W &\leftarrow&  relu(W - \eta_W (W H - X) H^T)
\end{eqnarray}

We then initialize $W$ and $H$ with non-negative noise and iteratively perform the updates until convergence.

\subsection{Normalization and preventing numerical issues}

We observed that numerical issues can sometimes be reduced by additionally replacing any zero-valued elements in the factor matrices with some small minimum allowable non-negative value immediately after the $relu()$ in Eqs. \ref{eqn:sgd_HUpdate_with_gradients_expanded} \ref{eqn:sgd_WUpdate_with_gradients_expanded}, such as $\epsilon = 1e-5$, for example.

We introduce the following two normalization methods, which we have observed to often perform well in our experiments, depending on whether we are performing alternating updates to jointly learn both $W$ and $H$ or performing unrolled inference to infer only $H$, followed by backpropagation-based learning of $W$.

\subsubsection{Normalization for the joint factorization case}

Note that we can scale $W$ by an arbitrary value $\alpha$ if we also scale $H$ by its inverse so that their product is unchanged. Depending on the update algorithm used, it is possible that one of the factor matrices might be scaled slightly larger on each update while the other is scaled slightly smaller so that one of $W$ and/or $H$ often tends toward infinity while the other tends toward zero. To avoid this problem, one of the factor matrices (typically $W$) is typically normalized to have e.g. unit column norm after each update \cite{liu2006nonnegative}. However, this prevents the basis vectors from becoming arbitrarily small, potentially resulting in a less sparse and/or less interpretable solutions.

We have empirically observed that optimization can be easier when the range of values in $W$ and $H$ are similar. We use the Numpy/PyTorch notation in the follow where ``$:$'' denotes selecting all rows or columns in the dimension where it is applied. Intuitively, if a particular column $W[:, i]$ does not contribute significantly to the approximation of $X$, then we would like its values as well as the corresponding activations in row $H[i, :]$ to be small and vice versa. We use the following normalization algorithm to achieve this.

For each column $W[:, i]$ in $W$, we compute its maximum value along with the maximum value of the corresponding row $H[i, :]$ in $H$. We then compute the mean of these two values. We scale the column $W[:, i]$ and row $H[i, :]$ so that their updated maximum value will be equal to this mean. We have found this simple algorithm to work well in our experiments. After training, it tends to result in both $W$ and $H$ having nearly identical maximum values, sparser learned factorizations and/or improved approximation error compared to the other normalization methods that we tried. We must be careful if mini-batch training is used, though, as the above maximum values are intended to be computed over the full $W$ and $H$ matrices. 

\subsubsection{Normalization for unrolled inference with backpropagation}
\label{sec:h_unrolled_normalization}

For the unrolled algorithm in which backpropagation is used together with another optimizer (e.g., RMSprop) to update $W$, we have found the following update method to perform well. The basis idea is that for each column $v_i$ in the data matrix $V$, we limit the maximum value of the corresponding inferred column $h_i$ in $H$ such that it is not allowed to have a maximum value larger than the maximum value in $v_i$. This normalization step involves computing the column-wise maximum values in $V$ at the start of inference. After each unrolled NMF right-update to matrix $H$, we then apply a column scaling step to scale the columns such that their corresponding maximum values do not exceed those of $V$ (if the maximum value is already less, then no scaling if performed). This is intended to prevent the inferred values in $H$ from exploding during the unrolled inference. Recall that backpropagation and another optimizer are used to update $W$. We observed that it was not necessary to explicitly apply any normalization to $W$ since we did not encounter exploding values. Although weight decay was used in the experiments, it was not needed in order to prevent numerical issues, and was only used due to its observed slightly beneficial effect on predictive performance in some cases.

\subsection{Automatic learning rate selection}

We empirically observe that setting the SGD learning rates $\eta_H$ and $\eta_W$ in Eq. \ref{eqn:sgd_HUpdate_with_gradients_expanded} automatically using the same method as in FISTA often works well. Specifically $\eta_H$ is set to $1/L_H$ where $L_H$ is the largest eigenvalue in $W^T W$. Likewise, $\eta_W$ is set to $1/L_W$ where $L_W$ is the largest eigenvalue in $H H^T$. In our experiments, we used the power method to compute the approximate largest eigenvalue.

\section{Implementation details for FISTA-accelerated NMF inference in the PFC block}
\label{sec:fista_details}

We adapt the Fast Iterative Shrinkage-Thresholding Algorithm (FISTA) to accelerate the NMF inference procedure in the PFC block. FISTA is an optimization method that combines gradient-based approaches with an acceleration technique \cite{beck2009fast}. Originally designed for sparse recovery, FISTA can be adapted for various applications, including matrix factorization. We use rectification (relu) as the proximal operator instead of the usual shrinkage threshold operator. The resulting algorithm uses the SGD NMF right-update steps with the FISTA steps that compute the momentum term. Both NMF and semi-NMF constraints are supported.

Given an input matrix $X$, we aim to infer the hidden matrix $H$ in the factorization $X \approx W_x H$. We assume that the weights $W_x$ remain fixed during the inference procedure. In the standard SGD-based NMF procedure reviewed in Section \ref{sec:sgd_nmf}, the inference procedure amounts to the repeated application of the $H$ update step in Eq \ref{eqn:sgd_HUpdate_with_gradients_expanded} followed by the normalization steps and FISTA momentum update steps. The procedure is as follows:

\begin{enumerate}
    \item Initialize \( H^0 \) and set \( Y^1 = H^0, t_1 = 1 \).
    \item For each iteration \( k \), update \( H \) by:
    \begin{enumerate}
        \item Updating \( H \) as \( H^{k+1} = relu(Y^k - \frac{1}{L} W^T (W Y^k - X)) \)
        \item Applying normalization scaling to $H$.
        \item Updating \( t \) as \( t_{k+1} = \frac{1 + \sqrt{1 + 4t_k^2}}{2} \).
        \item Updating \( Y \) as \( Y^{k+1} = H^{k+1} + \left(\frac{t_k - 1}{t_{k+1}}\right) (H^{k+1} - H^k) \).
    \end{enumerate}
    \item Repeat the process until convergence is achieved.
\end{enumerate}

We estimate the Lipschitz constant \( L \) using the power method. The normalization scaling step is described in Section \ref{sec:h_unrolled_normalization}. This step may not always be needed, but we leave it enabled in all experiments to prevent numerical issues in $H$.

\bibliographystyle{spiebib.bst}

\bibliography{vogel_paper_2023}

\end{document}